\PassOptionsToPackage{table}{xcolor}
\documentclass{article} %
\usepackage{iclr2027_conference,times}

\usepackage{amsmath,amsfonts,bm}

\def\eqref#1{equation~\ref{#1}}

\def\1{\bm{1}}

\DeclareMathAlphabet{\mathsfit}{\encodingdefault}{\sfdefault}{m}{sl}
\SetMathAlphabet{\mathsfit}{bold}{\encodingdefault}{\sfdefault}{bx}{n}

\DeclareMathOperator*{\argmax}{arg\,max}
\DeclareMathOperator*{\argmin}{arg\,min}

\usepackage{hyperref}
\usepackage{url}
\usepackage{booktabs}
\usepackage{graphicx}
\usepackage{multirow}
\usepackage{tabularx}
\usepackage{fvextra}
\usepackage{algorithm}
\usepackage{algpseudocode}

\AddToHook{env/figure/begin}{%
  \setlength{\abovecaptionskip}{2pt}%
}

\definecolor{TableLightBlue}{RGB}{232,242,252}
\definecolor{TableLightRed}{RGB}{252,234,234}
\definecolor{TableLightYellow}{RGB}{255,248,220}
\newcommand{\resultstd}[1]{\textsubscript{\(\pm #1\)}}
\newcommand{\tablecitep}[1]{\citep{#1}}
\newcommand{\smalltablecitep}[1]{{\scriptsize\citep{#1}}}
\newsavebox{\tablebox}
\newcommand{\fittable}[1]{%
    \sbox{\tablebox}{#1}%
    \ifdim\wd\tablebox>\linewidth
        \resizebox{\linewidth}{!}{\usebox{\tablebox}}%
    \else
        \usebox{\tablebox}%
    \fi
}

\title{Recurrent Latent Visual Search \\ for GUI Grounding}

\author{Kaiyu Wu, Beichen Zheng, Weiyao Huang, Keze Wang\thanks{Corresponding author (e-mail: kezewang@gmail.com)} \\
Sun Yat-sen University \\
\texttt{\{wuky28,zhengbch5,huangwy95\}@mail2.sysu.edu.cn} \\
\texttt{kezewang@gmail.com}}

\iclrfinalcopy
\begin{document}
\addtocontents{toc}{\protect\setcounter{tocdepth}{-1}}

\maketitle
\lhead{Preprint}

\begin{abstract}

GUI grounding is a critical capability for GUI agents powered by vision-language models, helping them execute user instructions by locating the corresponding elements in screenshots.
Single-step grounding struggles with small elements and dense layouts, motivating multi-step visual search.
However, existing approaches commonly rely on textual reasoning misaligned with visual space or costly multi-round interactions with external visual tools.
To make multi-step visual search an explicit spatial process within the model, we propose \emph{\textbf{ReLaViS}}, which performs \emph{\textbf{Re}current \textbf{La}tent \textbf{Vi}sual \textbf{S}earch} in a single interaction round.
At each step, a spatial search head uses the hidden state to query the screenshot's visual tokens, producing a spatial search distribution that explicitly represents the search focus.
This distribution then aggregates the visual tokens into latent visual evidence, which is recurrently fed back as the next input embedding to condition subsequent search.
We further introduce a GUI-aware coarse-to-fine inductive bias through trajectories constructed from flat element annotations, supervising search from the global interface through intermediate element groups to the target.
Built on Qwen2.5-VL-7B, ReLaViS improves ScreenSpot-Pro accuracy by 3.1 percentage points to 56.3\% with only a 3.5\% increase in inference FLOPs and outperforms the matched single-step baseline on all five benchmarks.

\end{abstract}

\section{Introduction}
\label{sec:introduction}

Advances in large language models~(LLMs) and vision-language models~(VLMs) enable GUI agents to understand user instructions and operate software interfaces~\citep{anthropic2024computeruse,wang2024mobile,cua2025}.
GUI grounding, which locates the interactive element specified by an instruction, is a core capability underlying actions such as clicking~\citep{yang2025aria,yuan2026se}.
Prevailing approaches formulate GUI grounding as single-step localization, directly predicting a coordinate, bounding box, or spatial distribution~\citep{xie2026scaling,tang2026gui,wu2026gui}.
Despite strong performance on standard interfaces, they still struggle with high-resolution screens, small targets, and dense layouts, where accurate grounding requires integrating global structure, local appearance, and instruction semantics~\citep{li2025screenspot,feizi2026grounding}.
To address these challenges, recent work improved grounding through multi-step visual search, including inspecting local regions~\citep{tang2026ui,liu2026drs} or revising predictions~\citep{zhao2025learning,wang2026measure} before returning a final location.

A key design choice is how to represent search states and update them across steps.
\emph{Thinking about images} expresses target semantics and spatial reasoning through chain of thought, yet the textual tokens are not directly aligned with visual space~\citep{shao2024visual,chen2026ui}.
\emph{Thinking with images} makes visual search explicit through tools such as zoom-in or cropping, but incurs additional test-time computation through multiple interaction rounds and image re-encoding~\citep{lei2025gui,zhang2026mvp,tang2026ui}.
\emph{Latent visual reasoning} instead propagates continuous visual representations internally, but existing formulations leave the multi-step spatial search process implicit~\citep{li2026latent,qin2025chain,wang2026monet}.

This raises a question: \emph{Can multi-step visual search become an explicit spatial process within a model's internal computation?}
Our key insight is to represent search as an evolving spatial focus over the screenshot's visual tokens.
Each search state specifies where to search and carries the corresponding visual evidence forward to guide subsequent search.
Search thus proceeds over the same encoded screenshot without additional interaction rounds or image re-encoding.

Beyond representing search states, we need to determine how their spatial focus should evolve toward the target.
Human visual search in GUIs provides useful inspiration, as people often search through likely regions and element groups before identifying the target~\citep{putkonen2025understanding}.
This motivates using the GUI hierarchy as a natural inductive bias to organize multi-step visual search from coarse to fine.
In practice, supervising this GUI-aware coarse-to-fine visual search remains challenging, as annotations of UI trees and search trajectories are scarce.

\begin{figure}[t]
    \setlength{\abovecaptionskip}{5pt} %
    \centering
    \includegraphics[
        width=0.92\linewidth,
        trim={1.87bp 2.28bp 5.96bp 2.62bp},
        clip
    ]{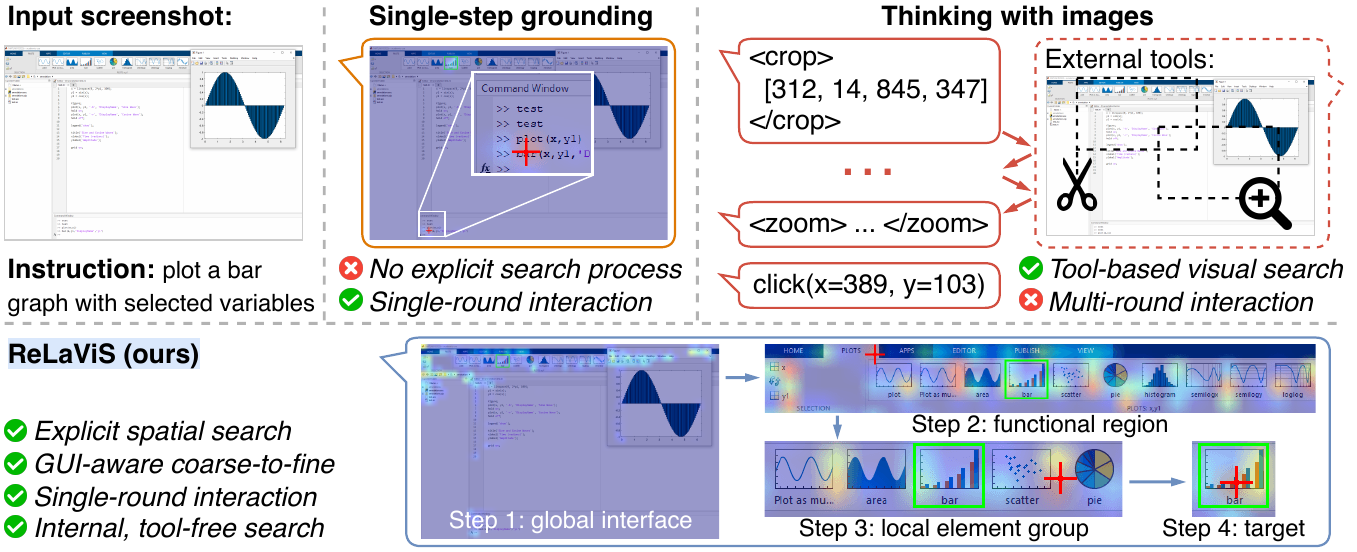}
    \caption{Conceptual comparison of grounding paradigms.
    ReLaViS performs GUI-aware coarse-to-fine visual search by updating explicit spatial search states within a single model interaction.}
    \label{fig:paradigm-comparison}
\end{figure}

Building on these insights, we propose \emph{\textbf{ReLaViS}}, a framework that performs \emph{\textbf{Re}current \textbf{La}tent \textbf{Vi}sual \textbf{S}earch} for GUI grounding (Fig.~\ref{fig:paradigm-comparison}).
Specifically, at each step, a spatial search head uses the hidden state to query the visual tokens, producing a spatial search distribution over them.
This distribution represents the current search focus and then aggregates the visual tokens into latent visual evidence, which is recurrently fed as the next input embedding to condition subsequent search.
We further supervise the search process with GUI-aware coarse-to-fine trajectories constructed from flat element annotations, progressing from the global interface through intermediate element groups to the target.
After the final search step, a deterministic readout converts the distribution into a click coordinate.

\textbf{Our contributions are threefold:}
(i) we formulate GUI grounding as recurrent latent visual search and propose ReLaViS, a framework that recurrently updates explicit spatial search states aligned with visual tokens within a single model interaction, without external tools;
(ii) we construct stepwise supervisory trajectories from flat GUI element annotations to instantiate a GUI-aware coarse-to-fine inductive bias without requiring scarce ground-truth UI trees;
and (iii) ReLaViS outperforms matched single-step baselines across all five GUI grounding benchmarks, including a 3.1-point accuracy gain on ScreenSpot-Pro (53.2\% to 56.3\%) with only 3.5\% more inference FLOPs, while controlled analyses support its key design choices.

\section{Related work}
\label{sec:related-work}

\textbf{GUI agents and GUI grounding.}
\label{sec:gui-agents-and-gui-grounding}
As GUI agents shift from structured metadata to direct screenshot perception, GUI grounding serves as the interface between instructions and actions~\citep{cua2025,anthropic2024computeruse}.
Existing methods differ in how they represent the target location.
\emph{Coordinate-based} methods generate textual coordinates or bounding boxes; prior work improved these predictions through supervised fine-tuning~(SFT)~\citep{wu2025atlas,xie2026scaling}, reinforcement learning~(RL)~\citep{yuan2026se,tang2026gui}, and on-policy self-distillation~\citep{zhang2026learn,huang2026trust}.
\emph{Coordinate-free} methods represent the target location as a spatial distribution over visual tokens; prior work refined its construction and supervision~\citep{wu2026gui,chen2025v2p,zhou2026gui}.
Regardless of the output representation, high-resolution screens, small targets, and dense layouts pose challenges for GUI grounding~\citep{li2025screenspot,feizi2026grounding}, motivating multi-step visual search.
To this end, ReLaViS builds on coordinate-free grounding~\citep{chen2025v2p} to perform supervised multi-step visual search.

\textbf{Thinking about images and thinking with images.}
\label{sec:thinking-about-images-and-thinking-with-images}
Existing multi-step approaches differ in how they represent and update search states.
\emph{Thinking about images} externalizes this process as textual chains of thought that describe target semantics and compare candidates~\citep{tan2026reason,chen2026ui}, yet remains confined to the textual space.
\emph{Thinking with images} instead updates visual evidence through adaptive zoom and region search~\citep{park2025r,wu2025dimo,tang2026ui,yao2026autofocus}, multi-view aggregation~\citep{zhang2026mvp,luo2025visual}, visual tool use~\citep{lei2025gui,chen2026gui,li2026guic2coarsetofineguigrounding}, and iterative visual feedback~\citep{zhao2025learning,li2025chain,wang2026measure}.
It relies on external tools to produce new image views, which are re-encoded and processed in additional model interactions.
This makes spatial search explicit at the observation level but incurs extra test-time computation.
Such limitations motivate ReLaViS, which performs spatial search internally within a single model interaction.

\textbf{Latent visual reasoning.}
\label{sec:latent-visual-reasoning}
Latent reasoning offers a mechanism for such internal search by propagating continuous representations rather than externalizing intermediate steps as textual tokens~\citep{hao2024training,cheng2024compressed}.
Recent work extended this paradigm to visual reasoning by reconstructing visual tokens from regions of interest~\citep{li2026latent}, incorporating perceptual priors from external models~\citep{qin2025chain}, and using auxiliary image supervision to learn visual thoughts~\citep{wang2026monet,yang2026machine}.
Together, these works established latent representations as viable internal visual reasoning states.
However, they do not explicitly track the search focus over the visual token grid or supervise the progression of their latent states as multi-step visual search.
In contrast, ReLaViS specializes latent visual reasoning for visual search by recurrently updating explicit spatial search states aligned with the visual token grid, and supervises the multi-step search process with a GUI-aware coarse-to-fine trajectory for GUI grounding.

\section{Methodology}
\label{sec:methodology}

To make multi-step visual search explicit within the model, ReLaViS recurrently updates spatial search states that track the search focus and carry visual evidence across steps.
Given a GUI screenshot $I$ and an instruction $q$, the model predicts a click coordinate $\hat{\mathbf{c}}=(x,y)$ through $T$ search steps within a single model interaction:
\begin{equation}
    (I,q)\rightarrow S_1\rightarrow\cdots
    \rightarrow S_T\rightarrow \hat{\mathbf{c}},
    \label{eq:overview-chain}
\end{equation}
where $S_t$ denotes the \emph{spatial search state} at step $t$.
In this section, we describe how ReLaViS represents and recurrently updates $S_t$~(Sec.~\ref{sec:recurrent-latent-visual-search}), constructs a GUI-aware coarse-to-fine search trajectory~(Sec.~\ref{sec:gui-aware-search-trajectory-construction}), and uses the resulting trajectory to supervise $S_{1:T}$~(Secs.~\ref{sec:search-process-supervision} and~\ref{sec:training-and-inference}).

\subsection{Recurrent latent visual search}
\label{sec:recurrent-latent-visual-search}

\begin{figure}[t]
    \centering
    \includegraphics[
        width=\linewidth,
        trim={0.47bp 3.40bp 2.63bp 0.00bp},
        clip
    ]{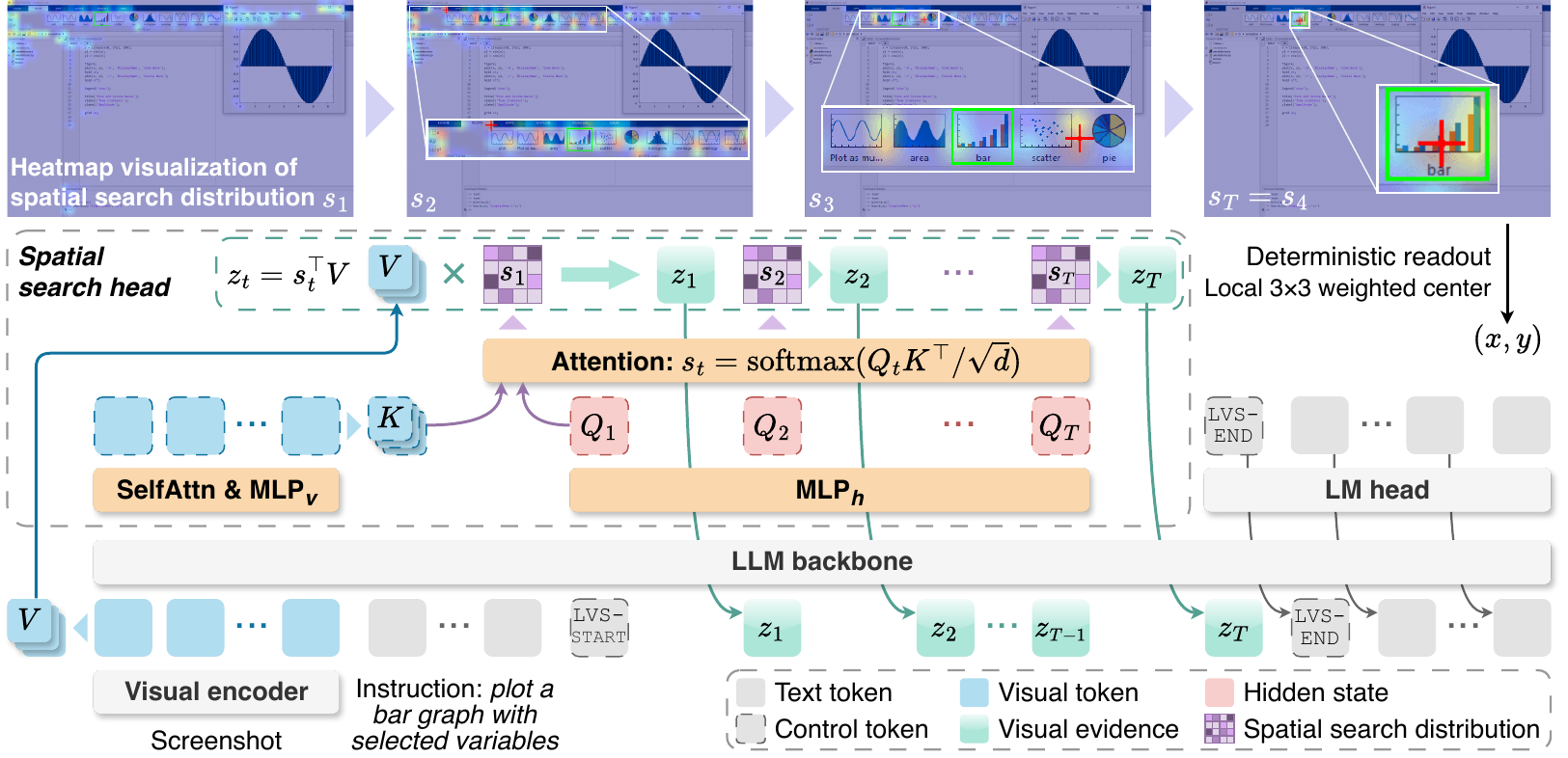}
    \caption{Inference pipeline of ReLaViS.
    At step $t$, the spatial search head attends to visual tokens $V$ using query $Q_t$ derived from the hidden state, producing the search distribution $s_t$.
    This distribution represents the search focus and aggregates $V$ into latent visual evidence $z_t=s_t^\top V$, which serves as the next input embedding to condition subsequent search.
    After $T$ steps, a deterministic readout maps $s_T$ to the click coordinate.
    Within a single autoregressive sequence, the search focus evolves from the global interface through element groups to the target, as illustrated by the top heatmaps.
    }
    \label{fig:relavis-inference-pipeline}
\end{figure}

Fig.~\ref{fig:relavis-inference-pipeline} shows the ReLaViS inference pipeline.
Specifically, we embed latent visual search (LVS) into the response sequence with three control symbols, \texttt{LVS-START}, \texttt{LVS-SLOT}, and \texttt{LVS-END}, delimiting a span $\{\ldots, \texttt{LVS-START}, \texttt{LVS-SLOT}^{\times T}, \texttt{LVS-END}, \ldots\}$.
When the model emits \texttt{LVS-START}, it switches from ordinary token decoding to recurrent latent visual search, with each \texttt{LVS-SLOT} reserving one of $T$ search steps.
After the final slot, \texttt{LVS-END} is inserted and token decoding resumes.
A deterministic readout converts the final state $S_T$ to the click coordinate.

To support explicit spatial representation and recurrent evidence propagation, we instantiate $S_t$ as
\begin{equation}
    S_t=(s_t,z_t),
    \quad s_t\in[0,1]^N,
    \quad \sum_{i=1}^{N}s_{t,i}=1,
    \quad z_t\in\mathbb{R}^{d},
    \label{eq:state-decomposition}
\end{equation}
where $N$ denotes the number of visual tokens and $d$ their representation dimension.
The \emph{spatial search distribution} $s_t$ assigns a normalized probability to each visual token and represents \emph{where} the model currently searches.
The \emph{latent visual evidence} $z_t$ represents \emph{what} the model observes at those locations and replaces the input embedding of the next token, the $t$-th \texttt{LVS-SLOT}.
The injected $z_t$ participates in subsequent autoregressive computation and recurrently conditions the construction of the next state $S_{t+1}$.
We next describe how ReLaViS constructs $s_t$ and $z_t$.

In order to explicitly represent and directly supervise the spatial search distribution on the visual token grid, without relying on textual coordinates, we introduce an attention-based \emph{spatial search head}.
Let $V=[v_1,\ldots,v_N]\in\mathbb{R}^{N\times d}$ denote the visual tokens extracted from the screenshot by the visual encoder and multimodal merger.
At search step $t$, let $h_t\in\mathbb{R}^{d}$ denote the last-layer hidden state at the position immediately before the $t$-th \texttt{LVS-SLOT}: \texttt{LVS-START} for $t=1$ and the preceding \texttt{LVS-SLOT} with input $z_{t-1}$ for $t>1$.
The spatial search head computes $s_t$ as
\begin{equation}
\begin{gathered}
    \widetilde V = \operatorname{SelfAttn}(V) = \operatorname{SelfAttn}(v_1, \ldots, v_N),\\
    Q_t = \operatorname{MLP}_{h}(h_t), \quad
    K = \operatorname{MLP}_{v}(\widetilde V), \quad
    s_{t} = \operatorname{softmax}\!\left(
        \frac{Q_tK^\top}{\sqrt d}
    \right).
\end{gathered}
\label{eq:spatial-search-head}
\end{equation}
Whereas prior coordinate-free methods used this attention distribution for single-step grounding, we predict $s_t$ at every search step and use it to construct latent visual evidence for subsequent search.

Recycling hidden states as inputs can introduce a distributional mismatch with pretrained input embeddings~\citep{deng2025llm}.
Motivated by this concern, we aggregate the original visual token embeddings according to $s_t$ to carry evidence from the current search focus into subsequent search:
\begin{equation}
    z_t=s_t^\top V=\sum_{i=1}^{N}s_{t,i}v_i.
    \label{eq:latent-visual-evidence}
\end{equation}
Because $s_t$ is normalized, $z_t$ remains within the convex hull of the input-space visual embeddings while retaining an explicit spatial interpretation through $s_t$.
Injecting $z_t$ closes the recurrent loop by producing the contextual hidden state at the next search step: $h_{t+1}=\operatorname{VLM}(I,q,z_{\leq t})$.
Appendix~\ref{app:recurrent-latent-visual-search} provides additional details on recurrent latent visual search, including the task template and spatial search head architecture.

\subsection{GUI-aware search trajectory construction}
\label{sec:gui-aware-search-trajectory-construction}

Hierarchy-rich GUI corpora primarily cover mobile or web platforms~\citep{li2024effects,gurbuz2026moving}, while desktop grounding data do not consistently pair UI trees with aligned instructions and target elements~\citep{feizi2026grounding}.
We therefore construct GUI-aware search trajectories from the flat GUI element annotations available in our training samples.
Let $E$ denote all GUI elements in the flat annotations, $b(e)=[x_{0,e},y_{0,e},x_{1,e},y_{1,e}]$ the bounding box of element $e\in E$, and $e^\star$ the~target.
As shown in Fig.~\ref{fig:method-overview}, we represent the trajectory as a nested sequence of candidate element sets:
\begin{equation}
    E=C_1\supseteq C_2\supseteq\cdots\supseteq C_T=\{e^\star\}.
    \label{eq:candidate-trajectory}
\end{equation}

To construct this trajectory, we use the prior that GUI element groups tend to have compact enclosing rectangles.
For a candidate set $C$, let $B(C)$ be its smallest enclosing rectangle and $\Pi(C)$ be the set of bipartitions induced by cuts between adjacent elements sorted by horizontal or vertical center coordinates.
We select the partition that minimizes the total area of the two enclosing rectangles:
\begin{equation}
    (C_a^\star,C_b^\star)
    =\argmin_{(C_a,C_b)\in\Pi(C)}
    \left[\operatorname{Area}(B(C_a))+\operatorname{Area}(B(C_b))\right].
    \label{eq:gui-partition}
\end{equation}
We then recurse into the partition branch containing the target, yielding a raw GUI-aware partition path from the full element set $E$ through spatially compact element groups to $e^\star$.

In order to remove adjacent redundant steps from the raw partition path, we retain key steps based on cumulative reduction in candidate count from the last retained set and local reduction at the next partition, forming a $T$-step trajectory.
The trajectory provides a spatial approximation to the GUI hierarchy, progressing from the global interface through local element groups to the target, as illustrated by trajectory visualizations (Appendix~\ref{app:dataset-visualizations}) and supported by close agreement with ground-truth UI-tree hierarchies (Appendix~\ref{app:agreement-with-ui-tree-hierarchies}).
We then map every element in $C_t$ to its support on the visual token grid.
Details on the trajectory construction are provided in Appendix~\ref{app:gui-aware-search-trajectory-construction}.

\subsection{Search process supervision}
\label{sec:search-process-supervision}

To guide the spatial search states $S_{1:T}$ along the constructed trajectory $C_{1:T}$, we introduce two complementary supervision objectives:
the \emph{candidate-wise coverage (CWC) loss} encourages each $s_t$ to maintain coverage over each candidate element in $C_t$;
the \emph{Fitts-Gaussian peak (FGP) loss} shapes the final distribution $s_T$ within the target element for coordinate readout.

\textbf{Candidate-wise coverage loss.}
\label{sec:candidate-wise-coverage-loss}
Let $\Omega(e)$ be the visual token support of $e$: tokens whose patches intersect with $b(e)$.
We aggregate its probability mass as $m_{t,e}=\sum_{v_i\in\Omega(e)}s_{t,i}$ and define
\begin{equation}
    \mathcal L_{\mathrm{CWC}}^{(t)}
    =\sum_{e\in C_t}
    \max\left(\frac{1}{|C_t|}-m_{t,e},\,0\right).
    \label{eq:cwc-loss}
\end{equation}
By applying a lower bound to each candidate's aggregate probability mass, CWC encourages coverage of each candidate and discourages collapse onto a few.
At the same time, this candidate-wise supervision preserves flexibility to emphasize grounding-relevant visual tokens within each candidate without prescribing a token-level target distribution.
The comparison of supervision objectives in Appendix~\ref{app:intermediate-supervision-objective} shows that CWC supervision outperforms token-level distributional supervision.

\textbf{Fitts-Gaussian peak loss.}
\label{sec:fitts-gaussian-peak-loss}
To capture the center-biased nature of GUI clicking, we construct a normalized target distribution $d_t$ on the visual token grid using 2D Gaussians centered on candidates in $C_t$ and truncated at their boundaries.
Their horizontal and vertical standard deviations match candidate widths and heights.
At the final step, $C_T=\{e^\star\}$, and we apply $d_T$ through the FGP loss:
\begin{equation}
    \mathcal L_{\mathrm{FGP}}^{(T)}
    =D_{\mathrm{KL}}(d_T\,\|\,s_T)
    =\sum_{i=1}^{N}d_{T,i}
    \log\frac{d_{T,i}}{s_{T,i}}.
    \label{eq:fgp-loss}
\end{equation}
While the final-step $\mathcal L_{\mathrm{CWC}}^{(T)}$ places probability mass inside the target element, $\mathcal L_{\mathrm{FGP}}^{(T)}$ shapes the distribution to peak near the target center, facilitating coordinate readout for precise click localization.

The overall training objective is
\begin{equation}
    \mathcal L
    =\mathcal L_{\mathrm{NTP}}
    +\lambda_{\mathrm{FGP}}\mathcal L_{\mathrm{FGP}}^{(T)}
    +\lambda_{\mathrm{CWC}}^{(T)}
        \mathcal L_{\mathrm{CWC}}^{(T)}
    +\lambda_{\mathrm{CWC}}^{(1:T-1)}
        \operatorname{mean}(\mathcal{L}_{\mathrm{CWC}}^{(1:T-1)}).
    \label{eq:overall-objective}
\end{equation}
Here, $\mathcal L_{\mathrm{NTP}}$ is the next-token prediction loss, and the three coefficients independently weight final peak shaping, final target coverage, and candidate-wise supervision at intermediate search steps.
Appendix~\ref{app:search-process-supervision} gives more details of the loss weighting and the construction of $d_t$.

\subsection{Training and inference}
\label{sec:training-and-inference}

The bottom panel of Fig.~\ref{fig:method-overview} summarizes our three-stage training protocol:
Stage~1 (\emph{spatial search head warm-up}) warms up the spatial search head for recurrent training through single-step localization ($T=1$), with the VLM frozen.
Stage~2 (\emph{teacher-forced recurrent training}) jointly trains the spatial search head and the VLM using teacher-forced latent visual evidence $z_t^{\mathrm{gt}}=d_t^\top V$ as input embeddings, allowing the model to learn from the ground-truth visual evidence without propagating unreliable early predictions.
Stage~3 (\emph{self-conditioned recurrent adaptation}) continues joint training with input embeddings $z_t=s_t^\top V$ constructed from predicted $s_t$ through a self-conditioned rollout.
This stage exposes the model to accumulated search errors and reduces the training--inference mismatch.
No-gradient probes predict $s_t$ for input construction, followed by one differentiable forward pass that recomputes $s_{1:T}$ and evaluates the objective in Eq.~(\ref{eq:overall-objective}), avoiding retention of the recurrent computation graph through the LLM backbone and its prohibitive memory cost.

At inference, the model performs $T$ search steps using only screenshot $I$ and instruction $q$ to produce $s_T$.
To reduce patch-level quantization error, a deterministic readout converts $s_T$ into $\hat{\mathbf{c}}$ by taking a probability-weighted average of token centers in the $3\times3$ neighborhood of the highest-probability token.
No element annotations, UI tree, zoom-in or cropping tool, or additional interaction rounds are required.
Appendix~\ref{app:training-and-inference} details the training procedures, cost analysis, and coordinate readout.

\section{Experiments}
\label{sec:experiments}

\textbf{Training data and configuration.}
Following the training-set size used in the controlled comparison in the GroundCUA study~\citep{feizi2026grounding}, we randomly select 100K training samples and a disjoint 5K validation set from GroundCUA.
Each sample contains a screenshot, an instruction, a target element, and flat GUI element annotations with non-nested bounding boxes for trajectory construction.
We initialize ReLaViS-7B from Qwen2.5-VL-7B~\citep{bai2025qwen25vltechnicalreport} and train for one epoch per stage, with default settings $T=4$, $\lambda_{\mathrm{FGP}}=1$, and $\lambda_{\mathrm{CWC}}^{(T)}=\lambda_{\mathrm{CWC}}^{(1:T-1)}=5$.

\textbf{Benchmarks and metrics.}
Evaluation covers ScreenSpot-Pro~\citep{li2025screenspot}, MMBench-GUI-L2~\citep{wang2025mmbench}, OSWorld-G-Refine~\citep{xie2026scaling}, ScreenSpot-v2~\citep{wu2025atlas}, and UI-Vision~\citep{nayak2025ui} (in-domain), which together span high-resolution professional interfaces, standard GUIs, and cross-platform settings.
We use point grounding accuracy~\citep{lin2025showui}: a prediction is correct when its click coordinate lies inside the ground-truth bounding box.

\textbf{Baselines.}
Public baselines include zero-shot general VLMs and coordinate-based or coordinate-free GUI-specialized models evaluated without external tools.
The latter are all based on Qwen2.5-VL-7B, a widely adopted backbone for training-based GUI grounding methods~\citep{xie2026scaling,tang2026gui,wu2026gui}, reducing confounding from pretraining and scale differences.
For controlled comparisons, we train two single-step-7B baselines using the same backbone and data as ReLaViS-7B, with matched training settings: the coordinate-based model generates textual click coordinates, while the coordinate-free $T=1$ variant of ReLaViS predicts a spatial distribution over visual tokens.
We also compare with thinking with images methods that use external tools to obtain new image views across multiple interaction rounds.
Implementation details are in Appendix~\ref{app:implementation-and-reproducibility-details}.

\subsection{Main results}
\label{sec:main-results}

\begin{table}[t]
    \centering
    \caption{Performance comparison across five GUI grounding benchmarks.
    For models trained in this work, results are reported as mean\resultstd{std} over three random seeds.
    ``--'' denotes unavailable results.
    Results with $\dagger$ are reproduced using official implementations; results for other baselines from prior work are taken from the corresponding papers.
    \textbf{Bold} values are best within and outside ``thinking with images'' separately; \underline{underlined} values are second-best only outside it.}
    \label{tab:main-results-all-benchmarks}
    \fittable{%
    \setlength{\tabcolsep}{2pt}
    \begin{tabular}{lccccccc}
        \toprule
        \multirow{2}{*}{\textbf{Model}}
        & \textbf{Training}
        & \textbf{ScreenSpot-}
        & \textbf{MMBench-} & \textbf{OSWorld-} & \textbf{ScreenSpot-}
        & \textbf{UI-} & \multirow{2}{*}{\textbf{Average}} \\
        & \textbf{data size} & \textbf{Pro} & \textbf{GUI-L2} & \textbf{G-Refine}
        & \textbf{v2} & \textbf{Vision} & \\
        \midrule
        \multicolumn{8}{c}{\textit{Zero-shot general VLMs}} \\
        GPT-4o~\tablecitep{hurst2024gpt} & --
        & 0.8 & 2.9 & -- & 80.7 & 1.4 & -- \\
        Operator~\tablecitep{cua2025} & --
        & -- & -- & 40.6 & 70.5 & -- & -- \\
        Qwen2.5-VL-3B$^\dagger$~\tablecitep{bai2025qwen25vltechnicalreport} & --
        & 18.4 & 55.2 & 38.8 & 69.6 & 7.4 & 37.9 \\
        Qwen2.5-VL-7B$^\dagger$~\tablecitep{bai2025qwen25vltechnicalreport} & --
        & 28.8 & 71.9 & 44.0 & 88.4 & 14.7 & 49.5 \\
        \midrule
        \multicolumn{8}{c}{\textit{GUI-specialized models (coordinate-based)}} \\
        SE-GUI-7B~\tablecitep{yuan2026se} & 3K
        & 47.3 & 34.5 & 33.9 & 90.3 & 16.7 & 44.5 \\
        Jedi-7B~\tablecitep{xie2026scaling} & 4M
        & 39.5 & 70.4 & 54.1 & 91.7 & 24.8 & 56.1 \\
        UI-TARS-1.5-7B~\tablecitep{qin2025ui} & --
        & 49.6 & 64.3 & \textbf{64.2} & 89.7 & 20.8 & 57.7 \\
        GUI-G$^2$-7B~\tablecitep{tang2026gui} & 100K
        & 47.5 & 79.5 & 61.9 & \underline{93.3} & 25.6 & 61.6 \\
        InfiGUI-G1-7B~\tablecitep{liu2026infigui} & 44K
        & 51.9 & \underline{80.8} & 59.9 & \textbf{93.5} & 26.1 & 62.4 \\
        \midrule
        \multicolumn{8}{c}{\textit{GUI-specialized models (coordinate-free)}} \\
        GUI-Actor-7B~\tablecitep{wu2026gui} & 9.6M
        & 44.6 & 70.9 & 47.0 & 92.1 & 21.9 & 55.3 \\
        V2P-7B~\tablecitep{chen2025v2p} & 9.6M
        & 52.5 & 79.9 & 52.5 & 92.4 & 28.8 & 61.2 \\
        \midrule
        \rowcolor{TableLightBlue}
        \multicolumn{8}{c}{\textit{Controlled comparison (identical training data)}} \\
        \rowcolor{TableLightBlue}
        Single-step-7B (coordinate-based) & 100K
        & 46.3\resultstd{1.40} & 78.7\resultstd{0.30} & 59.6\resultstd{0.20} & 90.0\resultstd{0.75} & 33.4\resultstd{0.35} & 61.6\resultstd{0.38} \\
        \rowcolor{TableLightBlue}
        Single-step-7B (coordinate-free) & 100K
        & \underline{53.2}\resultstd{0.52} & 80.2\resultstd{0.55} & 59.3\resultstd{1.18} & 90.4\resultstd{0.48} & \underline{43.1}\resultstd{0.42} & \underline{65.3}\resultstd{0.31} \\
        \rowcolor{TableLightBlue}
        \textbf{ReLaViS-7B} & 100K
        & \textbf{56.3}\resultstd{0.29} & \textbf{81.1}\resultstd{0.08} & \underline{62.5}\resultstd{1.01} & 90.7\resultstd{0.21} & \textbf{49.7}\resultstd{0.35} & \textbf{68.1}\resultstd{0.25} \\
        \midrule
        \rowcolor{TableLightYellow}
        \multicolumn{8}{c}{\textit{Thinking with images (external tools, multi-round)}} \\
        \rowcolor{TableLightYellow}
        MVP$^\dagger$~\smalltablecitep{zhang2026mvp} + Qwen2.5-VL-7B & --
        & 46.7 & 55.0 & 51.2 & 73.3 & 19.6 & 49.2 \\
        \rowcolor{TableLightYellow}
        UI-Zoomer$^\dagger$~\smalltablecitep{tang2026ui} + Qwen2.5-VL-7B & --
        & 39.9 & 74.2 & 54.9 & 90.7 & 23.9 & 56.7 \\
        \rowcolor{TableLightYellow}
        GUI-Spotlight$^\dagger$~\tablecitep{lei2025gui} & 18.5K
        & 53.3 & 72.2 & 56.7 & 88.8 & 23.5 & 58.9 \\
        \rowcolor{TableLightYellow}
        GUI-Cursor~\tablecitep{zhao2025learning} & 8K
        & 56.5 & -- & 58.0 & \textbf{93.9} & 25.7 & -- \\
        \rowcolor{TableLightYellow}
        Zoom-in + Qwen2.5-VL-7B & --
        & 49.8 & 77.3 & 59.2 & 88.1 & 23.9 & 59.6 \\
        \rowcolor{TableLightYellow}
        Zoom-in + single-step-7B (coordinate-free) & 100K
        & 62.3\resultstd{0.33} & 80.5\resultstd{0.11} & \textbf{68.3}\resultstd{1.07} & 84.7\resultstd{0.41} & 47.3\resultstd{0.44} & 68.6\resultstd{0.21} \\
        \rowcolor{TableLightYellow}
        \textbf{Zoom-in + ReLaViS-7B} & 100K
        & \textbf{63.5}\resultstd{0.57} & \textbf{80.9}\resultstd{0.16} & 68.2\resultstd{0.54} & 84.9\resultstd{0.59} & \textbf{52.3}\resultstd{0.36} & \textbf{70.0}\resultstd{0.19} \\
        \bottomrule
    \end{tabular}%
    }
\end{table}

As shown in Table~\ref{tab:main-results-all-benchmarks}, ReLaViS outperforms both matched single-step baselines on all five benchmarks.
ReLaViS raises average accuracy to 68.1\% from 61.6\% for the coordinate-based baseline and 65.3\% for the stricter coordinate-free baseline, gains of 6.5 and 2.8 percentage points (pp), respectively.
Among the compared methods without external tools, ReLaViS achieves the highest five-benchmark average (68.1\%) and average over the four out-of-domain benchmarks excluding UI-Vision (72.7\%, versus 71.5\% for the next-best InfiGUI-G1-7B).
These results highlight its strong overall performance across benchmarks and platforms.
Notably, on ScreenSpot-Pro, ReLaViS leads methods without external tools with 56.3\% accuracy, outperforming the coordinate-free baseline by 3.1~pp with only 3.5\% more inference FLOPs, demonstrating its advantage in professional and complex desktop interfaces at comparable inference cost (Table~\ref{tab:inference-cost}).
On ScreenSpot-v2, however, the gain is only 0.3~pp, indicating that the improvement varies across benchmarks and motivating an analysis of its association with target size.
Complete benchmark results are in Appendix~\ref{app:complete-benchmark-results}.

\begin{figure}[t]
    \centering
    \includegraphics[
        width=0.495\linewidth,
        trim={5.32bp 1.40bp 7.35bp 4.20bp},
        clip
    ]{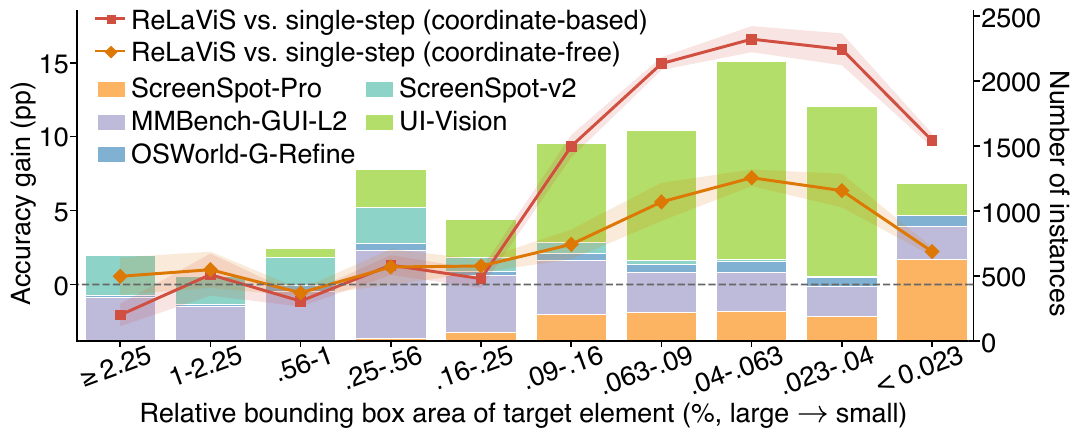}
    \hfill
    \includegraphics[
        width=0.495\linewidth,
        trim={3.69bp 2.86bp 3.38bp 0.00bp},
        clip
    ]{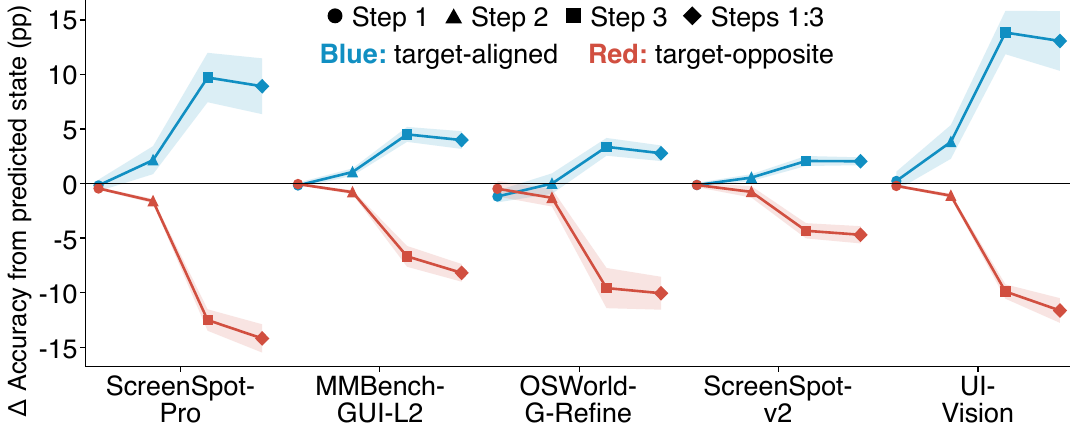}
    \caption{\emph{Left}: pooled accuracy gains of ReLaViS-7B over single-step baselines across target area bins; stacked bars show instance counts and benchmark composition.
    \emph{Right}: accuracy changes relative to predicted-state inference when target-aligned or target-opposite spatial search states are injected at $t=1,2,3$ individually or jointly.
    Curves show multi-seed means with std shading.}
    \label{fig:target-area-and-state-injection}
\end{figure}

\textbf{Where does ReLaViS help?}
\label{sec:where-does-relavis-help}
Fig.~\ref{fig:target-area-and-state-injection} (left) pools instances from all five benchmarks and bins them by relative target bounding box area.
ReLaViS performs comparably to both controlled single-step baselines in the three largest-target bins, while generally achieving larger gains in the smaller-target bins.
Benchmark composition exhibits a corresponding pattern: ScreenSpot-v2 contributes mainly to the larger-target bins, whereas UI-Vision and ScreenSpot-Pro contribute predominantly to the smaller-target bins.
This association is consistent with the smaller gain on ScreenSpot-v2 and larger gains on UI-Vision and ScreenSpot-Pro.
Within-benchmark results in Appendix~\ref{app:target-size-within-benchmark} show a similar tendency, further supporting this association and suggesting greater gains on small targets.

\textbf{How does ReLaViS compare to thinking with images?}
Across all five benchmarks, ReLaViS achieves an average accuracy of 68.1\%, exceeding four thinking with images baselines: MVP (49.2\%), UI-Zoomer (56.7\%), GUI-Spotlight (58.9\%), and zero-shot zoom-in (59.6\%).
ReLaViS requires one interaction round, compared with averages of 4.8, 2, 2.3, and 2 rounds for these baselines.
On ScreenSpot-Pro, ReLaViS achieves 56.3\% accuracy, outperforming these four baselines, while its latency is 2.19~s/sample, compared with 11.97, 2.34, 9.69, and 4.67~s/sample under the respective implementations, highlighting its favorable accuracy--efficiency trade-off.
Furthermore, adding zoom-in to ReLaViS raises its five-benchmark average to 70.0\%, exceeding zoom-in + single-step (coordinate-free) by 1.4~pp and leading the thinking with images group on three of five benchmarks.
These gains support the complementarity of acquiring new image views externally and organizing visual evidence internally.
However, on MMBench-GUI-L2 and ScreenSpot-v2, ReLaViS alone achieves comparable or higher accuracy than its zoom-in variant while completing grounding in a single interaction round without external tools or additional image re-encoding.

\subsection{Ablations and analyses}
\label{sec:ablations-and-analyses}

To evaluate the effectiveness of the key design choices of ReLaViS, we conduct ablations, interventions, and analyses organized around six research questions.
Table~\ref{tab:ablations} summarizes the ablation results.
Appendix~\ref{app:additional-results-and-analyses} provides further support through ablations and analyses, including studies of coordinate readout, trajectory construction, model scale, and search supervision.

\begin{table}[t]
    \centering
    \caption{Ablation results on ScreenSpot-Pro.
    \emph{Left}: search trajectory and latent visual evidence.
    \emph{Right}: search steps and training stages.
    Empty cells inherit ReLaViS settings; $\checkmark\checkmark$ denotes two epochs.
    \ensuremath{\Delta} denotes the Avg. change from full ReLaViS (pp).}
    \label{tab:ablations}
    \fittable{%
    \begin{tabular}{@{}c@{\hspace{0.02\linewidth}}c@{}}
    \begingroup%
        \setlength{\tabcolsep}{2pt}%
        \begin{tabular}[b]{lccccc}
            \toprule
            \multirow{2}{*}{\textbf{Group}}
            & \multirow{2}{*}{\textbf{Steps $T$}}
            & \textbf{Search}
            & \textbf{Visual}
            & \multirow{2}{*}{\textbf{Avg.}}
            & \multirow{2}{*}{\ensuremath{\boldsymbol{\Delta}}} \\
            & & \textbf{trajectory} & \textbf{evidence $z_t$} & & \\
            \midrule
            \cellcolor{TableLightBlue}\textbf{ReLaViS}
            & \cellcolor{TableLightBlue}4
            & \cellcolor{TableLightBlue}GUI-aware
            & \cellcolor{TableLightBlue}$s_t^\top V$
            & \cellcolor{TableLightBlue}\textbf{56.3}\resultstd{.29}
            & \cellcolor{TableLightBlue}-- \\
            \midrule
            \multirow{3}{*}{RQ1}
            & 1 & Only target at $T$ & & 53.2\resultstd{.52} & -3.1 \\
            & & Only target at $T$ & & 54.0\resultstd{.48} & -2.3 \\
            & & Repeated target & & 53.9\resultstd{.96} & -2.4 \\
            \midrule
            \multirow{2}{*}{RQ2}
            & & & \texttt{LVS-SLOT} & 54.5\resultstd{.71} & -1.8 \\
            & & & $h_t$ & 55.2\resultstd{.75} & -1.1 \\
            \midrule
            \multirow{2}{*}{RQ4}
            & & Geometric zoom & & 53.8\resultstd{.39} & -2.5 \\
            & & Shuffled GUI-aware & & 53.5\resultstd{.19} & -2.8 \\
            \bottomrule
        \end{tabular}%
    \endgroup%
    &
    \begingroup%
        \setlength{\tabcolsep}{2pt}%
        \begin{tabular}[b]{lc*{3}{w{c}{2em}}cc}
            \toprule
            \addlinespace[3pt]
            \multirow{2}{*}[-0.8ex]{\textbf{Group}}
            & \multirow{2}{*}[-0.8ex]{\textbf{Steps $T$}}
            & \multicolumn{3}{c}{\textbf{Training stage}}
            & \multirow{2}{*}[-0.8ex]{\textbf{Avg.}}
            & \multirow{2}{*}[-0.8ex]{\ensuremath{\boldsymbol{\Delta}}} \\
            \cmidrule(lr){3-5}
            & & \textbf{1} & \textbf{2} & \textbf{3} & & \\
            \addlinespace[0.8pt]
            \midrule
            \cellcolor{TableLightBlue}\textbf{ReLaViS}
            & \cellcolor{TableLightBlue}4
            & \cellcolor{TableLightBlue}$\checkmark$
            & \cellcolor{TableLightBlue}$\checkmark$
            & \cellcolor{TableLightBlue}$\checkmark$
            & \cellcolor{TableLightBlue}\textbf{56.3}\resultstd{0.29}
            & \cellcolor{TableLightBlue}-- \\
            \midrule
            \multirow{2}{*}{RQ5}
            & 3 & & & & 54.8\resultstd{1.05} & -1.5 \\
            & 5 & & & & 55.1\resultstd{0.93} & -1.2 \\
            \midrule
            \multirow{5}{*}{RQ6}
            & & $\checkmark$ & $\checkmark$ & $\times$ & 53.0\resultstd{0.44} & -3.3 \\
            & & $\checkmark$ & $\times$ & $\checkmark$ & 54.7\resultstd{1.22} & -1.6 \\
            & & $\times$ & $\checkmark$ & $\checkmark$ & 54.9\resultstd{0.03} & -1.4 \\
            & & $\checkmark$ & $\checkmark\checkmark$ & $\times$ & 53.6\resultstd{1.20} & -2.7 \\
            & & $\checkmark$ & $\times$ & $\checkmark\checkmark$ & 55.2\resultstd{0.53} & -1.1 \\
            \bottomrule
        \end{tabular}%
    \endgroup%
    \end{tabular}%
    }
\end{table}

\textbf{RQ1: Does coarse-to-fine multi-step search improve grounding?}
To separate the effects of recurrent multi-step computation and supervision, we compare the coordinate-free single-step baseline, two 4-step variants supervised on the target element only at the final step $T$ or at every step, and full ReLaViS with GUI-aware coarse-to-fine supervision.
As shown in Table~\ref{tab:ablations}, increasing $T$ from 1 to 4 with final-step supervision raises accuracy from 53.2\% to only 54.0\%, while repeating target-element supervision at every step yields 53.9\%.
ReLaViS reaches 56.3\%, exceeding these 4-step variants by 2.3 and 2.4~pp, respectively.
GUI-aware coarse-to-fine multi-step visual search thus provides gains beyond additional computation or supervision at every step alone.

\textbf{RQ2: How should visual evidence be propagated?}
We next examine the contribution of recurrent latent visual evidence propagation and compare $z_t=s_t^\top V$ with direct feedback of the contextual hidden state.
Propagating predicted latent visual evidence raises accuracy from 54.5\% with a fixed \texttt{LVS-SLOT} embedding to 56.3\%, indicating a benefit from propagating spatially selected visual evidence under the same trajectory supervision.
Directly feeding back $h_t$ reaches 55.2\%, outperforming the fixed token but remaining 1.1~pp below $s_t^\top V$.
These results support $s_t^\top V$ as an effective representation for recurrent propagation of latent visual evidence.

\textbf{RQ3: Do subsequent steps use preceding spatial search states?}
\label{sec:do-subsequent-steps-use-preceding-spatial-search-states}
To test whether subsequent grounding depends on preceding spatial search states, we intervene on predicted $s_t$ when computing the recurrent feedback $z_t=s_t^\top V$.
We replace $s_t$ at one or all preceding recurrent steps ($t=1,2,3$) with either a Gaussian distribution aligned with the target element or a shape-matched Gaussian shifted to the opposite grid location.
The model still predicts the final distribution $s_T=s_4$, from which the click coordinate is obtained.
Across the five benchmarks, joint target-aligned injection at steps 1:3 raises accuracy by 2.0--13.1~pp, whereas target-opposite injection lowers it by 4.7--14.2~pp (Fig.~\ref{fig:target-area-and-state-injection}, right).
Individual-step effects are small at the first step, modest at the second, and largest at the third.
These effects show that subsequent grounding functionally depends on the feedback derived from preceding spatial search states, with the strongest sensitivity at later recurrent steps.

\textbf{RQ4: Does the GUI-aware spatial prior improve grounding?}
With four steps in both settings, we construct the GUI-aware trajectory by recursively partitioning the candidate set into spatially compact GUI element groups, whereas geometric zoom linearly contracts a box from the full screenshot to the target bounding box.
Our GUI-aware trajectory reaches 56.3\%, exceeding geometric zoom at 53.8\% by 2.5~pp and showing that its spatial prior is more effective than generic contraction.
In addition, shuffling the first three GUI-aware steps while fixing the final target reduces accuracy to 53.5\%, 2.8~pp below the full model, highlighting the importance of coarse-to-fine order.

\textbf{RQ5: How does the number of search steps affect grounding?}
Table~\ref{tab:ablations} (right) shows that 4-step training and inference achieve 56.3\%, compared with 54.8\% for $T=3$ and 55.1\% for $T=5$.
These results highlight the importance of an appropriate number of search steps, as grounding accuracy does not increase monotonically with recurrent computation under the current setup.

\textbf{RQ6: Does three-stage training matter?}
\label{sec:does-three-stage-training-matter}
As shown in Table~\ref{tab:ablations}, removing Stage~3 reduces accuracy from 56.3\% to 53.0\%; replacing it with an additional Stage~2 epoch reaches only 53.6\%.
Thus, an extra teacher-forced epoch does not recover the benefit of adaptation to self-conditioned rollouts.
Removing Stage~2 yields 54.7\% with one Stage~3 epoch and 55.2\% even after extending Stage~3 to two epochs, supporting the value of initial recurrent training with reliable ground-truth evidence.
Omitting Stage~1 gives 54.9\%, confirming that spatial search head warm-up provides a useful initialization.
The three stages thus provide complementary benefits through initialization, stable recurrent learning, and training--inference adaptation, beyond additional training epochs alone.

\subsection{Recurrent search dynamics}
\label{sec:recurrent-search-dynamics}

\begin{figure}[t]
    \centering
    \includegraphics[
        width=\linewidth,
        trim={3.27bp 1.76bp 4.83bp 3.38bp},
        clip
    ]{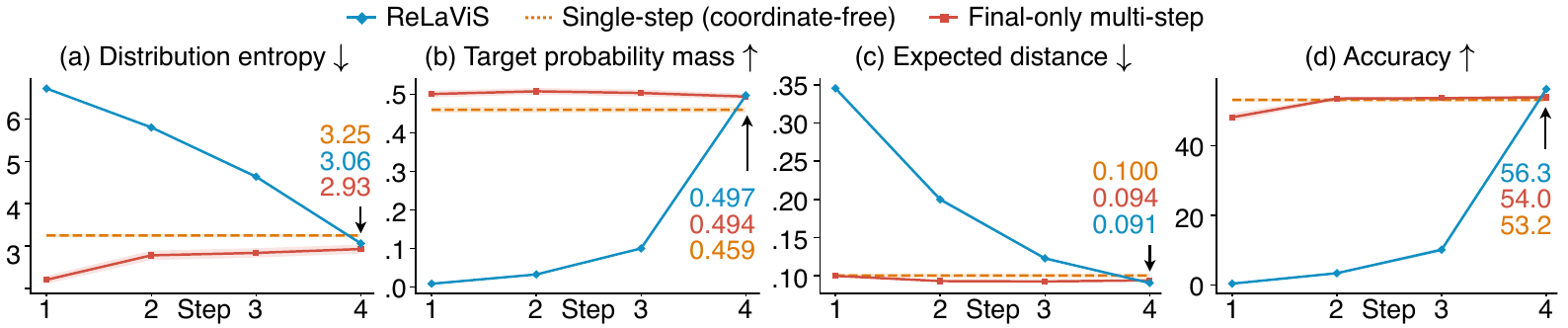}
    \caption{Stepwise search dynamics on ScreenSpot-Pro.
    Metric definitions are in Appendix~\ref{app:search-dynamics-metrics}.
    Dashed lines indicate single-step references; curves show multi-seed means with std shading.}
    \label{fig:recurrent-search-dynamics}
\end{figure}

Fig.~\ref{fig:recurrent-search-dynamics} compares the stepwise search dynamics of ReLaViS and two baselines from Table~\ref{tab:ablations} (group~RQ1, rows~1--2): the coordinate-free single-step baseline and the final-only multi-step variant, which receives supervision only at the final step.
Across four search steps, the entropy of the ReLaViS spatial search distribution decreases to 3.06, target probability mass rises from near zero to 0.497, and expected distance to the target element decreases from 0.345 to 0.091.
In contrast, the final-only variant starts with a highly concentrated search distribution, and its target probability mass and expected distance vary little across steps.
These coordinated trends show that ReLaViS progressively concentrates its spatial search toward the target before the final readout.

The predicted heatmaps in Figs.~\ref{fig:paradigm-comparison} and~\ref{fig:relavis-inference-pipeline} illustrate how ReLaViS progressively contracts its search focus from the global interface through local element groups (such as toolbars) to the target element; in contrast, the single-step variant incorrectly targets a non-interactive text element.
Visualizations and qualitative analyses in Appendix~\ref{app:visualizations-and-qualitative-analyses} illustrate how ReLaViS supports candidate disambiguation, semantic grounding, and fine-grained localization, and examine failure cases.
Stepwise diagnostics in Appendix~\ref{app:stepwise-search-state-diagnostics} further show that ReLaViS follows the GUI-aware coarse-to-fine trajectory more closely than the single-step and final-only variants.
With this GUI-aware search, ReLaViS reaches a better final spatial search state than these variants, with higher target probability mass, lower expected distance, and higher grounding accuracy (Fig.~\ref{fig:recurrent-search-dynamics}).

\section{Conclusion and limitations}
\label{sec:conclusion-and-limitations}

We introduce ReLaViS, a novel framework that formulates GUI grounding as explicit multi-step visual search within a single model interaction.
Guided by GUI-aware trajectories constructed from flat element annotations, its recurrent spatial search states carry visual evidence forward and progressively narrow the search toward the target without external tools.
Comprehensive experiments show that ReLaViS outperforms matched single-step baselines on all five benchmarks, with its gains on ScreenSpot-Pro highlighting the value of recurrent latent visual search for complex GUI grounding at modest inference overhead.
A current limitation is that ReLaViS mainly focuses on GUI grounding for click actions, leaving its extension to other actions, such as drag-and-drop gestures, for future work.
More broadly, extending ReLaViS to general visual search tasks remains promising.

\subsection*{AI use statement}

In this work, we use generative AI tools to assist with implementing methods.
We do not use generative AI tools to formulate or refine hypotheses, design or provide feedback on research methodology or experiments, or interpret results, and the rest of the required disclosure tasks are not applicable to this work.
Additionally, we use generative AI tools to assist with literature retrieval and discovery and to edit the manuscript to improve its readability and language.
We review all AI-assisted work.
In particular, we review and test AI-assisted code for correctness and verify the relevance of references identified with AI assistance.
We also review all AI-assisted manuscript edits to preserve technical meaning and accurately reflect our work.
We take responsibility for the final content of this work, including text, claims, or artifacts produced with the aid of generative AI.

\subsection*{Reproducibility statement}

We support reproducibility by documenting ReLaViS, the experimental protocol, and the supporting empirical evidence.
Secs.~\ref{sec:methodology} and~\ref{sec:experiments} present the core method and experimental setup.
Further information appears in Appendices~\ref{app:additional-details-of-relavis}, \ref{app:implementation-and-reproducibility-details}, and~\ref{app:additional-results-and-analyses}, which respectively provide the detailed task and method specification, data preparation and experimental configurations, and complete benchmark results and additional analyses.

\bibliography{iclr2027_conference}

@inproceedings{wu2026gui,
 author = {Wu, Qianhui and Cheng, Kanzhi and Yang, Rui and Zhang, Chaoyun and Yang, Jianwei and Jiang, Huiqiang and Mu, Jian and Peng, Baolin and Qiao, Bo and Tan, Reuben and Qin, Si and Liden, Lars and Lin, Qingwei and Zhang, Huan and Zhang, Tong and Zhang, Jianbing and Zhang, Dongmei and Gao, Jianfeng},
 booktitle = {Advances in Neural Information Processing Systems},
 doi = {10.52202/085713-0508},
 editor = {D. Belgrave and C. Zhang and H. Lin and R. Pascanu and P. Koniusz and M. Ghassemi and N. Chen},
 pages = {15101--15128},
 publisher = {Curran Associates, Inc.},
 title = {GUI-Actor: Coordinate-Free Visual Grounding for GUI Agents},
 url = {https://proceedings.neurips.cc/paper_files/paper/2025/file/16130af940e9dabb43c726119bd3b42e-Paper-Conference.pdf},
 volume = {38, Main Conference},
 year = {2025}
}

@misc{deng2025llm,
      title={LLM Latent Reasoning as Chain of Superposition}, 
      author={Jingcheng Deng and Liang Pang and Zihao Wei and Shicheng Xu and Zenghao Duan and Kun Xu and Yang Song and Huawei Shen and Xueqi Cheng},
      year={2026},
      eprint={2510.15522},
      archivePrefix={arXiv},
      primaryClass={cs.CL},
      url={https://arxiv.org/abs/2510.15522}, 
}

@misc{chen2025v2p,
      title={V2P: Visual Attention Calibration for GUI Grounding via Background Suppression and Center Peaking}, 
      author={Jikai Chen and Long Chen and Dong Wang and Qinglin Su and Zhixuan Chu and Bingguang Hao and Leilei Gan and Chenyi Zhuang and Jinjie Gu},
      year={2026},
      eprint={2508.13634},
      archivePrefix={arXiv},
      primaryClass={cs.AI},
      url={https://arxiv.org/abs/2508.13634}, 
}

@misc{bai2025qwen25vltechnicalreport,
      title={Qwen2.5-VL Technical Report}, 
      author={Shuai Bai and Keqin Chen and Xuejing Liu and Jialin Wang and Wenbin Ge and Sibo Song and Kai Dang and Peng Wang and Shijie Wang and Jun Tang and Humen Zhong and Yuanzhi Zhu and Mingkun Yang and Zhaohai Li and Jianqiang Wan and Pengfei Wang and Wei Ding and Zheren Fu and Yiheng Xu and Jiabo Ye and Xi Zhang and Tianbao Xie and Zesen Cheng and Hang Zhang and Zhibo Yang and Haiyang Xu and Junyang Lin},
      year={2025},
      eprint={2502.13923},
      archivePrefix={arXiv},
      primaryClass={cs.CV},
      url={https://arxiv.org/abs/2502.13923}, 
}

@misc{hu2021lora,
      title={LoRA: Low-Rank Adaptation of Large Language Models}, 
      author={Edward J. Hu and Yelong Shen and Phillip Wallis and Zeyuan Allen-Zhu and Yuanzhi Li and Shean Wang and Lu Wang and Weizhu Chen},
      year={2021},
      eprint={2106.09685},
      archivePrefix={arXiv},
      primaryClass={cs.CL},
      url={https://arxiv.org/abs/2106.09685}, 
}

@inproceedings{li2025screenspot,
author = {Li, Kaixin and Meng, Ziyang and Lin, Hongzhan and Luo, Ziyang and Tian, Yuchen and Ma, Jing and Huang, Zhiyong and Chua, Tat-Seng},
title = {ScreenSpot-Pro: GUI Grounding for Professional High-Resolution Computer Use},
year = {2025},
isbn = {9798400720352},
publisher = {Association for Computing Machinery},
address = {New York, NY, USA},
url = {https://doi.org/10.1145/3746027.3755688},
doi = {10.1145/3746027.3755688},
booktitle = {Proceedings of the 33rd ACM International Conference on Multimedia},
pages = {8778–8786},
numpages = {9},
location = {Dublin, Ireland},
series = {MM '25}
}

@InProceedings{luo2025visual,
    author    = {Luo, Tiange and Logeswaran, Lajanugen and Johnson, Justin and Lee, Honglak},
    title     = {Visual Test-time Scaling for GUI Agent Grounding},
    booktitle = {Proceedings of the IEEE/CVF International Conference on Computer Vision (ICCV)},
    month     = {October},
    year      = {2025},
    pages     = {19989-19998}
}

@inproceedings{li2026latent,
 author = {Li, Bangzheng and Sun, Ximeng and Liu, Jiang and Wang, Ze and Wu, Jialian and Yu, Xiaodong and Barsoum, Emad and Chen, Muhao and Liu, Zicheng},
 booktitle = {International Conference on Learning Representations},
 editor = {C. Vondrick and B. Hariharan and C. Raffel and L. Pinto and D. Yang and A. Faust},
 pages = {148076--148090},
 title = {Latent Visual Reasoning},
 url = {https://proceedings.iclr.cc/paper_files/paper/2026/file/efa0939caa8a906fd0a61a9c60e46e8b-Paper-Conference.pdf},
 volume = {2026},
 year = {2026}
}

@article{putkonen2025understanding,
title = {Understanding visual search in graphical user interfaces},
journal = {International Journal of Human-Computer Studies},
volume = {199},
pages = {103483},
year = {2025},
issn = {1071-5819},
doi = {https://doi.org/10.1016/j.ijhcs.2025.103483},
url = {https://www.sciencedirect.com/science/article/pii/S1071581925000400},
author = {Aini Putkonen and Yue Jiang and Jingchun Zeng and Olli Tammilehto and Jussi P.P. Jokinen and Antti Oulasvirta}
}

@misc{hurst2024gpt,
      title={GPT-4o System Card}, 
      author={{OpenAI}},
      year={2024},
      eprint={2410.21276},
      archivePrefix={arXiv},
      primaryClass={cs.CL},
      url={https://arxiv.org/abs/2410.21276}, 
}

@misc{hu2024dawn,
      title={The Dawn of GUI Agent: A Preliminary Case Study with Claude 3.5 Computer Use}, 
      author={Siyuan Hu and Mingyu Ouyang and Difei Gao and Mike Zheng Shou},
      year={2024},
      eprint={2411.10323},
      archivePrefix={arXiv},
      primaryClass={cs.AI},
      url={https://arxiv.org/abs/2411.10323}, 
}

@misc{bai2025qwen3,
      title={Qwen3-VL Technical Report}, 
      author={Shuai Bai and Yuxuan Cai and Ruizhe Chen and Keqin Chen and Xionghui Chen and Zesen Cheng and Lianghao Deng and Wei Ding and Chang Gao and Chunjiang Ge and Wenbin Ge and Zhifang Guo and Qidong Huang and Jie Huang and Fei Huang and Binyuan Hui and Shutong Jiang and Zhaohai Li and Mingsheng Li and Mei Li and Kaixin Li and Zicheng Lin and Junyang Lin and Xuejing Liu and Jiawei Liu and Chenglong Liu and Yang Liu and Dayiheng Liu and Shixuan Liu and Dunjie Lu and Ruilin Luo and Chenxu Lv and Rui Men and Lingchen Meng and Xuancheng Ren and Xingzhang Ren and Sibo Song and Yuchong Sun and Jun Tang and Jianhong Tu and Jianqiang Wan and Peng Wang and Pengfei Wang and Qiuyue Wang and Yuxuan Wang and Tianbao Xie and Yiheng Xu and Haiyang Xu and Jin Xu and Zhibo Yang and Mingkun Yang and Jianxin Yang and An Yang and Bowen Yu and Fei Zhang and Hang Zhang and Xi Zhang and Bo Zheng and Humen Zhong and Jingren Zhou and Fan Zhou and Jing Zhou and Yuanzhi Zhu and Ke Zhu},
      year={2025},
      eprint={2511.21631},
      archivePrefix={arXiv},
      primaryClass={cs.CV},
      url={https://arxiv.org/abs/2511.21631}, 
}

@inproceedings{xie2026scaling,
 author = {Xie, Tianbao and Deng, Jiaqi and Li, Xiaochuan and Yang, Junlin and Wu, Haoyuan and Chen, Jixuan and Hu, Wenjing and Wang, Xinyuan and Xu, Yuhui and Wang, Zekun and Xu, Yiheng and Wang, Junli and Sahoo, Doyen and Yu, Tao and Xiong, Caiming},
 booktitle = {Advances in Neural Information Processing Systems},
 doi = {10.52202/085713-0817},
 editor = {D. Belgrave and C. Zhang and H. Lin and R. Pascanu and P. Koniusz and M. Ghassemi and N. Chen},
 pages = {},
 publisher = {Curran Associates, Inc.},
 title = {Scaling Computer-Use Grounding via User Interface Decomposition and Synthesis},
 url = {https://proceedings.neurips.cc/paper_files/paper/2025/file/22c868099177ee278eb7baccec649f35-Paper-Datasets_and_Benchmarks_Track.pdf},
 volume = {38, Main Conference},
 year = {2025}
}

@misc{qin2025ui,
      title={UI-TARS: Pioneering Automated GUI Interaction with Native Agents}, 
      author={Yujia Qin and Yining Ye and Junjie Fang and Haoming Wang and Shihao Liang and Shizuo Tian and Junda Zhang and Jiahao Li and Yunxin Li and Shijue Huang and Wanjun Zhong and Kuanye Li and Jiale Yang and Yu Miao and Woyu Lin and Longxiang Liu and Xu Jiang and Qianli Ma and Jingyu Li and Xiaojun Xiao and Kai Cai and Chuang Li and Yaowei Zheng and Chaolin Jin and Chen Li and Xiao Zhou and Minchao Wang and Haoli Chen and Zhaojian Li and Haihua Yang and Haifeng Liu and Feng Lin and Tao Peng and Xin Liu and Guang Shi},
      year={2025},
      eprint={2501.12326},
      archivePrefix={arXiv},
      primaryClass={cs.AI},
      url={https://arxiv.org/abs/2501.12326}, 
}

@inproceedings{feizi2026grounding,
 author = {Feizi, Aarash and Nayak, Shravan and Jian, Xiangru and Lin, Kevin Qinghong and Li, Kaixin and Awal, Rabiul and Lu, Xing Han and Obando Ceron, Johan S and Rodriguez, Juan A. and Chapados, Nicolas and Vazquez, David and Romero-Soriano, Adriana and Rabbany, Reihaneh and Taslakian, Perouz and Pal, Christopher and Gella, Spandana and Mudumba, Sai Rajeswar},
 booktitle = {International Conference on Learning Representations},
 editor = {C. Vondrick and B. Hariharan and C. Raffel and L. Pinto and D. Yang and A. Faust},
 pages = {5181--5215},
 title = {Grounding Computer Use Agents on Human Demonstrations},
 url = {https://proceedings.iclr.cc/paper_files/paper/2026/file/0961fffb7adb81388b9b24ab6a3ff7ab-Paper-Conference.pdf},
 volume = {2026},
 year = {2026}
}

@inproceedings{yuan2026se,
 author = {Yuan, Xinbin and Zhang, Jian and Li, Kaixin and Cai, Zhuoxuan and Yao, Lujian and Chen, Jie and Wang, Enguang and Hou, Qibin and Chen, Jinwei and Jiang, Peng-Tao and Li, Bo},
 booktitle = {Advances in Neural Information Processing Systems},
 doi = {10.52202/085713-4251},
 editor = {D. Belgrave and C. Zhang and H. Lin and R. Pascanu and P. Koniusz and M. Ghassemi and N. Chen},
 pages = {127658--127679},
 publisher = {Curran Associates, Inc.},
 title = {SE-GUI: Enhancing Visual Grounding for GUI Agents via Self-Evolutionary Reinforcement Learning},
 url = {https://proceedings.neurips.cc/paper_files/paper/2025/file/b95c7e24501f5d1dddbc5e8526cda7ae-Paper-Conference.pdf},
 volume = {38, Main Conference},
 year = {2025}
}

@inproceedings{tang2026gui,
author = {Tang, Fei and Gu, Zhangxuan and Lu, Zhengxi and Liu, Xuyang and Shen, Shuheng and Meng, Changhua and Wang, Wen and Zhang, Wenqi and Shen, Yongliang and Lu, Weiming and Xiao, Jun and Zhuang, Yueting},
title = {GUI-G2: Gaussian reward modeling for GUI grounding},
year = {2026},
isbn = {978-1-57735-906-7},
publisher = {AAAI Press},
url = {https://doi.org/10.1609/aaai.v40i39.40606},
doi = {10.1609/aaai.v40i39.40606},
booktitle = {Proceedings of the Fortieth AAAI Conference on Artificial Intelligence and Thirty-Eighth Conference on Innovative Applications of Artificial Intelligence and Sixteenth Symposium on Educational Advances in Artificial Intelligence},
articleno = {3705},
numpages = {9},
series = {AAAI'26/IAAI'26/EAAI'26}
}

@inproceedings{liu2026infigui,
author = {Liu, Yuhang and Liu, Zeyu and Zhu, Shuanghe and Li, Pengxiang and Xie, Congkai and Wang, Jiasheng and Hu, Xueyu and Han, Xiaotian and Yuan, Jianbo and Wang, Xinyao and Zhang, Shengyu and Yang, Hongxia and Wu, Fei},
title = {InfiGUI-G1: advancing GUI grounding with adaptive exploration policy optimization},
year = {2026},
isbn = {978-1-57735-906-7},
publisher = {AAAI Press},
url = {https://doi.org/10.1609/aaai.v40i38.40500},
doi = {10.1609/aaai.v40i38.40500},
booktitle = {Proceedings of the Fortieth AAAI Conference on Artificial Intelligence and Thirty-Eighth Conference on Innovative Applications of Artificial Intelligence and Sixteenth Symposium on Educational Advances in Artificial Intelligence},
articleno = {3599},
numpages = {9},
series = {AAAI'26/IAAI'26/EAAI'26}
}

@inproceedings{chen2026ui,
 author = {Chen, Liangyu and Zhou, Hanzhang and cai, chenglin and Zhang, Jianan and Tong, Panrong and Zhang, Xu and Kong, Quyu and Liu, Chen and Liu, Yuqi and Wang, Wenxuan and Wang, Yue and Jin, Qin and HOI, Steven},
 booktitle = {International Conference on Learning Representations},
 editor = {C. Vondrick and B. Hariharan and C. Raffel and L. Pinto and D. Yang and A. Faust},
 pages = {54418--54444},
 title = {UI-Ins: Enhancing GUI Grounding with Multi-Perspective Instruction as Reasoning},
 url = {https://proceedings.iclr.cc/paper_files/paper/2026/file/5965f3a748a8d41415db2bfa44635cc3-Paper-Conference.pdf},
 volume = {2026},
 year = {2026}
}

@InProceedings{zhou2026gui,
author="Zhou, Shijie
and Lai, Viet Dac
and Tan, Hao
and Kil, Jihyung
and Zhu, Wanrong
and Chen, Changyou
and Zhang, Ruiyi",
editor="Favaro, Paolo
and Kukelova, Zuzana
and Maki, Atsuto
and Rohrbach, Anna
and Schindler, Konrad
and Tombari, Federico",
title="GUI-AIMA: Aligning Intrinsic Multimodal Attention with a Context Anchor for GUI Grounding",
booktitle="Computer Vision -- ECCV 2026",
year="2026",
publisher="Springer Nature Switzerland",
address="Cham",
pages="79--97",
isbn="978-3-032-37016-7"
}

@misc{lei2025gui,
      title={\textsc{GUI-Spotlight}: Adaptive Iterative Focus Refinement for Enhanced GUI Visual Grounding}, 
      author={Bin Lei and Nuo Xu and Ali Payani and Mingyi Hong and Chunhua Liao and Yu Cao and Caiwen Ding},
      year={2025},
      eprint={2510.04039},
      archivePrefix={arXiv},
      primaryClass={cs.CV},
      url={https://arxiv.org/abs/2510.04039}, 
}

@InProceedings{zhang2026mvp,
    author    = {Zhang, Yunzhu and Pan, Zeyu and Zeng, Zhengwen and Shen, Shuheng and Meng, Changhua and Zhu, Linchao},
    title     = {MVP: Multiple View Prediction Improves GUI Grounding},
    booktitle = {Proceedings of the IEEE/CVF Conference on Computer Vision and Pattern Recognition (CVPR)},
    month     = {June},
    year      = {2026},
    pages     = {27482-27492}
}

@InProceedings{wang2026monet,
    author    = {Wang, Qixun and Shi, Yang and Wang, Yifei and Zhang, Yuanxing and Wan, Pengfei and Gai, Kun and Ying, Xianghua and Wang, Yisen},
    title     = {Monet: Reasoning in Latent Visual Space Beyond Image and Language},
    booktitle = {Proceedings of the IEEE/CVF Conference on Computer Vision and Pattern Recognition (CVPR)},
    month     = {June},
    year      = {2026},
    pages     = {12030-12040}
}

@misc{zhang2026learn,
      title={Learn where to Click from Yourself: On-Policy Self-Distillation for GUI Grounding}, 
      author={Yan Zhang and Daiqing Wu and Huawen Shen and Can Ma and Yu Zhou},
      year={2026},
      eprint={2605.00642},
      archivePrefix={arXiv},
      primaryClass={cs.AI},
      url={https://arxiv.org/abs/2605.00642}, 
}

@InProceedings{yang2026machine,
    author    = {Yang, Zeyuan and Yu, Xueyang and Chen, Delin and Shen, Maohao and Gan, Chuang},
    title     = {Machine Mental Imagery: Empower Multimodal Reasoning with Latent Visual Tokens},
    booktitle = {Proceedings of the IEEE/CVF Conference on Computer Vision and Pattern Recognition (CVPR)},
    month     = {June},
    year      = {2026},
    pages     = {33510-33520}
}

@article{cua2025,
  title={Computer-Using Agent: Introducing a universal interface for AI to interact with the digital world},
  author={OpenAI},
  year={2025},
  url={https://openai.com/index/computer-using-agent},
}

@misc{anthropic2024computeruse,
  author = {{Anthropic}},
  title = {Developing a computer use model},
  year = {2024},
  month = oct,
  day = {22},
  url = {https://www.anthropic.com/news/developing-computer-use}
}

@misc{huang2026trust,
      title={Trust the Right Teacher: Quality-Aware Self-Distillation for GUI Grounding}, 
      author={Jingyuan Huang and Zuming Huang and Yucheng Shi and Tianze Yang and Xiaoming Zhai and Wei Chu and Ninghao Liu},
      year={2026},
      eprint={2606.18101},
      archivePrefix={arXiv},
      primaryClass={cs.AI},
      url={https://arxiv.org/abs/2606.18101}, 
}

@inproceedings{shao2024visual,
 author = {Shao, Hao and Qian, Shengju and Xiao, Han and Song, Guanglu and Zong, Zhuofan and Wang, Letian and Liu, Yu and Li, Hongsheng},
 booktitle = {Advances in Neural Information Processing Systems},
 doi = {10.52202/079017-0275},
 editor = {A. Globerson and L. Mackey and D. Belgrave and A. Fan and U. Paquet and J. Tomczak and C. Zhang},
 pages = {8612--8642},
 publisher = {Curran Associates, Inc.},
 title = {Visual CoT: Advancing Multi-Modal Language Models with a Comprehensive Dataset and Benchmark for Chain-of-Thought Reasoning},
 url = {https://proceedings.neurips.cc/paper_files/paper/2024/file/0ff38d72a2e0aa6dbe42de83a17b2223-Paper-Datasets_and_Benchmarks_Track.pdf},
 volume = {37},
 year = {2024}
}

@inproceedings{tan2026reason,
 author = {Tan, Huajie and Ji, Yuheng and Hao, Xiaoshuai and Chen, Xiansheng and Wang, Pengwei and Wang, Zhongyuan and Zhang, Shanghang},
 booktitle = {Advances in Neural Information Processing Systems},
 doi = {10.52202/085713-0204},
 editor = {D. Belgrave and C. Zhang and H. Lin and R. Pascanu and P. Koniusz and M. Ghassemi and N. Chen},
 pages = {5772--5822},
 publisher = {Curran Associates, Inc.},
 title = {Reason-RFT: Reinforcement Fine-Tuning for Visual Reasoning of Vision Language Models},
 url = {https://proceedings.neurips.cc/paper_files/paper/2025/file/08d70284b013c03ba89cd2b642bc864b-Paper-Conference.pdf},
 volume = {38, Main Conference},
 year = {2025}
}

@misc{hao2024training,
      title={Training Large Language Models to Reason in a Continuous Latent Space}, 
      author={Shibo Hao and Sainbayar Sukhbaatar and DiJia Su and Xian Li and Zhiting Hu and Jason Weston and Yuandong Tian},
      year={2026},
      eprint={2412.06769},
      archivePrefix={arXiv},
      primaryClass={cs.CL},
      url={https://arxiv.org/abs/2412.06769}, 
}

@misc{tang2026ui,
      title={UI-Zoomer: Uncertainty-Driven Adaptive Zoom-In for GUI Grounding}, 
      author={Fei Tang and Bofan Chen and Zhengxi Lu and Tongbo Chen and Songqin Nong and Tao Jiang and Wenhao Xu and Weiming Lu and Jun Xiao and Yueting Zhuang and Yongliang Shen},
      year={2026},
      eprint={2604.14113},
      archivePrefix={arXiv},
      primaryClass={cs.CV},
      url={https://arxiv.org/abs/2604.14113}, 
}

@misc{cheng2024compressed,
      title={Compressed Chain of Thought: Efficient Reasoning Through Dense Representations}, 
      author={Jeffrey Cheng and Benjamin Van Durme},
      year={2024},
      eprint={2412.13171},
      archivePrefix={arXiv},
      primaryClass={cs.CL},
      url={https://arxiv.org/abs/2412.13171}, 
}

@InProceedings{qin2025chain,
author="Qin, Yiming
and Wei, Bomin
and Ge, Jiaxin
and Kallidromitis, Konstantinos
and Fu, Stephanie
and Darrell, Trevor
and Wang, XuDong",
editor="Favaro, Paolo
and Kukelova, Zuzana
and Maki, Atsuto
and Rohrbach, Anna
and Schindler, Konrad
and Tombari, Federico",
title="Chain-of-Visual-Thought: Teaching VLMs to See and Think Better with Continuous Visual Tokens",
booktitle="Computer Vision -- ECCV 2026",
year="2026",
publisher="Springer Nature Switzerland",
address="Cham",
pages="445--462",
isbn="978-3-032-37359-5"
}

@misc{wang2024mobile,
      title={Mobile-Agent: Autonomous Multi-Modal Mobile Device Agent with Visual Perception}, 
      author={Junyang Wang and Haiyang Xu and Jiabo Ye and Ming Yan and Weizhou Shen and Ji Zhang and Fei Huang and Jitao Sang},
      year={2024},
      eprint={2401.16158},
      archivePrefix={arXiv},
      primaryClass={cs.CL},
      url={https://arxiv.org/abs/2401.16158}, 
}

@inproceedings{yang2025aria,
    title = "Aria-{UI}: Visual Grounding for {GUI} Instructions",
    author = "Yang, Yuhao  and
      Wang, Yue  and
      Li, Dongxu  and
      Luo, Ziyang  and
      Chen, Bei  and
      Huang, Chao  and
      Li, Junnan",
    editor = "Che, Wanxiang  and
      Nabende, Joyce  and
      Shutova, Ekaterina  and
      Pilehvar, Mohammad Taher",
    booktitle = "Findings of the Association for Computational Linguistics: ACL 2025",
    month = jul,
    year = "2025",
    address = "Vienna, Austria",
    publisher = "Association for Computational Linguistics",
    url = "https://aclanthology.org/2025.findings-acl.1152/",
    doi = "10.18653/v1/2025.findings-acl.1152",
    pages = "22418--22433",
    ISBN = "979-8-89176-256-5"
}

@InProceedings{lin2025showui,
    author    = {Lin, Kevin Qinghong and Li, Linjie and Gao, Difei and Yang, Zhengyuan and Wu, Shiwei and Bai, Zechen and Lei, Stan Weixian and Wang, Lijuan and Shou, Mike Zheng},
    title     = {ShowUI: One Vision-Language-Action Model for GUI Visual Agent},
    booktitle = {Proceedings of the IEEE/CVF Conference on Computer Vision and Pattern Recognition (CVPR)},
    month     = {June},
    year      = {2025},
    pages     = {19498-19508}
}

@InProceedings{wang2025mmbench,
    author    = {Wang, Xuehui and Wu, Zhenyu and Xie, JingJing and Ding, Zichen and Yang, Bowen and Li, Zehao and Liu, Zhaoyang and Li, Qingyun and Dong, Xuan and Chen, Zhe and Wang, Weiyun and Zhao, Xiangyu and Chen, Jixuan and Duan, Haodong and Xie, Tianbao and Yang, Chenyu and Su, Shiqian and Yu, Yue and Zhang, Yanting and Yue, Xiangyu and Su, Weijie and Zhu, Xizhou and Shen, Wei and Dai, Jifeng and Wang, Wenhai},
    title     = {MMBench-GUI: A Unified Hierarchical Evaluation Framework for Multi-Platform GUI Agents},
    booktitle = {Proceedings of the IEEE/CVF Conference on Computer Vision and Pattern Recognition (CVPR)},
    month     = {June},
    year      = {2026},
    pages     = {6239-6248}
}

@inproceedings{wu2025atlas,
 author = {Wu, Zhiyong and Wu, Zhenyu and Xu, Fangzhi and Wang, Yian and Sun, Qiushi and Jia, Chengyou and Cheng, Kanzhi and Ding, Zichen and Chen, Liheng and Liang, Paul Pu and Qiao, Yu},
 booktitle = {International Conference on Learning Representations},
 editor = {Y. Yue and A. Garg and N. Peng and F. Sha and R. Yu},
 pages = {5090--5108},
 title = {OS-ATLAS: Foundation Action Model for Generalist GUI Agents},
 url = {https://proceedings.iclr.cc/paper_files/paper/2025/file/0faa4bc5f522076947a030273629d4fe-Paper-Conference.pdf},
 volume = {2025},
 year = {2025}
}

@InProceedings{nayak2025ui,
  title = 	 {{UI}-Vision: A Desktop-centric {GUI} Benchmark for Visual Perception and Interaction},
  author =       {Nayak, Shravan and Jian, Xiangru and Lin, Kevin Qinghong and Rodriguez, Juan A. and Kalsi, Montek and Chapados, Nicolas and \"{O}zsu, M. Tamer and Agrawal, Aishwarya and Vazquez, David and Pal, Christopher and Taslakian, Perouz and Gella, Spandana and Rajeswar, Sai},
  booktitle = 	 {Proceedings of the 42nd International Conference on Machine Learning},
  pages = 	 {45817--45851},
  year = 	 {2025},
  editor = 	 {Singh, Aarti and Fazel, Maryam and Hsu, Daniel and Lacoste-Julien, Simon and Berkenkamp, Felix and Maharaj, Tegan and Wagstaff, Kiri and Zhu, Jerry},
  volume = 	 {267},
  series = 	 {Proceedings of Machine Learning Research},
  month = 	 {13--19 Jul},
  publisher =    {PMLR},
  url = 	 {https://proceedings.mlr.press/v267/nayak25a.html}
}

@inproceedings{li2024effects,
 author = {Li, Wei and Bishop, William and Li, Alice and Rawles, Chris and Campbell-Ajala, Folawiyo and Tyamagundlu, Divya and Riva, Oriana},
 booktitle = {Advances in Neural Information Processing Systems},
 doi = {10.52202/079017-2925},
 editor = {A. Globerson and L. Mackey and D. Belgrave and A. Fan and U. Paquet and J. Tomczak and C. Zhang},
 pages = {92130--92154},
 publisher = {Curran Associates, Inc.},
 title = {On the Effects of Data Scale on UI Control Agents},
 url = {https://proceedings.neurips.cc/paper_files/paper/2024/file/a79f3ef3b445fd4659f44648f7ea8ffd-Paper-Datasets_and_Benchmarks_Track.pdf},
 volume = {37},
 year = {2024}
}

@misc{gurbuz2026moving,
      title={ScreenParse: Moving Beyond Sparse Grounding with Complete Screen Parsing Supervision}, 
      author={A. Said Gurbuz and Sunghwan Hong and Ahmed Nassar and Marc Pollefeys and Peter Staar},
      year={2026},
      eprint={2602.14276},
      archivePrefix={arXiv},
      primaryClass={cs.CV},
      url={https://arxiv.org/abs/2602.14276}, 
}

@InProceedings{liu2026drs,
    author    = {Liu, Yichao and Shen, Huawen and Yu, Liu and Liu, Shiyu and Chen, Zeyu and Zhou, Yu},
    title     = {DRS-GUI: Dynamic Region Search for Training-Free GUI Grounding},
    booktitle = {Proceedings of the IEEE/CVF Conference on Computer Vision and Pattern Recognition (CVPR)},
    month     = {June},
    year      = {2026},
    pages     = {34606-34616}
}

@inproceedings{chen2026gui,
author = {Chen, Chen and Shao, Jiawei and Lu, Dakuan and Hu, Haoyi and Liu, Xiangcheng and Yao, Hantao and Liu, Wu},
title = {GUI-eyes: tool-augmented perception for visual grounding in GUI agents},
year = {2026},
isbn = {978-1-57735-906-7},
publisher = {AAAI Press},
url = {https://doi.org/10.1609/aaai.v40i35.40175},
doi = {10.1609/aaai.v40i35.40175},
booktitle = {Proceedings of the Fortieth AAAI Conference on Artificial Intelligence and Thirty-Eighth Conference on Innovative Applications of Artificial Intelligence and Sixteenth Symposium on Educational Advances in Artificial Intelligence},
articleno = {3274},
numpages = {9},
series = {AAAI'26/IAAI'26/EAAI'26}
}

@misc{zhao2025learning,
      title={Learning GUI Grounding with Spatial Reasoning from Visual Feedback}, 
      author={Yu Zhao and Wei-Ning Chen and Huseyin Atahan Inan and Samuel Kessler and Lu Wang and Lukas Wutschitz and Fangkai Yang and Chaoyun Zhang and Pasquale Minervini and Saravan Rajmohan and Robert Sim},
      year={2026},
      eprint={2509.21552},
      archivePrefix={arXiv},
      primaryClass={cs.CV},
      url={https://arxiv.org/abs/2509.21552}, 
}

@misc{li2025chain,
      title={Chain-of-Ground: Improving GUI Grounding via Iterative Reasoning and Reference Feedback}, 
      author={Aiden Yiliu Li and Bizhi Yu and Daoan Lei and Tianhe Ren and Shilong Liu},
      year={2025},
      eprint={2512.01979},
      archivePrefix={arXiv},
      primaryClass={cs.AI},
      url={https://arxiv.org/abs/2512.01979}, 
}

@misc{li2026guic2coarsetofineguigrounding,
      title={GUI-C$^2$: Coarse-to-Fine GUI Grounding via Difficulty-Aware Reinforcement Learning}, 
      author={Junlong Li and Chao Hao and Lap-Pui Chau and Yi Wang},
      year={2026},
      eprint={2605.30884},
      archivePrefix={arXiv},
      primaryClass={cs.CV},
      url={https://arxiv.org/abs/2605.30884}, 
}

@inproceedings{wang2026measure,
    title = "Measure Twice, Click Once: Co-evolving Proposer and Visual Critic via Reinforcement Learning for {GUI} Grounding",
    author = "Wang, Wenkai  and
      Li, Xiyun  and
      Guo, Hongcan  and
      Yu, Wenhao  and
      Fang, Tianqing  and
      Mi, Haitao  and
      Yu, Dong  and
      Zhang, Shengyu",
    editor = "Liakata, Maria  and
      Moreira, Viviane P.  and
      Zhang, Jiajun  and
      Jurgens, David",
    booktitle = "Proceedings of the 64th Annual Meeting of the {A}ssociation for {C}omputational {L}inguistics (Volume 1: Long Papers)",
    month = jul,
    year = "2026",
    address = "San Diego, California, United States",
    publisher = "Association for Computational Linguistics",
    url = "https://aclanthology.org/2026.acl-long.960/",
    doi = "10.18653/v1/2026.acl-long.960",
    pages = "20964--20984",
    ISBN = "979-8-89176-390-6"
}

@misc{yao2026autofocus,
      title={AutoFocus: Uncertainty-Aware Active Visual Search for GUI Grounding}, 
      author={Ruilin Yao and Shegnwu Xiong and Tianyu Zou and Shili Xiong and Yi Rong},
      year={2026},
      eprint={2605.02630},
      archivePrefix={arXiv},
      primaryClass={cs.CV},
      url={https://arxiv.org/abs/2605.02630}, 
}

@misc{muryn2025screen2ax,
      title={Screen2AX: Vision-Based Approach for Automatic macOS Accessibility Generation}, 
      author={Viktor Muryn and Marta Sumyk and Mariya Hirna and Sofiya Garkot and Maksym Shamrai},
      year={2025},
      eprint={2507.16704},
      archivePrefix={arXiv},
      primaryClass={cs.LG},
      url={https://arxiv.org/abs/2507.16704}, 
}

@inproceedings{wu2025dimo,
    title = "{D}i{M}o-{GUI}: Advancing Test-time Scaling in {GUI} Grounding via Modality-Aware Visual Reasoning",
    author = "Wu, Hang  and
      Chen, Hongkai  and
      Cai, Yujun  and
      Liu, Chang  and
      Ye, Qingwen  and
      Yang, Ming-Hsuan  and
      Wang, Yiwei",
    editor = "Christodoulopoulos, Christos  and
      Chakraborty, Tanmoy  and
      Rose, Carolyn  and
      Peng, Violet",
    booktitle = "Proceedings of the 2025 Conference on Empirical Methods in Natural Language Processing",
    month = nov,
    year = "2025",
    address = "Suzhou, China",
    publisher = "Association for Computational Linguistics",
    url = "https://aclanthology.org/2025.emnlp-main.1334/",
    doi = "10.18653/v1/2025.emnlp-main.1334",
    pages = "26246--26256",
    ISBN = "979-8-89176-332-6"
}

@inproceedings{park2025r,
    title = "{R}-{VLM}: Region-Aware Vision Language Model for Precise {GUI} Grounding",
    author = "Park, Joonhyung  and
      Tang, Peng  and
      Das, Sagnik  and
      Appalaraju, Srikar  and
      Singh, Kunwar Yashraj  and
      Manmatha, R.  and
      Ghadar, Shabnam",
    editor = "Che, Wanxiang  and
      Nabende, Joyce  and
      Shutova, Ekaterina  and
      Pilehvar, Mohammad Taher",
    booktitle = "Findings of the Association for Computational Linguistics: ACL 2025",
    month = jul,
    year = "2025",
    address = "Vienna, Austria",
    publisher = "Association for Computational Linguistics",
    url = "https://aclanthology.org/2025.findings-acl.501/",
    doi = "10.18653/v1/2025.findings-acl.501",
    pages = "9669--9685",
    ISBN = "979-8-89176-256-5"
}
\bibliographystyle{iclr2027_conference}

\clearpage
\appendix
\addtocontents{toc}{\protect\setcounter{tocdepth}{2}}
\renewcommand{\contentsname}{Appendix contents}
\tableofcontents

\section{Additional details of ReLaViS}
\label{app:additional-details-of-relavis}

\subsection{Recurrent latent visual search}
\label{app:recurrent-latent-visual-search}

\textbf{Task template.}
\label{app:task-template}
We follow the tool-call template used in the official Qwen2.5-VL~\citep{bai2025qwen25vltechnicalreport} implementation for GUI interaction.
Our system prompt exposes a \texttt{computer\_use} tool whose \texttt{left\_click} action identifies the click target through an LVS span in place of textual coordinates:
\begin{Verbatim}[fontsize=\scriptsize,breaklines=true,breakanywhere=true]
You are a helpful assistant. The user will give you an instruction, and you MUST left click on the corresponding UI element via tool call. If you are not sure about where to click, guess a most likely one.

# Tools

You may call one or more functions to assist with the user query.

You are provided with function signatures within <tools></tools> XML tags:
<tools>
{"type": "function", "function": {"name": "computer_use", "description": "Use a mouse to interact with a computer.
* Make sure to click any buttons, links, icons, etc with the cursor tip in the center of the element.
* You can only use the left_click action to interact with the computer.", "parameters": {"properties": {"action": {"description": "The action to perform. The available actions are:
* `left_click`: Click the left mouse button on the UI element specified by lvr tokens.", "enum": ["left_click"], "type": "string"}, "lvr": {"description": "<|object_ref_start|><|lvr|><|object_ref_end|>: Latent visual reasoning tokens identifying the UI element to left click. Required only by `action=left_click`.", "type": "special_token"}, "required": ["action"], "type": "object"}}}}
</tools>

For each function call, return a json object with function name and arguments within <tool_call></tool_call> XML tags:
<tool_call>
{"name": <function-name>, "arguments": <args-json-object>}
</tool_call>
\end{Verbatim}
The corresponding tool-call response is as follows:
\begin{Verbatim}[fontsize=\scriptsize,breaklines=true,breakanywhere=true]
<tool_call>
{"name": "computer_use", "arguments": {"action": "left_click", "lvr": <|object_ref_start|><|lvr|><|lvr|><|lvr|><|lvr|><|object_ref_end|>}}
</tool_call>
\end{Verbatim}
The exact tokens \texttt{<|object\_ref\_start|>}, \texttt{<|lvr|>}, and \texttt{<|object\_ref\_end|>} correspond to the semantic symbols \texttt{LVS-START}, \texttt{LVS-SLOT}, and \texttt{LVS-END}, respectively.
The boundary tokens \texttt{<|object\_ref\_start|>} and \texttt{<|object\_ref\_end|>} already belong to the pretrained vocabulary, whereas we initialize \texttt{<|lvr|>} with the mean of the pretrained vocabulary embeddings.
We exclude every \texttt{LVS-SLOT} and \texttt{LVS-END} position from the next-token prediction loss, while retaining supervision on \texttt{LVS-START} and the textual tool-call prefix and suffix.

\textbf{Visual token grid.}
\label{app:grid-of-visual-tokens}
We use the default dynamic-resolution image preprocessing of Qwen2.5-VL, which resizes each screenshot while preserving its aspect ratio.
The visual encoder partitions the resized screenshot into $14\times14$ pixel patches and initially produces a grid of size $G_h\times G_w$.
The multimodal merger applies a spatial merge factor $M=2$ to each $2\times2$ group of patch embeddings.
The post-merge grid shared by the LLM backbone and the spatial search head therefore has dimensions $H_v=G_h/M$ and $W_v=G_w/M$, yielding $N=H_vW_v$ visual tokens, denoted as $V$.
Each post-merge visual token corresponds to a $28\times28$ pixel region in the resized screenshot, and we constrain the resized image area to the range from $4\times28^2$ to $16{,}384\times28^2$, yielding 4 to 16,384 tokens per screenshot.

\textbf{Spatial search head architecture.}
\label{app:spatial-search-head-architecture}
The $\operatorname{SelfAttn}$ block in Eq.~(\ref{eq:spatial-search-head}) uses 8-head self-attention, a dropout rate of $0.1$, a residual connection, and LayerNorm.
The hidden state and visual token $\operatorname{MLP}$s use independent \texttt{Linear-GELU-Linear} projections.
Their intermediate width $r$ equals the backbone hidden width $d$.
The head contains $8d^2+10d$ parameters (102.8M with Qwen2.5-VL-7B), introducing only a lightweight parameter overhead.

\subsection{GUI-aware search trajectory construction}
\label{app:gui-aware-search-trajectory-construction}

\begin{figure}[t]
    \centering
    \includegraphics[
        width=\linewidth,
        trim={0.95bp 2.62bp 1.44bp 3.21bp},
        clip
    ]{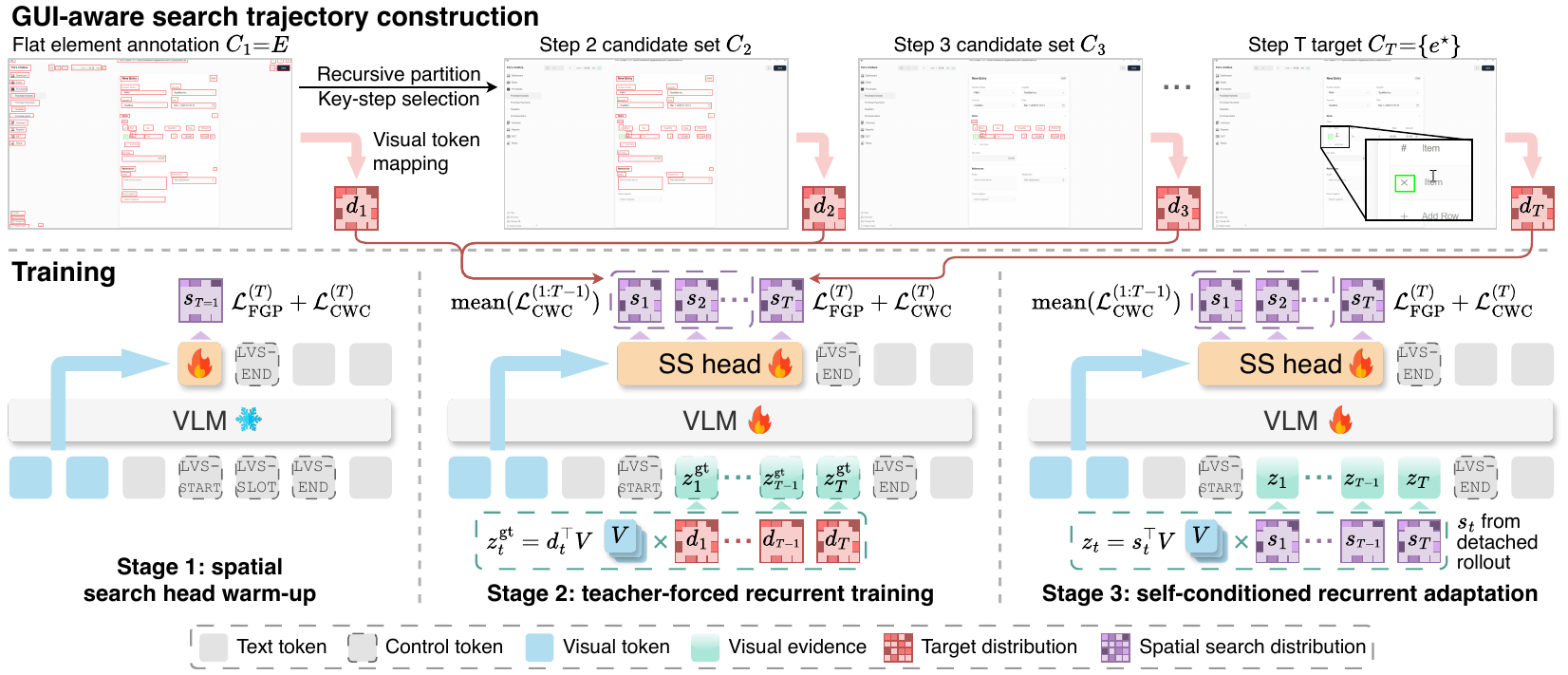}
    \caption{Overview of trajectory construction and training.
    \emph{Top}: ReLaViS converts flat GUI element annotations $E$ into a search trajectory $C_{1:T}$ and maps each candidate set $C_t$ to a target distribution $d_t$ over the visual token grid.
    \emph{Bottom}: Stage~1 trains only the spatial search head with a frozen VLM and $T=1$; Stage~2 uses teacher-forced inputs $z_t^{\mathrm{gt}}=d_t^\top V$; and Stage~3 uses self-conditioned inputs $z_t=s_t^\top V$ from a detached rollout.
    $\mathcal L_{\mathrm{CWC}}$ supervises all search steps, while $\mathcal L_{\mathrm{FGP}}$ additionally supervises the final step.
    ``SS head'' denotes the spatial search head.}
    \label{fig:method-overview}
\end{figure}

Fig.~\ref{fig:method-overview} summarizes GUI-aware search trajectory construction and the training procedure.

\begin{algorithm}[t]
\caption{Constructing the target-containing recursive partition.}
\label{alg:target-containing-recursive-partition}
\begin{algorithmic}[1]
\Require Full element set $E$ and target element $e^\star\in E$
\Ensure Raw partition path $(R_0,\ldots,R_L)$ with $R_0=E$ and $R_L=\{e^\star\}$
\State $R_0\gets E$; $L\gets 0$
\While{$|R_L|>1$}
    \State $\Pi(R_L)\gets\emptyset$
    \ForAll{$\alpha\in\{x,y\}$}
        \State $(e_1,\ldots,e_n)\gets
        \operatorname{sort}(R_L,\operatorname{Center}_\alpha)$
        \For{$p=1,\ldots,n-1$}
            \State Add $(\{e_1,\ldots,e_p\},\{e_{p+1},\ldots,e_n\})$ to $\Pi(R_L)$
        \EndFor
    \EndFor
    \State $(C_a^\star,C_b^\star)\gets\displaystyle\argmin_{(C_a,C_b)\in\Pi(R_L)}
    \left[\operatorname{Area}(B(C_a))+\operatorname{Area}(B(C_b))\right]$
    \State $R_{L+1}\gets C_a^\star$ if $e^\star\in C_a^\star$, and $R_{L+1}\gets C_b^\star$ otherwise
    \State $L\gets L+1$
\EndWhile
\State \Return $(R_0,\ldots,R_L)$
\end{algorithmic}
\end{algorithm}

\textbf{Target-containing recursive partition.}
\label{app:target-containing-recursive-partition}
Alg.~\ref{alg:target-containing-recursive-partition} starts from the full element set $E$, with $R_0=E$, and recursively narrows the current element set $R_i$ toward $e^\star$.
To keep the partition search tractable, we restrict $\Pi(R_i)$ to bipartitions induced by cuts between adjacent elements sorted by horizontal or vertical center coordinates.
Then the algorithm selects $(C_a^\star,C_b^\star)$ using the objective in Eq.~(\ref{eq:gui-partition}) and retains the subset containing $e^\star$ as $R_{i+1}$.
Ties in center coordinates are broken by original annotation order; for equal-area partitions, the algorithm favors the $x$-axis, then the first cut in sorted order.
The recursion terminates at $R_L=\{e^\star\}$, producing the nested raw partition path $(R_0,\ldots,R_L)$.

\begin{algorithm}[t]
\caption{Selecting key steps and constructing a fixed-step trajectory.}
\label{alg:key-step-selection-and-fixed-step-trajectory}
\begin{algorithmic}[1]
\Require Raw partition path $(R_0,\ldots,R_L)$, number of steps $T\geq2$, and local-reduction threshold~$k$
\Ensure Fixed-step trajectory $P^\star=(C_1,\ldots,C_T)$, $C_1=R_0=E$, and $C_T=R_L=\{e^\star\}$
\State $\mathcal{R}\gets\{1,0\}\cup
\left\{\frac{|R_i|-|R_j|}{|R_i|}:0\leq i<j<L\right\}$
\State $P^\star\gets()$
\ForAll{$\rho\in\operatorname{sort}_{\downarrow}(\mathcal{R})$}
    \State $P\gets(R_0)$; $i\gets0$
    \State Define $\mathcal{J}(i,\rho)=\left\{j:i<j<L,\ \frac{|R_i|-|R_j|}{|R_i|}\geq\rho,\ |R_j|-|R_{j+1}|>k\right\}$
    \While{$\mathcal{J}(i,\rho)\neq\emptyset$}
        \State $j\gets\min\mathcal{J}(i,\rho)$
        \State Append $R_j$ to $P$; $i\gets j$
    \EndWhile
    \If{$\operatorname{last}(P)\neq R_L$}
        \State Append $R_L$ to $P$
    \EndIf
    \If{$|P|\geq T$}
        \State $P^\star\gets(R_0)\mathbin{\|}\operatorname{suffix}_{T-1}(P)$
        \State \Return $P^\star$
    \EndIf
    \If{$|P|>|P^\star|$}
        \State $P^\star\gets P$
    \EndIf
\EndFor
\While{$|P^\star|<T$}
    \State Append $R_L$ to $P^\star$
\EndWhile
\State \Return $P^\star$
\end{algorithmic}
\end{algorithm}

\textbf{Key-step selection and fixed-step trajectory.}
\label{app:key-step-selection-and-fixed-step-trajectory}
Alg.~\ref{alg:key-step-selection-and-fixed-step-trajectory} removes redundant steps from the partition path and converts the result into a fixed-step trajectory.
The cumulative-reduction condition $({|R_i|-|R_j|})/{|R_i|}\geq\rho$ requires sufficient reduction from the last retained step, while the local-reduction condition $|R_j|-|R_{j+1}|>k$ retains only steps followed by a substantial reduction in candidate count.
For each $\rho\in\mathcal{R}$ in descending order, the algorithm constructs a new path $P$ from $R_0$ by repeatedly selecting the earliest subsequent intermediate step satisfying both conditions.
When no further intermediate step qualifies, it appends $R_L$.
At the first threshold yielding $|P|\geq T$, it returns $P$ unchanged if $|P|=T$; otherwise, it truncates $P$ to $R_0$ and its last $T-1$ steps.
If all paths are shorter than $T$, it pads the first longest path with copies of $R_L$ to length $T$.
The returned $P^\star=(C_1,\ldots, C_T)$ is the $T$-step trajectory used in the main text.

\textbf{Visual token mapping.}
\label{app:visual-token-mapping}
Let $b(v_i)$ be the rectangular patch corresponding to post-merge visual token $v_i$.
For an element $e$, the corresponding visual token support is
\begin{equation}
    \Omega(e)=\{v_i\in V:\operatorname{Area}(b(v_i)\cap b(e))>0\}.
    \label{eq:token-support}
\end{equation}
The support includes every visual token whose patch has a positive-area intersection with the element bounding box.
If $\Omega(e)$ is empty, we add the token whose patch contains the bounding box center, ensuring that every candidate element $e$ has nonempty support $\Omega(e)$.

\subsection{Search process supervision}
\label{app:search-process-supervision}

\textbf{Gaussian target distributions.}
\label{app:gaussian-target-distributions}
Following V2P~\citep{chen2025v2p}, we construct the target distribution $d_t$ using size-adaptive 2D Gaussians.
For each element, we set $\mu_{x,e}=(x_{0,e}+x_{1,e})/2$, $\sigma_{x,e}=x_{1,e}-x_{0,e}$, $\mu_{y,e}=(y_{0,e}+y_{1,e})/2$, and $\sigma_{y,e}=y_{1,e}-y_{0,e}$.
Let $\mathcal G(\cdot;\mu,\sigma)$ denote the density of a
one-dimensional Gaussian distribution with mean $\mu$ and standard
deviation $\sigma$.
At step $t$, we integrate each candidate's Gaussian density over $b(v_i)\cap b(e)$, take the maximum across $e\in C_t$, and normalize over visual tokens:
\begin{equation}
    \widetilde d_{t,i}
    =\max_{e\in C_t}
    \int_{(x,y)\in b(v_i)\cap b(e)}
    \mathcal G(x;\mu_{x,e},\sigma_{x,e})
    \mathcal G(y;\mu_{y,e},\sigma_{y,e})
    \,dx\,dy,
    \quad
    d_{t,i}
    =\frac{\widetilde d_{t,i}}
    {\sum_j\widetilde d_{t,j}}.
    \label{eq:target-distribution}
\end{equation}
The intersection truncates each density at its element boundary, and the maximum selects the largest contribution across candidates to define $\widetilde d_{t, i}$ for each visual token.

\textbf{Loss weighting.}
\label{app:loss-weighting}
Because the click coordinate directly depends on $s_T$, we assign equal weights to $\mathcal L_{\mathrm{CWC}}^{(T)}$ and the mean CWC loss over intermediate steps.
The FGP weight controls the contribution of final peak shaping relative to CWC supervision.
Validation-set selection yields $\lambda_{\mathrm{CWC}}^{(T)}=\lambda_{\mathrm{CWC}}^{(1:T-1)}=5$ and $\lambda_{\mathrm{FGP}}=1$.
As shown in the ablation of loss weighting (Appendix~\ref{app:intermediate-cwc-loss-weight}), the tested lower and higher intermediate CWC weights both reduce grounding accuracy relative to this setting.

\subsection{Training and inference}
\label{app:training-and-inference}

\textbf{Training procedure.}
\label{app:training-procedure}
In our implementation, we initialize the full response sequence with $T$ \texttt{LVS-SLOT} tokens.
Since $s_T$ does not depend on $z_T$, we inject latent visual evidence only into the first $T-1$ slots, saving one no-gradient backbone forward pass in Stage~3.
Stage~2 constructs the recurrent inputs directly from $d_{1:T-1}$ and $V$.
Stage~3 fills the slots sequentially: for each $t<T$, a backbone probe and the spatial search head predict $s_t$ without gradients using the previously filled slots, and the resulting $z_t$ replaces the $t$-th slot embedding.
The probe predictions are detached, while the aggregation $z_t=s_t^\top V$ remains differentiable with respect to $V$.
After all inputs are prepared, one differentiable forward pass over the full sequence computes all $T$ search distributions and the applicable losses in Eq.~(\ref{eq:overall-objective}).

\textbf{Forward-pass cost.}
\label{app:forward-pass-cost}
We count LLM backbone forward passes, with visual tokens $V$ computed once and reused during both training and inference.
Stages~1 and~2 require one differentiable pass each, since their inputs require no backbone probes.
Stage~3 requires $T-1$ no-gradient probes plus one differentiable pass, for $T$ passes in total.
Inference requires $T$ backbone passes for search, one per predicted $s_t$, plus additional passes for autoregressive decoding of the tool-call prefix and suffix.
Appendix~\ref{app:training-and-inference-cost} reports training time, inference latency, FLOPs, and memory.

\textbf{Coordinate readout.}
\label{app:coordinate-readout}
$\mathcal L_{\mathrm{FGP}}^{(T)}$ encourages $s_T$ to concentrate around the target center, making the center of the highest-probability token a natural readout choice.
However, each visual token covers a patch of the screenshot, so selecting the peak-token center quantizes the predicted coordinate to the visual token grid.
Even when the peak token contains the target center, its center can differ from the target center by up to half a patch width or height.
For GUI elements smaller than or comparable to the patch size, this offset may place the predicted coordinate outside the target boundary.
We therefore refine the coordinate using the local probability distribution around the peak.
Based on validation performance, we use a $3\times3$ neighborhood around the peak token $v^\star=\argmax_{v_i\in V}s_{T,i}$, clipped to the $H_v\times W_v$ grid.
Let $\mathcal N(v^\star)$ denote the visual tokens in this neighborhood.
Using zero-based column and row indices, we define the normalized patch center of $v_i$ as
\begin{equation}
    \operatorname{Center}(v_i)=\left(
    (\operatorname{Col}(v_i)+\tfrac12)/W_v,\;
    (\operatorname{Row}(v_i)+\tfrac12)/H_v
    \right).
\end{equation}
The final click coordinate is
\begin{equation}
    \hat{\mathbf{c}}=
    \frac{\sum_{v_i\in\mathcal N(v^\star)}s_{T,i}\operatorname{Center}(v_i)}
         {\sum_{v_i\in\mathcal N(v^\star)}s_{T,i}}.
    \label{eq:coordinate-readout}
\end{equation}
This local weighted average interpolates among the centers of the peak token and its adjacent tokens, mitigating patch-level quantization while excluding distant probability mass from the readout.
Multiplying the two normalized components by the original image width and height gives pixel coordinates.
The ablation in Appendix~\ref{app:coordinate-readout-ablation} shows that this local readout outperforms both selecting the peak-token center and averaging over a $5\times5$ neighborhood.

\section{Implementation and reproducibility details}
\label{app:implementation-and-reproducibility-details}

\subsection{Dataset and constructed trajectories}
\label{app:dataset-statistics}

\textbf{Dataset sampling.}
We construct the sample pool from the 700K-instruction subset of GroundCUA~\citep{feizi2026grounding}, whose authors collected expert demonstrations on desktop applications and deduplicated the resulting data to reduce interface redundancy.
We discard samples with a missing screenshot, an unparseable target bounding box, or no valid click coordinate, leaving 697,892 valid grounding samples.
Using random seed 42, we select 100,000 samples for training and 5,000 for validation.

\begin{table}[t]
    \centering
    \caption{Statistics of fixed-step trajectory construction over the 105,000 training and validation samples.
    Exact, truncated, and padded refer to mutually exclusive outcomes when adjusting the key-step-selected path to $T$ steps.
    Candidate-set sizes are sample means in order from $C_1$ to $C_T$.
    }
    \label{tab:trajectory-statistics}
    \fittable{%
    \begin{tabular}{ccccc}
        \toprule
        \textbf{$T$} & \textbf{Exact} & \textbf{Truncated} & \textbf{Padded}
        & \textbf{Mean candidate-set sizes, $|C_1|\rightarrow\cdots\rightarrow|C_T|$} \\
        \midrule
        3 & 102,798 (97.90\%) & 0 (0.00\%) & 2,202 (2.10\%)
        & $74.25\rightarrow11.35\rightarrow1.00$ \\
        \rowcolor{TableLightBlue}
        \textbf{4} & 94,159 (89.68\%) & 69 (0.07\%) & 10,772 (10.26\%)
        & $74.25\rightarrow28.01\rightarrow10.11\rightarrow1.00$ \\
        5 & 78,449 (74.71\%) & 167 (0.16\%) & 26,384 (25.13\%)
        & $74.25\rightarrow39.24\rightarrow19.19\rightarrow8.00\rightarrow1.00$ \\
        \bottomrule
    \end{tabular}%
    }
\end{table}

\textbf{Trajectory statistics.}
We report trajectory statistics over the 105,000 training and validation samples.
Before key-step selection, the number of candidate sets in each target-containing recursive partition path, including both endpoints, has a mean of 9.36, a median of 9, and an interquartile range of 7--11.
As illustrated by Alg.~\ref{alg:key-step-selection-and-fixed-step-trajectory}, we categorize fixed-step trajectory construction as exact, truncated, or padded.
Based on validation performance, we set the local-reduction threshold to $k=2$ for all settings and use $T=4$ by default.
Table~\ref{tab:trajectory-statistics} shows that truncation remains below 0.2\% for $T\in\{3,4,5\}$, whereas padding increases with the target number of steps.
At the default $T=4$, the mean candidate-set size contracts from 74.25 to 28.01, 10.11, and 1.00.
Moreover, 89.68\% of samples provide exactly two intermediate key steps and require neither truncation nor padding.
Compared with $T=3$, four steps provide one additional intermediate step; compared with $T=5$, they reduce padding from 25.13\% to 10.26\%, providing data-level context for the default $T=4$.

\begin{figure}[t]
    \centering
    \includegraphics[
        width=\linewidth,
        trim={0.00bp 3.34bp 3.00bp 2.58bp},
        clip
    ]{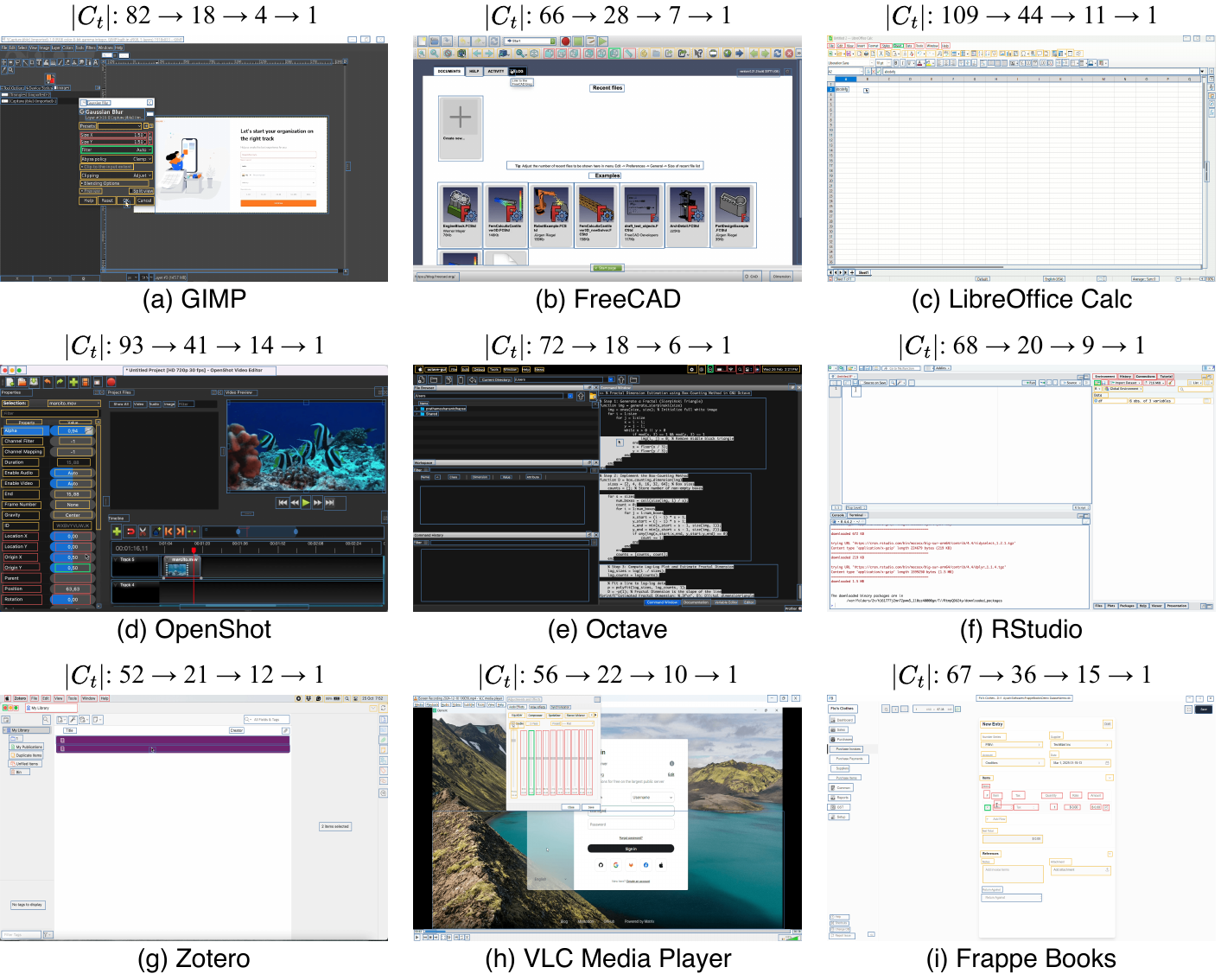}
    \caption{Constructed GUI-aware search trajectories with $T=4$ across representative applications.
    The sequence above each panel reports $|C_1|\rightarrow\cdots\rightarrow|C_4|$, while blue, yellow, red, and green bounding boxes mark candidate elements in $C_1\setminus C_2$, $C_2\setminus C_3$, $C_3\setminus C_4$, and $C_4=\{e^\star\}$, respectively.
    The candidate sets contract from the full element set $E$ through intermediate groups to the target.}
    \label{fig:gui-trajectory-examples}
\end{figure}

\textbf{Trajectory visualization.}
\label{app:dataset-visualizations}
Fig.~\ref{fig:gui-trajectory-examples} presents constructed GUI-aware search trajectories for nine common applications spanning image editing, CAD, office productivity, video editing, scientific computing, reference management, media playback, and accounting.
In these examples, the intermediate candidate sets narrow the search through broad functional regions and local element groups before reaching the target, illustrating a GUI-aware coarse-to-fine search process.
Together with the close agreement with UI-tree hierarchies in Appendix~\ref{app:agreement-with-ui-tree-hierarchies}, these results illustrate how our trajectory construction can approximate the GUI hierarchy using flat GUI element annotations.

\subsection{Training configuration}
\label{app:training-configuration}

\textbf{Model and trainable modules.}
We initialize ReLaViS-7B from Qwen2.5-VL-7B~\citep{bai2025qwen25vltechnicalreport} and randomly initialize the spatial search head.
In Stages~2 and~3, we apply LoRA adapters~\citep{hu2021lora} to all linear layers of the LLM backbone except the output head, using rank 32, $\alpha=64$, and dropout $0.05$.
Stage~1 trains only the spatial search head with the VLM frozen.
Stages~2 and~3 train the spatial search head, LoRA adapters, visual encoder, and multimodal merger while keeping the base weights of the LLM backbone and token embeddings frozen.

\textbf{Optimization.}
All three stages use AdamW with weight decay $0.1$ and a cosine learning-rate schedule with warm-up over the first $3\%$ of steps in each stage.
The spatial search head uses a learning rate of $10^{-4}$ in every stage; in Stages~2 and~3, the LoRA adapters in the LLM backbone, the multimodal merger, and the visual encoder use learning rates of $10^{-4}$, $10^{-5}$, and $2\times10^{-6}$, respectively.
Training uses BF16 precision and a global batch size of 128 with 16 gradient-accumulation steps.
Each stage runs for one epoch, corresponding to 782 optimization steps on the 100K training set.

\textbf{Hyperparameters, random seeds, and checkpoint selection.}
We select $k$, $T$, and the FGP and CWC loss weights based on point grounding accuracy on the 5,000-sample GroundCUA validation set.
The selected settings include $k=2$, $T=4$, and the loss weights described in Appendix~\ref{app:loss-weighting}: unit weights for $\mathcal L_{\mathrm{NTP}}$ and $\mathcal L_{\mathrm{FGP}}^{(T)}$, and $\lambda_{\mathrm{CWC}}^{(T)}=\lambda_{\mathrm{CWC}}^{(1:T-1)}=5$.
To ensure the stability and reproducibility of results, we train each model with three random seeds (42, 43, and 44) in the main controlled comparisons, ablations, and analyses, and report the mean and standard deviation across runs.
After each training stage completes all optimization steps, its final checkpoint initializes the next stage.
We evaluate the final checkpoint of each run, without selecting checkpoints by validation or test performance.

\subsection{Comparison and ablation configuration}
\label{app:comparison-and-ablation-configuration}

\textbf{Controlled comparisons.}
Single-step-7B (coordinate-based) directly generates a textual click coordinate.
Single-step-7B (coordinate-free) sets $T=1$ and uses the spatial search head to predict a spatial search distribution over visual tokens, followed by the same deterministic coordinate readout as ReLaViS-7B.
All three models use the same 100K training set, evaluation protocol, and optimization settings described above.
Since single-step-7B (coordinate-based) has no spatial search head, we train it with coordinate-generation SFT throughout all three stages, updating the same VLM modules as in Stages~2 and~3 of ReLaViS-7B.

\textbf{Ablations and analyses.}
Except for the changes specified in each experiment, all ablation variants retain the ReLaViS-7B configuration.
We evaluate these variants on ScreenSpot-Pro, whose high-resolution professional desktop interfaces and small targets provide a focused setting for examining complex grounding cases.

\textbf{Thinking with images.}
The four baselines for thinking with images cover the mechanisms discussed in Sec.~\ref{sec:thinking-about-images-and-thinking-with-images}.
MVP~\citep{zhang2026mvp} employs multi-view aggregation, and UI-Zoomer~\citep{tang2026ui} utilizes adaptive zoom and region search.
GUI-Spotlight~\citep{lei2025gui} learns to use visual tools, and GUI-Cursor~\citep{zhao2025learning} performs iterative refinement from visual feedback.
MVP and UI-Zoomer are training-free; we reproduce them using their official implementations with Qwen2.5-VL-7B as the backbone.
We directly evaluate the public GUI-Spotlight checkpoint, which is based on Qwen2.5-VL-7B.
GUI-Cursor has no publicly released implementation at this time, so we use the results reported in its paper.
We also implement a simple zoom-in baseline to explore the potential for combining ReLaViS with thinking with images.
The baseline first grounds the instruction on the full screenshot and defines a window centered on the predicted click, with width and height each one-third of the corresponding screenshot dimension.
If the window extends beyond the image boundaries, we shift it to fit within the screenshot while preserving its size.
We then crop the selected region and resize it to the original screenshot resolution with bicubic interpolation, producing a $3\times$ linear zoom.
The same model grounds the original instruction on this enlarged crop, and we map the prediction back to the original screenshot as the final output.
We apply this identical implementation to Qwen2.5-VL-7B, single-step-7B (coordinate-free), and ReLaViS-7B.

\subsection{Training and inference cost}
\label{app:training-and-inference-cost}

\begin{table}[t]
    \centering
    \caption{Training cost. Times are approximate wall-clock hours on 4 NVIDIA L40-48G GPUs.}
    \label{tab:training-cost}
    \fittable{%
    \begin{tabular}{lcccc}
        \toprule
        \multirow{2}{*}[-0.6ex]{\textbf{Model}}
        & \multicolumn{4}{c}{\textbf{Training time (h)}} \\
        \cmidrule(l){2-5}
        & \textbf{Stage~1} & \textbf{Stage~2} & \textbf{Stage~3} & \textbf{Total} \\
        \midrule
        Single-step-7B (coordinate-based) & $\sim$30 & $\sim$30 & $\sim$30 & $\sim$90 \\
        Single-step-7B (coordinate-free) & $\sim$11 & $\sim$29 & $\sim$29 & $\sim$69 \\
        \textbf{ReLaViS-7B} & $\sim$11 & $\sim$29 & $\sim$48 & $\sim$88 \\
        \bottomrule
    \end{tabular}%
    }
\end{table}

\textbf{Training cost.}
All training runs use 4 NVIDIA L40-48G GPUs with a per-device batch size of 2.
Table~\ref{tab:training-cost} reports the approximate training time of each stage.
ReLaViS-7B requires approximately 88 hours for the complete training procedure, comparable to the 90 hours required by single-step-7B (coordinate-based) and higher than the 69 hours required by single-step-7B (coordinate-free).
The additional training time relative to single-step-7B (coordinate-free) occurs in Stage~3, where ReLaViS performs sequential no-gradient probes before the differentiable main pass, consistent with the forward-pass analysis in Appendix~\ref{app:forward-pass-cost}.

\begin{table}[t]
    \centering
    \caption{Inference cost.
    Interaction rounds are averaged across all five benchmarks.
    Latency, FLOPs, and peak allocated GPU memory are reported for ScreenSpot-Pro.}
    \label{tab:inference-cost}
    \fittable{%
    \begin{tabular}{lcccc}
        \toprule
        \multirow{2}{*}{\textbf{Model}}
        & \textbf{Interaction} & \textbf{Latency} & \textbf{TFLOPs} & \textbf{Peak memory} \\
        & \textbf{rounds} & \textbf{(s/sample)} & \textbf{(/sample)} & \textbf{(GiB)} \\
        \midrule
        Single-step-7B (coordinate-based) & 1.0 & 2.29 & 172.00 & 18.98 \\
        Single-step-7B (coordinate-free) & 1.0 & 2.06 & 173.51 & 18.78 \\
        \textbf{ReLaViS-7B} & 1.0 & 2.19 & 179.65 & 19.54 \\
        \midrule
        Zoom-in + Qwen2.5-VL-7B & 2.0 & 4.67 & 344.01 & 18.98 \\
        Zoom-in + single-step-7B (coordinate-free) & 2.0 & 4.28 & 347.03 & 18.78 \\
        \textbf{Zoom-in + ReLaViS-7B} & 2.0 & 4.57 & 359.31 & 19.54 \\
        \midrule
        MVP~\smalltablecitep{zhang2026mvp} + Qwen2.5-VL-7B & 4.8 & 11.97 & 786.39 & 18.00 \\
        UI-Zoomer~\smalltablecitep{tang2026ui} + Qwen2.5-VL-7B & 2.0 & 2.34 & 148.70 & 39.67 \\
        GUI-Spotlight~\tablecitep{lei2025gui} & 2.3 & 9.69 & 532.09 & 45.05 \\
        \bottomrule
    \end{tabular}
    }
\end{table}

\textbf{Inference cost.}
We average interaction rounds across all five benchmarks and measure latency and peak allocated memory on ScreenSpot-Pro using a single L40 GPU, BF16 precision, a batch size of 1, and deterministic decoding.
End-to-end latency includes image preprocessing, model generation, and prediction postprocessing.
We analytically estimate FLOPs from each sample's actual token and patch counts and report the average in TFLOPs/sample.
The estimate covers the visual encoder, LLM backbone prefill and decoding, and method-specific operations, including the spatial search head and visual evidence propagation for ReLaViS.
For thinking with images baselines, latency measurements and FLOPs estimates cover all interaction rounds.
Peak allocated memory measures the maximum live tensor allocation over inference.
Inference cost results for MVP, UI-Zoomer, and GUI-Spotlight serve as references under their official implementations.

As shown in Table~\ref{tab:inference-cost}, ReLaViS improves grounding accuracy while keeping inference cost close to the single-step baselines.
Relative to single-step-7B (coordinate-free), ReLaViS-7B gains 3.1~pp in ScreenSpot-Pro accuracy (Table~\ref{tab:main-results-all-benchmarks}).
The corresponding increases are 6.3\% in mean latency, 3.5\% in FLOPs, and 0.76~GiB in peak allocated memory.
Compared with single-step-7B (coordinate-based), ReLaViS improves accuracy by 10.0~pp with 4.4\% lower mean latency, 4.4\% more FLOPs, and 0.56~GiB more peak allocated memory.
Recurrent search thus improves grounding with modest increases in FLOPs and memory while keeping latency comparable to that of single-step grounding.

This favorable accuracy--efficiency trade-off also holds with external zoom-in.
Zoom-in + ReLaViS achieves 63.5\% ScreenSpot-Pro accuracy, compared with 62.3\% for zoom-in + single-step-7B (coordinate-free), with 6.8\% higher mean latency and 3.5\% more FLOPs.
Without zoom-in, ReLaViS already exceeds zoom-in + Qwen2.5-VL-7B in accuracy (56.3\% versus 49.8\%) while reducing latency by 53.1\% and FLOPs by 47.8\%.

\subsection{Evaluation benchmarks and metrics}
\label{app:evaluation-benchmarks-and-metrics}

\textbf{Benchmarks.}
We evaluate on ScreenSpot-Pro~\citep{li2025screenspot}, MMBench-GUI-L2~\citep{wang2025mmbench}, OSWorld-G-Refine~\citep{xie2026scaling}, ScreenSpot-v2~\citep{wu2025atlas}, and UI-Vision~\citep{nayak2025ui}, spanning high-resolution professional desktop interfaces and general desktop, web, and mobile interfaces.
As noted by the GroundCUA~\citep{feizi2026grounding} authors, UI-Vision overlaps with GroundCUA in terms of application coverage, and we therefore treat it as an in-domain benchmark.
The cross-benchmark average is the arithmetic mean of the headline scores from the five benchmarks.

\textbf{Search dynamics metrics.}
\label{app:search-dynamics-metrics}
For a spatial search distribution $s_t$ over $N$ visual tokens, we measure its concentration using its entropy, defined as $\operatorname{Entropy}(s_t)=-\sum_{i=1}^{N}s_{t,i}\log s_{t,i}$.
Thus, a lower value indicates a more concentrated spatial search distribution.

The probability mass of the target element $e^\star$ is $m_{t,e^\star}=\sum_{v_i\in\Omega(e^\star)}s_{t,i}$.
A higher target probability mass indicates that more probability is assigned to the visual token support of the target element.

To measure proximity to the target element $e^\star$, let visual token $v_i$ have coordinates $(x_i,y_i)$ on an $H_v\times W_v$ grid, where $x_i\in\{0,\ldots, W_v-1\}$ and $y_i\in\{0,\ldots, H_v-1\}$.
We define the distance from token $v_i$ to the nearest target-support token in $\Omega(e^\star)$, with normalization by the grid diagonal, as
\begin{equation}
    \operatorname{Distance}(v_i,e^\star)
    =\frac{
        \min_{v_j\in\Omega(e^\star)}
        \lVert(x_i,y_i)-(x_j,y_j)\rVert_2
    }{
        \sqrt{H_v^2+W_v^2}
    }.
    \label{eq:token-to-target-distance}
\end{equation}
The expected distance under the spatial search distribution $s_t$ is
\begin{equation}
    \operatorname{ExpectedDistance}(s_t,e^\star)
    =\sum_{i=1}^{N}s_{t,i}\,\operatorname{Distance}(v_i,e^\star).
    \label{eq:expected-target-distance}
\end{equation}
Target-support tokens have zero distance; a lower expected distance indicates a search focus closer to the target element on the visual token grid.

\section{Additional results and analyses}
\label{app:additional-results-and-analyses}

\subsection{Target-size analysis within each benchmark}
\label{app:target-size-within-benchmark}

\begin{table}[t]
    \centering
    \caption{Accuracy gains (pp) of ReLaViS-7B over matched single-step baselines across relative target area bins within each benchmark.
    Relative target area is the target bounding box area divided by the screenshot area.
    Gains are reported as mean\resultstd{std} over three random seeds; $n$ is the instance count in each bin.
    Within each benchmark and baseline comparison, blue and red indicate the highest and lowest mean gains, respectively.}
    \label{tab:target-size-within-benchmark}
    \fittable{%
    \renewcommand{\arraystretch}{1.15}%
    \newcommand{\sizecell}[3][-0.5pt]{%
        \raisebox{#1}[\height][\depth]{%
            \begin{tabular}[c]{@{}w{c}{4.5em}@{}}#2\\[-3pt]#3\end{tabular}}}%
    \begin{tabular}{l|ccc|ccc}
        \toprule
        & \multicolumn{3}{c|}{\textbf{vs. single-step (coordinate-based)}}
        & \multicolumn{3}{c}{\textbf{vs. single-step (coordinate-free)}} \\
        \midrule
        \textbf{Benchmark}
        & \sizecell[0pt]{\textbf{Large}}{\scriptsize $\geq 0.25\%$}
        & \sizecell[0pt]{\textbf{Medium}}{\scriptsize $[0.09\%, 0.25\%)$}
        & \sizecell[0pt]{\textbf{Small}}{\scriptsize $< 0.09\%$}
        & \sizecell[0pt]{\textbf{Large}}{\scriptsize $\geq 0.25\%$}
        & \sizecell[0pt]{\textbf{Medium}}{\scriptsize $[0.09\%, 0.25\%)$}
        & \sizecell[0pt]{\textbf{Small}}{\scriptsize $< 0.09\%$} \\
        \midrule
        ScreenSpot-Pro
            & \cellcolor{TableLightBlue}\sizecell{+13.9\resultstd{4.81}}{\scriptsize $n = 36$}
            & \cellcolor{TableLightRed}\sizecell{+4.5\resultstd{3.81}}{\scriptsize $n = 273$}
            & \sizecell{+11.1\resultstd{0.94}}{\scriptsize $n = 1{,}272$}
            & \cellcolor{TableLightRed}\sizecell{+0.9\resultstd{4.24}}{\scriptsize $n = 36$}
            & \sizecell{+2.6\resultstd{1.68}}{\scriptsize $n = 273$}
            & \cellcolor{TableLightBlue}\sizecell{+3.3\resultstd{0.69}}{\scriptsize $n = 1{,}272$} \\
        MMBench-GUI-L2
            & \cellcolor{TableLightRed}\sizecell{+0.1\resultstd{0.72}}{\scriptsize $n = 1{,}649$}
            & \sizecell{+1.2\resultstd{0.47}}{\scriptsize $n = 857$}
            & \cellcolor{TableLightBlue}\sizecell{+6.6\resultstd{0.74}}{\scriptsize $n = 1{,}088$}
            & \sizecell{+0.8\resultstd{0.53}}{\scriptsize $n = 1{,}649$}
            & \cellcolor{TableLightRed}\sizecell{+0.7\resultstd{0.34}}{\scriptsize $n = 857$}
            & \cellcolor{TableLightBlue}\sizecell{+1.0\resultstd{0.69}}{\scriptsize $n = 1{,}088$} \\
        OSWorld-G-Refine
            & \cellcolor{TableLightRed}\sizecell{-0.5\resultstd{2.20}}{\scriptsize $n = 131$}
            & \sizecell{0.0\resultstd{2.60}}{\scriptsize $n = 77$}
            & \cellcolor{TableLightBlue}\sizecell{+5.6\resultstd{2.98}}{\scriptsize $n = 302$}
            & \sizecell{+3.1\resultstd{2.02}}{\scriptsize $n = 131$}
            & \cellcolor{TableLightRed}\sizecell{-0.4\resultstd{4.92}}{\scriptsize $n = 77$}
            & \cellcolor{TableLightBlue}\sizecell{+4.7\resultstd{3.77}}{\scriptsize $n = 302$} \\
        ScreenSpot-v2
            & \cellcolor{TableLightRed}\sizecell{-0.3\resultstd{0.52}}{\scriptsize $n = 1{,}019$}
            & \sizecell{+2.8\resultstd{2.02}}{\scriptsize $n = 200$}
            & \cellcolor{TableLightBlue}\sizecell{+11.9\resultstd{2.88}}{\scriptsize $n = 53$}
            & \cellcolor{TableLightRed}\sizecell{+0.3\resultstd{0.34}}{\scriptsize $n = 1{,}019$}
            & \sizecell{+0.5\resultstd{2.18}}{\scriptsize $n = 200$}
            & \cellcolor{TableLightBlue}\sizecell{+0.6\resultstd{2.18}}{\scriptsize $n = 53$} \\
        UI-Vision
            & \cellcolor{TableLightRed}\sizecell{-1.3\resultstd{1.85}}{\scriptsize $n = 359$}
            & \sizecell{+11.2\resultstd{0.36}}{\scriptsize $n = 1{,}048$}
            & \cellcolor{TableLightBlue}\sizecell{+18.9\resultstd{0.58}}{\scriptsize $n = 4{,}071$}
            & \cellcolor{TableLightRed}\sizecell{+0.1\resultstd{2.02}}{\scriptsize $n = 359$}
            & \sizecell{+3.7\resultstd{1.03}}{\scriptsize $n = 1{,}048$}
            & \cellcolor{TableLightBlue}\sizecell{+7.9\resultstd{0.90}}{\scriptsize $n = 4{,}071$} \\
        \bottomrule
    \end{tabular}%
    }
\end{table}

To reduce confounding from benchmark composition in the pooled analysis (Fig.~\ref{fig:target-area-and-state-injection}, left), we further examine the association between accuracy gains and target size within each benchmark.
Table~\ref{tab:target-size-within-benchmark} reports the accuracy gains of ReLaViS-7B over the two matched single-step baselines across all five benchmarks, using the same three relative target area bins ordered from large to medium to small.
Against the coordinate-based baseline, the small-target bin has the largest mean gain and the large-target bin has the smallest mean gain in four of the five benchmarks.
Similarly, against the coordinate-free baseline, the small-target bin has the largest mean gain in all five benchmarks, and the large-target bin has the smallest mean gain in ScreenSpot-Pro, ScreenSpot-v2, and UI-Vision.
These within-benchmark results further support the association observed in the pooled analysis and suggest greater gains on small targets.

\subsection{Results at a smaller model scale}
\label{app:smaller-model-scale}

\begin{table}[t]
    \centering
    \caption{Performance comparison of models based on Qwen2.5-VL-3B across five GUI grounding benchmarks.
    For models trained in this work, accuracy (\%) is reported as mean\resultstd{std} over three random seeds.
    $\dagger$ denotes a result reproduced using the official implementation.
    \textbf{Bold} values indicate the best results.}
    \label{tab:model-scale-3b}
    \fittable{%
    \setlength{\tabcolsep}{2pt}
    \begin{tabular}{lcccccc}
        \toprule
        \multirow{2}{*}{\textbf{Model}}
        & \textbf{ScreenSpot-} & \textbf{MMBench-} & \textbf{OSWorld-}
        & \textbf{ScreenSpot-} & \textbf{UI-} & \multirow{2}{*}{\textbf{Average}} \\
        & \textbf{Pro} & \textbf{GUI-L2} & \textbf{G-Refine}
        & \textbf{v2} & \textbf{Vision} & \\
        \midrule
        \multicolumn{7}{c}{\textit{Zero-shot general VLMs}} \\
        Qwen2.5-VL-3B$^\dagger$~\tablecitep{bai2025qwen25vltechnicalreport}
        & 18.4 & 55.2 & 38.8 & 69.6 & 7.4 & 37.9 \\
        \midrule
        \multicolumn{7}{c}{\textit{Controlled comparison (identical training data)}} \\
        Single-step-3B (coordinate-free)
        & 46.0\resultstd{0.89} & 73.6\resultstd{0.41} & 53.4\resultstd{0.10}
        & 85.8\resultstd{0.28} & 31.4\resultstd{0.06} & 58.1\resultstd{0.17} \\
        \textbf{ReLaViS-3B}
        & \textbf{48.3}\resultstd{0.33} & \textbf{75.6}\resultstd{0.12} & \textbf{55.6}\resultstd{0.82}
        & \textbf{87.3}\resultstd{0.33} & \textbf{39.1}\resultstd{0.38} & \textbf{61.2}\resultstd{0.18} \\
        \bottomrule
    \end{tabular}
    }
\end{table}

We evaluate ReLaViS on Qwen2.5-VL-3B to examine whether the gains from recurrent latent visual search persist at a smaller model scale.
We compare ReLaViS-3B with single-step-3B (coordinate-free), its $T=1$ variant.
Both models use the same settings as their corresponding 7B models except for the smaller model scale and halved learning rates.
Table~\ref{tab:model-scale-3b} shows that ReLaViS-3B outperforms the single-step baseline on all five benchmarks, improving average accuracy by 3.1~pp, from 58.1\% to 61.2\%.
Together with the 7B results in Table~\ref{tab:main-results-all-benchmarks}, these results demonstrate the effectiveness of ReLaViS across different model scales within the Qwen2.5-VL family.

\subsection{Additional ablations}
\label{app:additional-ablations-and-analyses}

\textbf{Coordinate readout.}
\label{app:coordinate-readout-ablation}
We compare four deterministic coordinate readouts on ScreenSpot-Pro while keeping the ReLaViS-7B model and evaluation setting fixed.
Selecting the center of the highest-probability visual token yields 46.7\resultstd{0.48}\% accuracy.
Region readout groups tokens whose probabilities exceed 30\% of the highest token probability into connected regions and selects the region with the highest mean token probability.
Taking its probability-weighted center improves accuracy to 53.5\resultstd{0.10}\%.
Local probability-weighted averaging around the peak performs better: a $5\times5$ neighborhood achieves 55.1\resultstd{0.40}\%, whereas the default $3\times3$ neighborhood achieves the highest accuracy of 56.3\resultstd{0.29}\%.
The 9.6~pp gain over the peak-token center supports the use of local interpolation to mitigate patch-level quantization error.
The 1.2~pp advantage over the broader $5\times5$ neighborhood suggests that a compact window limits the contribution of less relevant probability mass.

\textbf{Intermediate CWC loss weight.}
\label{app:intermediate-cwc-loss-weight}
We vary $\lambda_{\mathrm{CWC}}^{(1:T-1)}$ while keeping all other training and evaluation configurations unchanged.
This experiment examines how the weight on intermediate-step candidate-wise supervision affects point grounding accuracy.
On ScreenSpot-Pro, the default weight of $5$ yields 56.3\resultstd{0.29}\% accuracy, compared with 54.3\resultstd{1.27}\% at $15$ and 52.9\resultstd{0.63}\% at $1$.
The tested lower and higher weights both reduce grounding accuracy relative to the default weight.

\textbf{Intermediate supervision objective.}
\label{app:intermediate-supervision-objective}
We replace the intermediate CWC term in Eq.~(\ref{eq:overall-objective}) with a direct token-level distributional objective that combines a KL loss and a global negative loss.
The KL loss $\mathcal L_{\mathrm{KL}}^{(t)}=D_{\mathrm{KL}}(d_t\,\|\,s_t)$ matches $s_t$ to the Gaussian target distribution $d_t$.
The global negative loss $\mathcal L_{\mathrm{neg}}^{(t)}=\sum_{v_i\notin\bigcup_{e\in C_t}\Omega(e)}s_{t,i}$ penalizes probability mass outside the visual token support of the candidate elements.
We average each loss over the $T-1$ intermediate steps and include these averages in the overall objective with weights $1$ and $5$, respectively, and keep all other training and evaluation configurations unchanged.
On ScreenSpot-Pro, this alternative achieves 54.8\resultstd{0.86}\% accuracy, 1.5~pp below CWC at 56.3\resultstd{0.29}\%.
These results favor candidate-wise coverage supervision over the tested token-level distributional objective.

\subsection{Search trajectory and state analyses}
\label{app:search-trajectory-and-state-analyses}

\begin{table}[t]
    \centering
    \caption{Agreement with reference trajectories derived from UI trees on Screen2AX-Tree.
    IoUs are averaged over targets within each screenshot and then across the 1,125 screenshots, which contain 29,901 targets in total.
    The two intermediate candidate sets are jointly matched to reference steps in coarse-to-fine order.
    }
    \label{tab:screen2ax-tree-agreement}
    \fittable{
    \begin{tabular}{lcc}
        \toprule
        \textbf{Search trajectory} & $\boldsymbol{C_2}$ \textbf{IoU (\%)} $\uparrow$ & $\boldsymbol{C_3}$ \textbf{IoU (\%)} $\uparrow$ \\
        \midrule
        \textbf{GUI-aware} & \textbf{75.1} & \textbf{74.7} \\
        Geometric zoom & 62.4 & 57.9 \\
        \bottomrule
    \end{tabular}
    }
\end{table}

\textbf{Agreement with UI-tree hierarchies.}
\label{app:agreement-with-ui-tree-hierarchies}
We evaluate how closely our constructed trajectories approximate a GUI hierarchy using 1,125 macOS screenshots with nonempty UI-tree annotations from Screen2AX-Tree~\citep{muryn2025screen2ax}.
We treat the leaf nodes of the UI tree as a flat set of elements $E$.
For each target in $E$, we follow its parent links to recover the root-to-target ancestor chain, which determines the reference trajectory steps.
Starting from $E$, we intersect the current candidate set with the elements enclosed by the bounding box of each successive node in this chain (e.g., windows, groups, and toolbars) and retain only steps that strictly reduce the set, ending with the target element.
For each target, we measure agreement with the reference GUI hierarchy by matching the two intermediate candidate sets $C_2, C_3$ of each evaluated four-step search trajectory to reference steps.
We select the pair of reference steps that maximizes the sum of the two matched set IoUs while preserving coarse-to-fine order, with IoU computed over element ID sets.
As shown in Table~\ref{tab:screen2ax-tree-agreement}, the GUI-aware trajectory improves set IoU by 12.7~pp at $C_2$ and by 16.8~pp at $C_3$ over geometric zoom, indicating a closer approximation to the GUI hierarchy.

\begin{figure}[t]
    \centering
    \includegraphics[
        width=0.6\linewidth,
        trim={2.21bp 2.34bp 2.44bp 3.36bp},
        clip
    ]{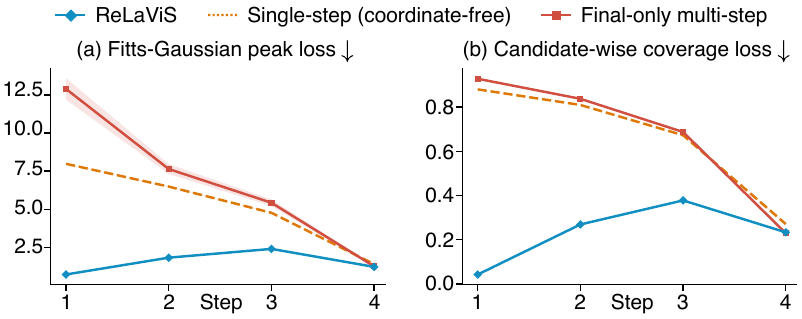}
    \caption{Stepwise diagnostic $\mathcal L_{\mathrm{FGP}}$ (a) and $\mathcal L_{\mathrm{CWC}}$ (b) on the 5,000-sample GroundCUA validation set.
    Curves show multi-seed means with std shading.
    }
    \label{fig:candidate-loss-dynamics}
\end{figure}

\textbf{Stepwise search-state diagnostics.}
\label{app:stepwise-search-state-diagnostics}
At each step, we evaluate $\mathcal L_{\mathrm{CWC}}^{(t)}$ against the candidate set $C_t$ and the KL divergence used by $\mathcal L_{\mathrm{FGP}}$ against the Gaussian target distribution $d_t$.
The intermediate-step KL serves only as a diagnostic and does not contribute to training.
For the single-step model, we score its single predicted distribution against each of the four step-specific targets.
Fig.~\ref{fig:candidate-loss-dynamics} shows that ReLaViS yields substantially lower losses than the single-step (coordinate-free) and final-only multi-step variants at the first three search steps, while all three models have comparable losses at the final step.
Lower intermediate $\mathcal L_{\mathrm{CWC}}$ indicates better candidate coverage, while lower diagnostic KL indicates closer agreement with $d_t$.
These diagnostics show that ReLaViS's intermediate spatial search distributions align more closely with the constructed GUI-aware trajectories.

\subsection{Visualizations and qualitative analyses}
\label{app:visualizations-and-qualitative-analyses}

\begin{figure}[t]
    \centering
    \includegraphics[
        width=0.9535\linewidth,
        trim={0.36bp 4.95bp 3.23bp 4.98bp},
        clip
    ]{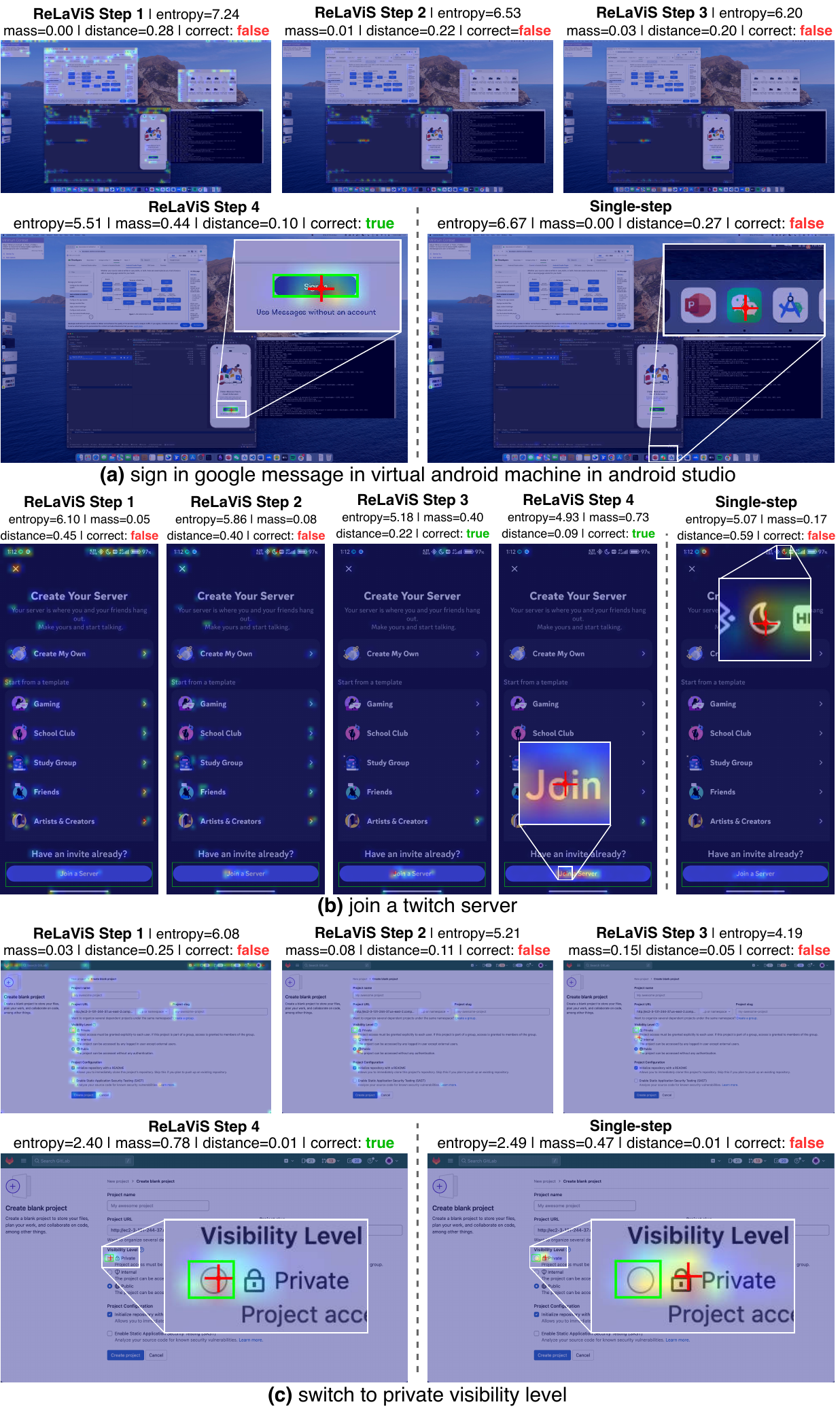}
    \caption{Successful cases from ScreenSpot-Pro (panel (a)) and ScreenSpot-v2 (panels (b) and (c)).}
    \label{fig:success-ss-pro-and-ss-v2}
\end{figure}

\begin{figure}[t]
    \centering
    \includegraphics[
        width=0.9204\linewidth,
        trim={0.00bp 3.87bp 2.99bp 4.99bp},
        clip
    ]{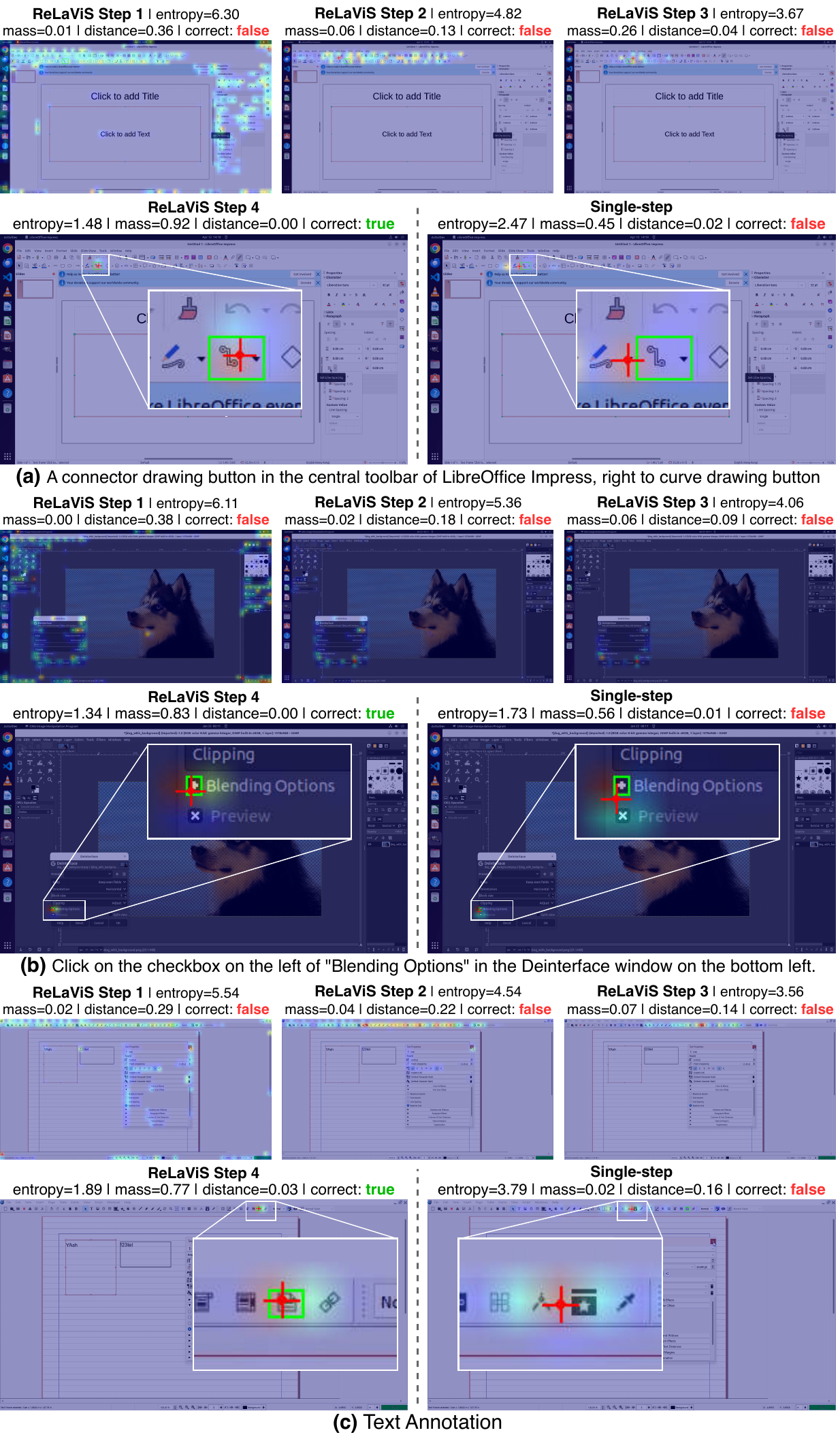}
    \caption{Successful cases from MMBench-GUI-L2 (panel (a)), OSWorld-G-Refine (panel (b)), and UI-Vision (panel (c)).}
    \label{fig:success-mmbench-osworld-uivision}
\end{figure}

\begin{figure}[t]
    \centering
    \includegraphics[
        width=0.9325\linewidth,
        trim={0.36bp 4.78bp 3.18bp 4.98bp},
        clip
    ]{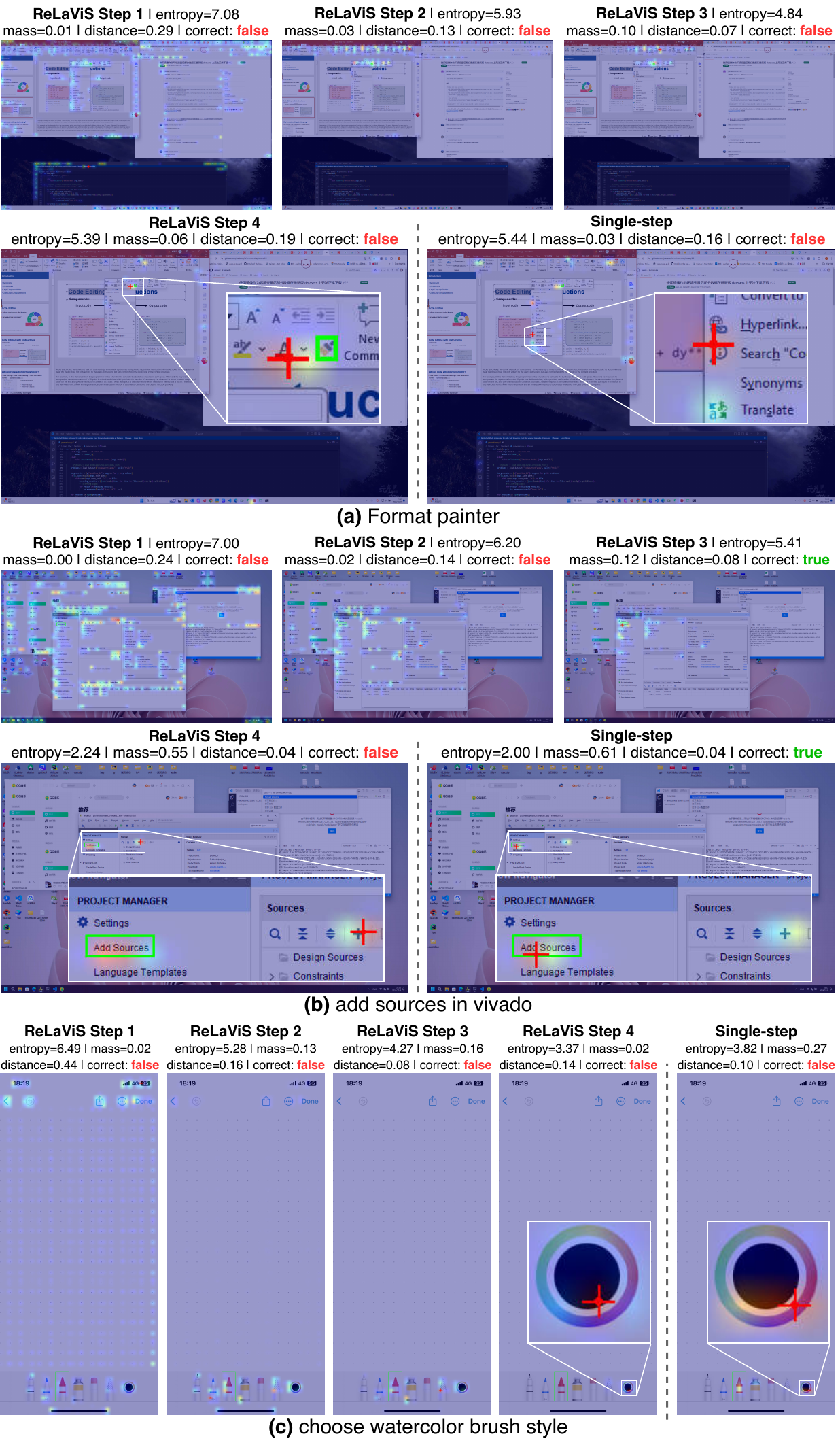}
    \caption{Failure cases from ScreenSpot-Pro (panels (a) and (b)) and ScreenSpot-v2 (panel (c)).}
    \label{fig:failure-ss-pro-and-ss-v2}
\end{figure}

Figs.~\ref{fig:success-ss-pro-and-ss-v2} and~\ref{fig:success-mmbench-osworld-uivision} present six successful cases across all five benchmarks and desktop, mobile, and web platforms.
Fig.~\ref{fig:failure-ss-pro-and-ss-v2} presents three failure cases from ScreenSpot-Pro and ScreenSpot-v2.
For each case, the figures visualize the spatial search distribution across four ReLaViS steps and report its entropy, target probability mass, expected distance to the target element, and click correctness, together with the matched single-step (coordinate-free) result.
Across all six successful cases, entropy and expected distance decrease monotonically, target probability mass increases monotonically, and ReLaViS produces a correct final click where single-step (coordinate-free) fails.
These case-level trajectories qualitatively complement the aggregate dynamics in Sec.~\ref{sec:recurrent-search-dynamics}.

The successful cases illustrate how recurrent search supports candidate disambiguation, instruction-conditioned semantic grounding, and fine-grained localization.
The ScreenSpot-Pro and UI-Vision cases in Figs.~\ref{fig:success-ss-pro-and-ss-v2}(a) and~\ref{fig:success-mmbench-osworld-uivision}(c) show search states consistent with the GUI-aware trajectory: the former first emphasizes Android Studio over other windows; the latter reduces the probability mass assigned to text-related regions before selecting the target toolbar icon.
Both examples illustrate candidate disambiguation as probability mass progressively shifts away from irrelevant regions.
In Fig.~\ref{fig:success-ss-pro-and-ss-v2}(c), the search focus narrows from the full form to the Visibility Level group and then to the Private radio button.
This progression shows that recurrent search grounds a compositional instruction by resolving both the relevant setting group and the required control role.
In Fig.~\ref{fig:success-ss-pro-and-ss-v2}(b), ``Twitch'' does not match any visible interface label, yet the search focus converges on the interactive ``Join a Server'' control.
This case shows that recurrent search remains grounded in the corresponding interactive control when the instruction is noisy.
In Fig.~\ref{fig:success-mmbench-osworld-uivision}(a), the search focus narrows to the toolbar and then to the connector drawing button adjacent to the referenced curve drawing button.
In Fig.~\ref{fig:success-mmbench-osworld-uivision}(b), it narrows to the dialog and then to the small checkbox beside ``Blending Options''.
By resolving the enclosing functional region before distinguishing nearby controls, this coarse-to-fine progression supports fine-grained localization in dense layouts and for small targets.

In the failure case shown in Fig.~\ref{fig:failure-ss-pro-and-ss-v2}(a), the search focus approaches the extremely small Format painter control by step~3.
The deterministic final click lies visually closer to the target than the single-step (coordinate-free) prediction but remains outside the target boundary, consistent with limited spatial resolution on the visual token grid.
In Fig.~\ref{fig:failure-ss-pro-and-ss-v2}(b), ReLaViS shifts from the annotated ``Add Sources'' control at step~3 to the plus icon in the Sources panel at step~4.
Both controls are plausible entry points for ``add sources'', suggesting instruction--annotation ambiguity in this case.
In Fig.~\ref{fig:failure-ss-pro-and-ss-v2}(c), the search contracts to the brush toolbar but ultimately selects the color control instead of the annotated brush.
This failure to associate the abstract visual style ``watercolor'' with the annotated control reveals a limited ability to distinguish abstract, non-standard visual semantics.
These failure cases show that ReLaViS can narrow the search focus to the relevant region, while the final click still drifts under limited visual token grid resolution, instruction--annotation ambiguity, or a limited ability to distinguish abstract visual semantics.

\clearpage
\subsection{Complete benchmark results}
\label{app:complete-benchmark-results}

Tables~\ref{tab:screenspot-pro-detailed}, \ref{tab:mmbench-gui-l2-detailed}, \ref{tab:osworld-g-refine-detailed}, \ref{tab:screenspot-v2-detailed}, and~\ref{tab:ui-vision-detailed} provide category-level results across the five benchmarks, supplementing the benchmark-level summary in Table~\ref{tab:main-results-all-benchmarks}.
Following each benchmark's official taxonomy, the tables break down point grounding accuracy (\%) by target type, platform, software domain, or task category.
For models trained in this work, results are reported as mean\resultstd{std} over three random seeds.
In these tables, ``--'' denotes unavailable results.
Results marked with $\dagger$ are reproduced using official implementations.
Results for other baselines from prior work are taken from the corresponding papers.
Within each column, \textbf{bold} values are the best results within and outside ``thinking with images'' separately; \underline{underlined} values are the second-best results only outside it.

\begin{table}[t]
    \centering
    \caption{Performance comparison on ScreenSpot-Pro.}
    \label{tab:screenspot-pro-detailed}
    \fittable{%
    \setlength{\tabcolsep}{1pt}
    \begin{tabular}{l*{6}{cc}ccc}
        \toprule
        \multirow{2}{*}[-0.4ex]{\textbf{Model}}
        & \multicolumn{2}{c}{\textbf{CAD}} & \multicolumn{2}{c}{\textbf{Dev}}
        & \multicolumn{2}{c}{\textbf{Creative}} & \multicolumn{2}{c}{\textbf{Scientific}}
        & \multicolumn{2}{c}{\textbf{Office}} & \multicolumn{2}{c}{\textbf{OS}}
        & \multicolumn{3}{c}{\textbf{Average}} \\
        \cmidrule(lr){2-3}\cmidrule(lr){4-5}\cmidrule(lr){6-7}
        \cmidrule(lr){8-9}\cmidrule(lr){10-11}\cmidrule(lr){12-13}\cmidrule(lr){14-16}
        & \textbf{Text} & \textbf{Icon} & \textbf{Text} & \textbf{Icon}
        & \textbf{Text} & \textbf{Icon} & \textbf{Text} & \textbf{Icon}
        & \textbf{Text} & \textbf{Icon} & \textbf{Text} & \textbf{Icon}
        & \textbf{Text} & \textbf{Icon} & \textbf{Avg} \\
        \midrule
        \multicolumn{16}{c}{\textit{Zero-shot general VLMs}} \\
        GPT-4o~{\tablecitep{hurst2024gpt}}
        & 2.0 & 0.0 & 1.3 & 0.0 & 1.0 & 0.0 & 2.1 & 0.0 & 1.1 & 0.0 & 0.0 & 0.0 & 1.3 & 0.0 & 0.8 \\
        Claude Computer~{\tablecitep{hu2024dawn}}
        & 14.5 & 3.7 & 22.0 & 3.9 & 25.9 & 3.4 & 33.9 & 15.8 & 30.1 & 16.3 & 11.0 & 4.5 & 23.4 & 7.1 & 17.1 \\
        Qwen2.5-VL-3B$^\dagger$~{\tablecitep{bai2025qwen25vltechnicalreport}}
        & 14.7 & 1.6 & 24.0 & 2.1 & 28.3 & 2.1 & 39.6 & 8.2 & 40.7 & 18.9 & 12.2 & 1.1 & 27.0 & 4.5 & 18.4 \\
        Qwen2.5-VL-7B$^\dagger$~{\tablecitep{bai2025qwen25vltechnicalreport}}
        & 21.8 & 1.6 & 46.8 & 4.8 & 36.4 & 7.0 & 56.3 & 8.2 & 55.9 & 20.8 & 39.3 & 10.1 & 41.9 & 7.8 & 28.8 \\
        Qwen2.5-VL-72B~{\tablecitep{bai2025qwen25vltechnicalreport}}
        & -- & -- & -- & -- & -- & -- & -- & -- & -- & -- & -- & -- & -- & -- & 53.3 \\
        Qwen3-VL-8B~{\tablecitep{bai2025qwen3}}
        & 56.9 & 10.9 & \underline{75.3} & 22.8 & \underline{68.2} & 16.1 & 78.5 & 32.7 & \textbf{80.8} & 39.6 & \underline{71.0} & 20.2 & \textbf{71.1} & 22.8 & 52.7 \\
        Qwen3-VL-30B-A3B~{\tablecitep{bai2025qwen3}}
        & 51.8 & 15.6 & \textbf{76.0} & 24.8 & \textbf{69.2} & 20.3 & 76.4 & 27.3 & \textbf{80.8} & 37.7 & \textbf{75.7} & 38.2 & \underline{70.6} & 26.3 & \underline{53.7} \\
        \midrule
        \multicolumn{16}{c}{\textit{GUI-specialized models (coordinate-based)}} \\
        SE-GUI-7B~{\tablecitep{yuan2026se}}
        & 51.3 & \textbf{42.2} & 68.2 & 19.3 & 57.6 & 9.1 & 75.0 & 28.2 & 78.5 & 43.4 & 49.5 & 25.8 & 63.5 & 21.0 & 47.3 \\
        Jedi-7B~{\tablecitep{xie2026scaling}}
        & 38.0 & 14.1 & 42.9 & 11.0 & 50.0 & 11.9 & 72.9 & 25.5 & 75.1 & 47.2 & 33.6 & 16.9 & 52.6 & 18.2 & 39.5 \\
        UI-TARS-1.5-7B~{\tablecitep{qin2025ui}}
        & -- & -- & -- & -- & -- & -- & -- & -- & -- & -- & -- & -- & -- & -- & 49.6 \\
        GUI-G$^2$-7B~{\tablecitep{tang2026gui}}
        & 55.8 & 12.5 & 68.8 & 17.2 & 57.1 & 15.4 & 77.1 & 24.5 & 74.0 & 32.7 & 57.9 & 21.3 & 64.7 & 19.6 & 47.5 \\
        InfiGUI-G1-7B~{\tablecitep{liu2026infigui}}
        & 57.4 & 23.4 & 74.7 & 24.1 & 64.6 & 15.4 & \underline{80.6} & 31.8 & 75.7 & 39.6 & 57.0 & 29.2 & 68.4 & 25.2 & 51.9 \\
        UI-Ins-7B~{\tablecitep{chen2026ui}}
        & \textbf{60.9} & 20.3 & \underline{75.3} & 18.6 & 65.2 & 18.9 & \textbf{81.3} & 29.1 & \underline{79.7} & 37.7 & 57.0 & 25.8 & -- & -- & 52.2 \\
        GroundNext-7B (SFT)~{\tablecitep{feizi2026grounding}}
        & 46.2 & 32.8 & 68.2 & \underline{38.6} & 54.5 & 20.3 & 70.8 & 37.3 & 76.8 & 49.1 & 45.8 & 33.7 & 59.9 & 33.6 & 50.2 \\
        GroundNext-7B (RL)~{\tablecitep{feizi2026grounding}}
        & 50.2 & \underline{34.3} & 73.4 & \textbf{40.0} & 59.6 & 23.8 & 70.1 & \underline{42.7} & 74.6 & 54.7 & 53.3 & 30.3 & 60.5 & 33.6 & 52.9 \\
        \midrule
        \multicolumn{16}{c}{\textit{GUI-specialized models (coordinate-free)}} \\
        GUI-Actor-7B~{\tablecitep{wu2026gui}}
        & -- & -- & -- & -- & -- & -- & -- & -- & -- & -- & -- & -- & -- & -- & 44.6 \\
        V2P-7B~{\tablecitep{chen2025v2p}}
        & \underline{58.4} & 12.5 & 67.5 & 24.8 & 62.6 & 16.1 & 73.6 & 33.6 & 75.7 & 43.4 & 56.1 & 32.6 & 65.8 & 25.8 & 52.5 \\
        \midrule
        \rowcolor{TableLightBlue}
        \multicolumn{16}{c}{\textit{Controlled comparison (identical training data)}} \\
        \rowcolor{TableLightBlue}
        Single-step-7B (coordinate-based)
        & 49.2\resultstd{3.66} & 21.4\resultstd{1.81} & 68.0\resultstd{0.99} & 23.9\resultstd{1.05} & 57.4\resultstd{0.77} & 14.5\resultstd{1.76} & 63.2\resultstd{6.02} & 30.9\resultstd{0.91} & 75.3\resultstd{2.55} & 40.3\resultstd{5.76} & 46.1\resultstd{0.54} & 20.6\resultstd{3.44} & 60.3\resultstd{2.04} & 23.6\resultstd{0.42} & 46.3\resultstd{1.40} \\
        \rowcolor{TableLightBlue}
        Single-step-7B (coordinate-free)
        & 58.2\resultstd{1.47} & 26.6\resultstd{1.56} & 65.2\resultstd{4.32} & 36.1\resultstd{3.11} & 62.5\resultstd{0.58} & \underline{24.7}\resultstd{1.76} & 60.7\resultstd{1.75} & 41.8\resultstd{1.58} & 74.6\resultstd{1.96} & \underline{57.9}\resultstd{4.75} & 62.0\resultstd{2.16} & \underline{39.3}\resultstd{3.37} & 63.9\resultstd{0.39} & \underline{35.8}\resultstd{1.33} & 53.2\resultstd{0.52} \\
        \rowcolor{TableLightBlue}
        \textbf{ReLaViS-7B}
        & 57.9\resultstd{2.03} & 30.2\resultstd{2.39} & 69.9\resultstd{2.28} & \textbf{40.0}\resultstd{1.83} & 67.2\resultstd{0.87} & \textbf{25.6}\resultstd{0.81} & 65.5\resultstd{2.89} & \textbf{43.0}\resultstd{1.39} & 78.3\resultstd{1.18} & \textbf{62.3}\resultstd{4.99} & 63.2\resultstd{1.08} & \textbf{45.3}\resultstd{1.72} & 67.1\resultstd{0.30} & \textbf{38.9}\resultstd{0.82} & \textbf{56.3}\resultstd{0.29} \\
        \midrule
        \rowcolor{TableLightYellow}
        \multicolumn{16}{c}{\textit{Thinking with images (external tools, multi-round)}} \\
        \rowcolor{TableLightYellow}
        \begin{tabular}[c]{@{}l@{}}MVP$^\dagger$~{\tablecitep{zhang2026mvp}}\\\quad + Qwen2.5-VL-7B\end{tabular}
        & 49.8 & 12.5 & 70.1 & 25.5 & 58.6 & 11.9 & 68.1 & 22.7 & 69.5 & 41.5 & 58.9 & 27.0 & 62.0 & 22.0 & 46.7 \\
        \rowcolor{TableLightYellow}
        \begin{tabular}[c]{@{}l@{}}UI-Zoomer$^\dagger$~{\tablecitep{tang2026ui}}\\\quad + Qwen2.5-VL-7B\end{tabular}
        & 47.2 & 12.5 & 58.4 & 21.4 & 46.5 & 13.3 & 48.6 & 22.7 & 64.4 & 26.4 & 45.8 & 29.2 & 52.0 & 20.4 & 39.9 \\
        \rowcolor{TableLightYellow}
        GUI-Spotlight$^\dagger$~{\tablecitep{lei2025gui}}
        & 64.5 & 17.2 & \textbf{75.3} & 24.1 & 66.2 & 19.6 & 78.5 & 20.9 & 80.8 & 35.9 & \textbf{65.4} & 29.2 & 71.7 & 23.5 & 53.3 \\
        \rowcolor{TableLightYellow}
        GUI-Cursor~{\tablecitep{zhao2025learning}}
        & \textbf{80.5} & 33.1 & 65.7 & 18.2 & 62.4 & 25.0 & \textbf{83.3} & 32.7 & 84.2 & 43.4 & \textbf{65.4} & 31.5 & 73.3 & 29.3 & 56.5 \\
        \rowcolor{TableLightYellow}
        \begin{tabular}[c]{@{}l@{}}Zoom-in\\\quad + Qwen2.5-VL-7B\end{tabular}
        & 50.3 & 21.9 & 73.4 & 20.7 & 64.1 & 21.0 & 76.4 & 19.1 & 77.4 & 41.5 & 55.1 & 29.2 & 66.0 & 23.7 & 49.8 \\
        \rowcolor{TableLightYellow}
        \begin{tabular}[c]{@{}l@{}}Zoom-in\\\quad + single-step-7B (coordinate-free)\end{tabular}
        & 70.7\resultstd{0.59} & 32.3\resultstd{2.39} & 73.6\resultstd{1.35} & \textbf{49.4}\resultstd{3.80} & 68.7\resultstd{2.81} & 31.7\resultstd{0.40} & 73.6\resultstd{0.70} & \textbf{51.2}\resultstd{2.93} & 81.7\resultstd{1.82} & 66.7\resultstd{4.75} & 64.5\resultstd{3.74} & 53.2\resultstd{3.24} & 72.5\resultstd{0.52} & 45.8\resultstd{0.39} & 62.3\resultstd{0.33} \\
        \rowcolor{TableLightYellow}
        \textbf{\begin{tabular}[c]{@{}l@{}}Zoom-in\\\quad + ReLaViS-7B\end{tabular}}
        & 69.9\resultstd{1.06} & \textbf{34.4}\resultstd{7.16} & 75.1\resultstd{1.63} & \textbf{49.4}\resultstd{3.11} & \textbf{69.7}\resultstd{0.51} & \textbf{33.6}\resultstd{2.52} & 75.2\resultstd{0.40} & 49.4\resultstd{1.89} & \textbf{84.4}\resultstd{0.86} & \textbf{75.5}\resultstd{1.89} & \textbf{65.4}\resultstd{3.37} & \textbf{55.4}\resultstd{1.72} & \textbf{73.6}\resultstd{0.36} & \textbf{47.2}\resultstd{1.54} & \textbf{63.5}\resultstd{0.57} \\
        \bottomrule
    \end{tabular}%
    }
\end{table}

\begin{table}[t]
    \centering
    \caption{Performance comparison on MMBench-GUI-L2.}
    \label{tab:mmbench-gui-l2-detailed}
    \fittable{%
    \setlength{\tabcolsep}{1pt}
    \begin{tabular}{l*{6}{cc}c}
        \toprule
        \multirow{2}{*}[-0.4ex]{\textbf{Model}}
        & \multicolumn{2}{c}{\textbf{Windows}} & \multicolumn{2}{c}{\textbf{macOS}}
        & \multicolumn{2}{c}{\textbf{Linux}} & \multicolumn{2}{c}{\textbf{iOS}}
        & \multicolumn{2}{c}{\textbf{Android}} & \multicolumn{2}{c}{\textbf{Web}}
        & \multirow{2}{*}[-0.4ex]{\textbf{Avg}} \\
        \cmidrule(lr){2-3}\cmidrule(lr){4-5}\cmidrule(lr){6-7}
        \cmidrule(lr){8-9}\cmidrule(lr){10-11}\cmidrule(lr){12-13}
        & \textbf{Basic} & \textbf{Adv.} & \textbf{Basic} & \textbf{Adv.}
        & \textbf{Basic} & \textbf{Adv.} & \textbf{Basic} & \textbf{Adv.}
        & \textbf{Basic} & \textbf{Adv.} & \textbf{Basic} & \textbf{Adv.} & \\
        \midrule
        \multicolumn{14}{c}{\textit{Zero-shot general VLMs}} \\
        GPT-4o~\tablecitep{hurst2024gpt}
        & 1.5 & 1.1 & 8.7 & 4.3 & 1.1 & 1.0 & 5.1 & 3.3 & 2.5 & 1.4 & 3.2 & 2.9 & 2.9 \\
        Qwen2.5-VL-3B$^\dagger$~\tablecitep{bai2025qwen25vltechnicalreport}
        & 57.2 & 26.5 & 62.3 & 31.5 & 49.2 & 23.5 & 81.2 & 56.4 & 83.7 & 57.2 & 70.3 & 42.5 & 55.2 \\
        Qwen2.5-VL-7B$^\dagger$~\tablecitep{bai2025qwen25vltechnicalreport}
        & 68.6 & 45.2 & 70.7 & 54.6 & 48.7 & 35.7 & \underline{94.9} & 82.4 & 93.8 & 80.9 & 86.5 & 70.8 & 71.9 \\
        \midrule
        \multicolumn{14}{c}{\textit{GUI-specialized models (coordinate-based)}} \\
        SE-GUI-7B~\tablecitep{yuan2026se}
        & -- & -- & -- & -- & -- & -- & -- & -- & -- & -- & -- & -- & 34.5 \\
        Jedi-7B~\tablecitep{xie2026scaling}
        & -- & -- & -- & -- & -- & -- & -- & -- & -- & -- & -- & -- & 70.4 \\
        UI-TARS-1.5-7B~\tablecitep{qin2025ui}
        & 68.3 & 39.0 & 69.0 & 44.5 & 64.4 & 37.8 & 88.5 & 69.4 & 90.5 & 69.3 & 81.0 & 56.5 & 64.3 \\
        GUI-G$^2$-7B~\tablecitep{tang2026gui}
        & -- & -- & -- & -- & -- & -- & -- & -- & -- & -- & -- & -- & 79.5 \\
        InfiGUI-G1-7B~\tablecitep{liu2026infigui}
        & 82.7 & \textbf{61.8} & 83.8 & 63.9 & 72.3 & 52.0 & \underline{94.9} & \textbf{89.4} & \underline{95.2} & \textbf{85.6} & \textbf{93.5} & \textbf{76.3} & \underline{80.8} \\
        GroundNext-7B (SFT)~\tablecitep{feizi2026grounding}
        & \underline{83.8} & \underline{60.7} & \textbf{86.7} & 69.9 & \underline{75.4} & \underline{61.2} & 94.3 & 83.3 & 94.9 & 79.4 & 91.0 & 70.5 & 80.4 \\
        \midrule
        \multicolumn{14}{c}{\textit{GUI-specialized models (coordinate-free)}} \\
        GUI-Actor-7B~\tablecitep{wu2026gui}
        & -- & -- & -- & -- & -- & -- & -- & -- & -- & -- & -- & -- & 70.9 \\
        V2P-7B~\tablecitep{chen2025v2p}
        & -- & -- & -- & -- & -- & -- & -- & -- & -- & -- & -- & -- & 79.9 \\
        \midrule
        \rowcolor{TableLightBlue}
        \multicolumn{14}{c}{\textit{Controlled comparison (identical training data)}} \\
        \rowcolor{TableLightBlue}
        Single-step-7B (coordinate-based)
        & 82.9\resultstd{1.18} & 55.4\resultstd{2.03} & 82.0\resultstd{1.33} & 66.5\resultstd{0.87} & 71.6\resultstd{0.31} & 49.0\resultstd{0.51} & \textbf{95.0}\resultstd{0.36} & 83.4\resultstd{0.93} & \textbf{96.0}\resultstd{0.32} & 81.3\resultstd{0.43} & \underline{91.2}\resultstd{0.37} & 71.3\resultstd{1.85} & 78.7\resultstd{0.30} \\
        \rowcolor{TableLightBlue}
        Single-step-7B (coordinate-free)
        & 82.2\resultstd{0.85} & 57.0\resultstd{1.28} & \underline{84.9}\resultstd{1.00} & \textbf{72.1}\resultstd{0.88} & 74.0\resultstd{1.84} & 55.4\resultstd{2.12} & 94.2\resultstd{1.21} & \underline{86.7}\resultstd{0.81} & 93.5\resultstd{0.74} & 83.4\resultstd{1.23} & 88.7\resultstd{1.41} & 73.4\resultstd{1.17} & 80.2\resultstd{0.55} \\
        \rowcolor{TableLightBlue}
        \textbf{ReLaViS-7B}
        & \textbf{85.6}\resultstd{0.98} & 58.8\resultstd{0.37} & 84.4\resultstd{0.89} & \underline{71.5}\resultstd{2.69} & \textbf{76.1}\resultstd{1.09} & \textbf{61.7}\resultstd{0.88} & 93.7\resultstd{0.74} & 85.4\resultstd{0.17} & 93.0\resultstd{0.85} & \underline{83.6}\resultstd{0.32} & 89.4\resultstd{0.85} & \underline{76.1}\resultstd{0.82} & \textbf{81.1}\resultstd{0.08} \\
        \midrule
        \rowcolor{TableLightYellow}
        \multicolumn{14}{c}{\textit{Thinking with images (external tools, multi-round)}} \\
        \rowcolor{TableLightYellow}
        \begin{tabular}[c]{@{}l@{}}MVP$^\dagger$~\tablecitep{zhang2026mvp}\\\quad + Qwen2.5-VL-7B\end{tabular}
        & 63.1 & 41.9 & 54.2 & 45.7 & 44.5 & 39.3 & 69.4 & 63.6 & 56.7 & 61.4 & 54.5 & 54.6 & 55.0 \\
        \rowcolor{TableLightYellow}
        \begin{tabular}[c]{@{}l@{}}UI-Zoomer$^\dagger$~\tablecitep{tang2026ui}\\\quad + Qwen2.5-VL-7B\end{tabular}
        & 77.1 & 50.0 & 81.7 & 60.1 & 75.4 & 49.5 & 90.8 & 73.6 & \textbf{91.9} & 72.7 & 89.0 & 65.3 & 74.2 \\
        \rowcolor{TableLightYellow}
        GUI-Spotlight$^\dagger$~\tablecitep{lei2025gui}
        & 77.5 & 50.4 & 77.4 & 53.8 & 63.9 & 45.9 & 91.7 & 75.2 & 85.7 & 70.4 & 88.7 & 70.1 & 72.2 \\
        \rowcolor{TableLightYellow}
        GUI-Cursor~\tablecitep{zhao2025learning}
        & -- & -- & -- & -- & -- & -- & -- & -- & -- & -- & -- & -- & -- \\
        \rowcolor{TableLightYellow}
        \begin{tabular}[c]{@{}l@{}}Zoom-in\\\quad + Qwen2.5-VL-7B\end{tabular}
        & 81.9 & 58.8 & 81.2 & 67.9 & 69.6 & 52.6 & 90.1 & \textbf{77.9} & 90.2 & \textbf{77.8} & 91.0 & 72.4 & 77.3 \\
        \rowcolor{TableLightYellow}
        \begin{tabular}[c]{@{}l@{}}Zoom-in\\\quad + single-step-7B (coordinate-free)\end{tabular}
        & \textbf{87.0}\resultstd{0.42} & \textbf{64.6}\resultstd{1.81} & 88.4\resultstd{1.00} & \textbf{75.7}\resultstd{0.77} & 81.0\resultstd{1.99} & 61.9\resultstd{0.78} & 88.6\resultstd{0.48} & 77.8\resultstd{1.75} & 89.5\resultstd{0.59} & 76.9\resultstd{1.47} & 90.8\resultstd{0.98} & 74.8\resultstd{1.84} & 80.5\resultstd{0.11} \\
        \rowcolor{TableLightYellow}
        \textbf{\begin{tabular}[c]{@{}l@{}}Zoom-in\\\quad + ReLaViS-7B\end{tabular}}
        & 86.0\resultstd{1.33} & 64.0\resultstd{1.32} & \textbf{89.7}\resultstd{1.17} & 75.4\resultstd{0.29} & \textbf{82.2}\resultstd{1.05} & \textbf{64.1}\resultstd{1.06} & \textbf{91.8}\resultstd{0.49} & 76.4\resultstd{1.09} & 90.5\resultstd{0.65} & 75.4\resultstd{0.59} & \textbf{91.9}\resultstd{0.97} & \textbf{75.3}\resultstd{0.56} & \textbf{80.9}\resultstd{0.16} \\
        \bottomrule
    \end{tabular}%
    }
\end{table}

\begin{table}[t]
    \centering
    \caption{Performance comparison on OSWorld-G-Refine.}
    \label{tab:osworld-g-refine-detailed}
    \fittable{%
    \setlength{\tabcolsep}{4pt}
    \begin{tabular}{l*{4}{w{c}{6em}}c}
        \toprule
        \multirow{2}{*}{\textbf{Model}}
        & \textbf{Text}
        & \textbf{Element}
        & \textbf{Layout}
        & \textbf{Fine-grained}
        & \multirow{2}{*}{\textbf{Overall}} \\
        & \textbf{matching}
        & \textbf{recognition}
        & \textbf{understanding}
        & \textbf{manipulation}
        & \\
        \midrule
        \multicolumn{6}{c}{\textit{Zero-shot general VLMs}} \\
        Operator~\tablecitep{cua2025}
        & 51.3 & 42.4 & 46.6 & 31.5 & 40.6 \\
        Qwen2.5-VL-3B$^\dagger$~\tablecitep{bai2025qwen25vltechnicalreport}
        & 52.1 & 41.5 & 46.6 & 22.4 & 38.8 \\
        Qwen2.5-VL-7B$^\dagger$~\tablecitep{bai2025qwen25vltechnicalreport}
        & 57.5 & 46.1 & 51.0 & 31.6 & 44.0 \\
        \midrule
        \multicolumn{6}{c}{\textit{GUI-specialized models (coordinate-based)}} \\
        SE-GUI-7B~\tablecitep{yuan2026se}
        & -- & -- & -- & -- & 33.9 \\
        Jedi-7B~\tablecitep{xie2026scaling}
        & 65.9 & 55.5 & 57.7 & 46.9 & 54.1 \\
        UI-TARS-1.5-7B~\tablecitep{qin2025ui}
        & 52.6 & \textbf{75.4} & \textbf{72.4} & \textbf{66.7} & \textbf{64.2} \\
        GUI-G$^2$-7B~\tablecitep{tang2026gui}
        & -- & -- & -- & -- & 61.9 \\
        InfiGUI-G1-7B~\tablecitep{liu2026infigui}
        & \textbf{72.0} & 63.6 & 66.8 & 46.3 & 59.9 \\
        \midrule
        \multicolumn{6}{c}{\textit{GUI-specialized models (coordinate-free)}} \\
        GUI-Actor-7B~\tablecitep{wu2026gui}
        & -- & -- & -- & -- & 47.0 \\
        V2P-7B~\tablecitep{chen2025v2p}
        & -- & -- & -- & -- & 52.5 \\
        \midrule
        \rowcolor{TableLightBlue}
        \multicolumn{6}{c}{\textit{Controlled comparison (identical training data)}} \\
        \rowcolor{TableLightBlue}
        Single-step-7B (coordinate-based)
        & 68.2\resultstd{1.01} & 64.7\resultstd{0.17} & 65.2\resultstd{1.05} & \underline{47.6}\resultstd{0.38} & 59.6\resultstd{0.20} \\
        \rowcolor{TableLightBlue}
        Single-step-7B (coordinate-free)
        & 67.4\resultstd{1.01} & 66.9\resultstd{1.75} & 68.4\resultstd{2.20} & 38.4\resultstd{0.38} & 59.3\resultstd{1.18} \\
        \rowcolor{TableLightBlue}
        \textbf{ReLaViS-7B}
        & \underline{68.8}\resultstd{0.80} & \underline{71.4}\resultstd{1.43} & \underline{71.2}\resultstd{0.00} & 40.4\resultstd{2.01} & \underline{62.5}\resultstd{1.01} \\
        \midrule
        \rowcolor{TableLightYellow}
        \multicolumn{6}{c}{\textit{Thinking with images (external tools, multi-round)}} \\
        \rowcolor{TableLightYellow}
        MVP$^\dagger$~\smalltablecitep{zhang2026mvp} + Qwen2.5-VL-7B
        & 61.7 & 53.9 & 60.9 & 36.8 & 51.2 \\
        \rowcolor{TableLightYellow}
        UI-Zoomer$^\dagger$~\smalltablecitep{tang2026ui} + Qwen2.5-VL-7B
        & 61.7 & 60.9 & 61.3 & 40.4 & 54.9 \\
        \rowcolor{TableLightYellow}
        GUI-Spotlight$^\dagger$~\tablecitep{lei2025gui}
        & 68.2 & 60.6 & 63.6 & 45.4 & 56.7 \\
        \rowcolor{TableLightYellow}
        GUI-Cursor~\tablecitep{zhao2025learning}
        & 70.1 & 58.5 & 62.1 & 52.3 & 58.0 \\
        \rowcolor{TableLightYellow}
        Zoom-in + Qwen2.5-VL-7B
        & 67.1 & 66.4 & 70.0 & 40.1 & 59.2 \\
        \rowcolor{TableLightYellow}
        Zoom-in + single-step-7B (coordinate-free)
        & \textbf{73.9}\resultstd{1.01} & 75.0\resultstd{0.93} & 77.3\resultstd{1.60} & \textbf{53.7}\resultstd{1.66} & \textbf{68.3}\resultstd{1.07} \\
        \rowcolor{TableLightYellow}
        \textbf{Zoom-in + ReLaViS-7B}
        & \textbf{73.9}\resultstd{1.33} & \textbf{75.3}\resultstd{0.46} & \textbf{77.6}\resultstd{1.60} & 52.4\resultstd{1.37} & 68.2\resultstd{0.54} \\
        \bottomrule
    \end{tabular}
    }
\end{table}

\begin{table}[t]
    \centering
    \caption{Performance comparison on ScreenSpot-v2.}
    \label{tab:screenspot-v2-detailed}
    \fittable{%
    \setlength{\tabcolsep}{1pt}
    \begin{tabular}{lccccccc}
        \toprule
        \multirow{2}{*}[-0.6ex]{\textbf{Model}}
        & \multicolumn{2}{c}{\textbf{Mobile}}
        & \multicolumn{2}{c}{\textbf{Desktop}}
        & \multicolumn{2}{c}{\textbf{Web}}
        & \multirow{2}{*}[-0.6ex]{\textbf{Avg}} \\
        \cmidrule(lr){2-3}\cmidrule(lr){4-5}\cmidrule(lr){6-7}
        & \textbf{Text} & \textbf{Icon}
        & \textbf{Text} & \textbf{Icon}
        & \textbf{Text} & \textbf{Icon} & \\
        \midrule
        \multicolumn{8}{c}{\textit{Zero-shot general VLMs}} \\
        GPT-4o~\smalltablecitep{hurst2024gpt} + OmniParser-v2
        & 95.5 & 74.6 & 92.3 & 60.9 & 88.0 & 59.6 & 80.7 \\
        Operator~\tablecitep{cua2025}
        & 47.3 & 41.5 & 90.2 & 80.3 & 92.8 & 84.3 & 70.5 \\
        Qwen2.5-VL-3B$^\dagger$~\tablecitep{bai2025qwen25vltechnicalreport}
        & 83.3 & 64.9 & 76.1 & 45.2 & 73.2 & 62.0 & 69.6 \\
        Qwen2.5-VL-7B$^\dagger$~\tablecitep{bai2025qwen25vltechnicalreport}
        & 98.8 & 87.4 & 90.0 & 73.8 & 92.9 & 78.9 & 88.4 \\
        Qwen3-VL-8B~\tablecitep{bai2025qwen3}
        & \textbf{99.7} & 87.7 & 94.8 & 83.6 & 95.3 & 85.7 & 92.1 \\
        Qwen3-VL-30B-A3B~\tablecitep{bai2025qwen3}
        & \underline{99.0} & 87.7 & 95.4 & 82.9 & 95.3 & 83.7 & 91.7 \\
        \midrule
        \multicolumn{8}{c}{\textit{GUI-specialized models (coordinate-based)}} \\
        SE-GUI-7B~\tablecitep{yuan2026se}
        & -- & -- & -- & -- & -- & -- & 90.3 \\
        Jedi-7B~\tablecitep{xie2026scaling}
        & 96.9 & 87.2 & 95.9 & 87.9 & 94.4 & 84.2 & 91.7 \\
        UI-TARS-1.5-7B~\tablecitep{qin2025ui}
        & 95.9 & 84.8 & 94.9 & 80.7 & 90.6 & 86.2 & 89.7 \\
        GUI-G$^2$-7B~\tablecitep{tang2026gui}
        & 98.3 & \textbf{91.9} & 95.4 & 89.3 & 94.0 & 87.7 & 93.3 \\
        InfiGUI-G1-7B~\tablecitep{liu2026infigui}
        & \underline{99.0} & \textbf{91.9} & 94.3 & 82.1 & \textbf{97.9} & \underline{89.2} & \underline{93.5} \\
        UI-Ins-7B~\tablecitep{chen2026ui}
        & \underline{99.0} & \underline{90.5} & \textbf{97.9} & 81.4 & \underline{97.4} & \textbf{91.6} & \textbf{94.0} \\
        GroundNext-7B (SFT)~\tablecitep{feizi2026grounding}
        & 97.2 & 84.8 & 94.3 & 90.0 & 91.5 & 74.9 & 89.3 \\
        GroundNext-7B (RL)~\tablecitep{feizi2026grounding}
        & 96.6 & 88.2 & 95.4 & 87.9 & 94.9 & 75.9 & 90.4 \\
        \midrule
        \multicolumn{8}{c}{\textit{GUI-specialized models (coordinate-free)}} \\
        GUI-Actor-7B~\tablecitep{wu2026gui}
        & 97.6 & 88.2 & 96.9 & 85.7 & 93.2 & 86.7 & 92.1 \\
        V2P-7B~\tablecitep{chen2025v2p}
        & 98.1 & 88.0 & 96.1 & 89.7 & 95.4 & 84.4 & 92.4 \\
        \midrule
        \rowcolor{TableLightBlue}
        \multicolumn{8}{c}{\textit{Controlled comparison (identical training data)}} \\
        \rowcolor{TableLightBlue}
        Single-step-7B (coordinate-based)
        & 98.3\resultstd{0.45} & 89.5\resultstd{0.52} & \underline{97.0}\resultstd{0.32} & 83.9\resultstd{1.99} & 91.1\resultstd{0.36} & 78.1\resultstd{2.92} & 90.0\resultstd{0.75} \\
        \rowcolor{TableLightBlue}
        Single-step-7B (coordinate-free)
        & 97.8\resultstd{0.44} & 89.4\resultstd{0.80} & 94.1\resultstd{0.32} & \underline{90.5}\resultstd{1.37} & 92.4\resultstd{0.54} & 78.1\resultstd{1.46} & 90.4\resultstd{0.48} \\
        \rowcolor{TableLightBlue}
        \textbf{ReLaViS-7B}
        & 96.1\resultstd{0.67} & 87.3\resultstd{0.80} & 94.8\resultstd{0.65} & \textbf{92.3}\resultstd{0.46} & 93.1\resultstd{0.54} & 80.9\resultstd{0.24} & 90.7\resultstd{0.21} \\
        \midrule
        \rowcolor{TableLightYellow}
        \multicolumn{8}{c}{\textit{Thinking with images (external tools, multi-round)}} \\
        \rowcolor{TableLightYellow}
        MVP$^\dagger$~\smalltablecitep{zhang2026mvp} + Qwen2.5-VL-7B
        & 86.8 & 61.3 & 84.4 & 51.6 & 81.4 & 61.6 & 73.3 \\
        \rowcolor{TableLightYellow}
        UI-Zoomer$^\dagger$~\smalltablecitep{tang2026ui} + Qwen2.5-VL-7B
        & \textbf{98.5} & \textbf{85.9} & \textbf{97.2} & 87.3 & 91.1 & 82.7 & 90.7 \\
        \rowcolor{TableLightYellow}
        GUI-Spotlight$^\dagger$~\tablecitep{lei2025gui}
        & 97.7 & 79.1 & 95.0 & 76.2 & \textbf{93.2} & \textbf{84.0} & 88.8 \\
        \rowcolor{TableLightYellow}
        GUI-Cursor~\tablecitep{zhao2025learning}
        & -- & -- & -- & -- & -- & -- & \textbf{93.9} \\
        \rowcolor{TableLightYellow}
        Zoom-in + Qwen2.5-VL-7B
        & 89.5 & 83.3 & \textbf{97.2} & 83.3 & 90.0 & 83.5 & 88.1 \\
        \rowcolor{TableLightYellow}
        Zoom-in + single-step-7B (coordinate-free)
        & 84.2\resultstd{2.37} & 77.8\resultstd{2.88} & 92.8\resultstd{1.47} & 88.6\resultstd{2.00} & 88.1\resultstd{0.83} & 78.3\resultstd{0.87} & 84.7\resultstd{0.41} \\
        \rowcolor{TableLightYellow}
        \textbf{Zoom-in + ReLaViS-7B}
        & 84.2\resultstd{0.98} & 76.4\resultstd{1.39} & 93.3\resultstd{0.00} & \textbf{90.7}\resultstd{2.42} & 87.5\resultstd{0.71} & 79.7\resultstd{0.43} & 84.9\resultstd{0.59} \\
        \bottomrule
    \end{tabular}
    }
\end{table}

\begin{table}[t]
    \centering
    \caption{Performance comparison on UI-Vision.}
    \label{tab:ui-vision-detailed}
    \fittable{%
    \begin{tabular}{lcccc}
        \toprule
        \textbf{Model}
        & \textbf{Basic}
        & \textbf{Functional}
        & \textbf{Spatial}
        & \textbf{Overall} \\
        \midrule
        \multicolumn{5}{c}{\textit{Zero-shot general VLMs}} \\
        GPT-4o~\tablecitep{hurst2024gpt}
        & 1.6 & 1.5 & 1.0 & 1.4 \\
        Qwen2.5-VL-3B$^\dagger$~\tablecitep{bai2025qwen25vltechnicalreport}
        & 5.3 & 11.9 & 5.2 & 7.4 \\
        Qwen2.5-VL-7B$^\dagger$~\tablecitep{bai2025qwen25vltechnicalreport}
        & 17.4 & 19.1 & 7.4 & 14.7 \\
        Qwen3-VL-8B~\tablecitep{bai2025qwen3}
        & 27.8 & 29.6 & 9.4 & 21.9 \\
        Qwen3-VL-30B-A3B~\tablecitep{bai2025qwen3}
        & 31.2 & 31.9 & 14.6 & 25.6 \\
        \midrule
        \multicolumn{5}{c}{\textit{GUI-specialized models (coordinate-based)}} \\
        SE-GUI-7B~\tablecitep{yuan2026se}
        & -- & -- & -- & 16.7 \\
        Jedi-7B~\tablecitep{xie2026scaling}
        & 32.3 & 30.5 & 12.8 & 24.8 \\
        UI-TARS-1.5-7B~\tablecitep{qin2025ui}
        & -- & -- & -- & 20.8 \\
        GUI-G$^2$-7B~\tablecitep{tang2026gui}
        & -- & -- & -- & 25.6 \\
        InfiGUI-G1-7B~\tablecitep{liu2026infigui}
        & 36.2 & 31.9 & 11.5 & 26.1 \\
        \midrule
        \multicolumn{5}{c}{\textit{GUI-specialized models (coordinate-free)}} \\
        GUI-Actor-7B~\tablecitep{wu2026gui}
        & -- & -- & -- & 21.9 \\
        V2P-7B~\tablecitep{chen2025v2p}
        & -- & -- & -- & 28.8 \\
        \midrule
        \rowcolor{TableLightBlue}
        \multicolumn{5}{c}{\textit{Controlled comparison (identical training data)}} \\
        \rowcolor{TableLightBlue}
        Single-step-7B (coordinate-based)
        & 40.5\resultstd{0.30} & 38.4\resultstd{0.70} & 21.3\resultstd{0.30} & 33.4\resultstd{0.35} \\
        \rowcolor{TableLightBlue}
        Single-step-7B (coordinate-free)
        & \underline{54.7}\resultstd{0.41} & \underline{50.9}\resultstd{0.67} & \underline{23.8}\resultstd{1.62} & \underline{43.1}\resultstd{0.42} \\
        \rowcolor{TableLightBlue}
        \textbf{ReLaViS-7B}
        & \textbf{62.6}\resultstd{0.42} & \textbf{57.9}\resultstd{0.55} & \textbf{28.7}\resultstd{0.98} & \textbf{49.7}\resultstd{0.35} \\
        \midrule
        \rowcolor{TableLightYellow}
        \multicolumn{5}{c}{\textit{Thinking with images (external tools, multi-round)}} \\
        \rowcolor{TableLightYellow}
        MVP$^\dagger$~\smalltablecitep{zhang2026mvp} + Qwen2.5-VL-7B
        & 24.3 & 24.0 & 10.4 & 19.6 \\
        \rowcolor{TableLightYellow}
        UI-Zoomer$^\dagger$~\smalltablecitep{tang2026ui} + Qwen2.5-VL-7B
        & 31.0 & 28.4 & 12.5 & 23.9 \\
        \rowcolor{TableLightYellow}
        GUI-Spotlight$^\dagger$~\tablecitep{lei2025gui}
        & 31.6 & 30.2 & 8.8 & 23.5 \\
        \rowcolor{TableLightYellow}
        GUI-Cursor~\tablecitep{zhao2025learning}
        & 35.0 & 31.2 & 12.1 & 25.7 \\
        \rowcolor{TableLightYellow}
        Zoom-in + Qwen2.5-VL-7B
        & 28.1 & 28.7 & 14.7 & 23.9 \\
        \rowcolor{TableLightYellow}
        Zoom-in + single-step-7B (coordinate-free)
        & 57.2\resultstd{0.57} & 53.6\resultstd{0.34} & 31.1\resultstd{0.54} & 47.3\resultstd{0.44} \\
        \rowcolor{TableLightYellow}
        \textbf{Zoom-in + ReLaViS-7B}
        & \textbf{62.5}\resultstd{0.09} & \textbf{57.1}\resultstd{0.75} & \textbf{37.3}\resultstd{0.80} & \textbf{52.3}\resultstd{0.36} \\
        \bottomrule
    \end{tabular}
    }
\end{table}

\end{document}